\documentclass[times, review, 10pt]{elsarticle}

\usepackage{algorithm}
\usepackage{algorithmic}
\usepackage{amsmath}
\usepackage{amssymb}
\usepackage{multirow}
\usepackage{makecell}
\usepackage{caption}
\usepackage{enumitem}
\usepackage{url}
\usepackage{booktabs} 
\usepackage{hyperref}
\usepackage{placeins}  
\FloatBarrier

\begin{document}

\begin{frontmatter}

\title{Dynamic Graph Structure Learning via Resistance Curvature Flow}

\author[label1]{Chaoqun Fei}
\author[label1]{Huanjiang Liu}
\author[label1]{Tinglve Zhou}
\author[label2]{Yangyang Li \footnote{Corresponding author: Yangyang Li.}}
\author[label3]{Tianyong Hao}
\affiliation[label1]{organization={School of Artificial Intelligence, South China Normal University},
            city={Foshan},
            postcode={528225},
            state={Guangdong},
            country={China}}

\affiliation[label2]{organization={School of Computer Science, South China Normal University},
            city={Guangzhou},
            postcode={510631},
            state={Guangdong},
            country={China}}
            
\affiliation[label3]{organization={State Key Laboratory of Mathematical Sciences, Academy of Mathematics and Systems Science, Chinese Academy of Sciences},
            city={Beijing},
            postcode={100190},
            country={China}}

\begin{abstract}
Geometric Representation Learning (GRL) aims to approximate the non-Euclidean topology of high-dimensional data through discrete graph structures, grounded in the manifold hypothesis. However, traditional static graph construction methods based on Euclidean distance often fail to capture the intrinsic curvature characteristics of the data manifold. Although Ollivier-Ricci Curvature Flow (OCF) has proven to be a powerful tool for dynamic topological optimization, its core reliance on Optimal Transport (Wasserstein distance) leads to prohibitive computational complexity, severely limiting its application in large-scale datasets and deep learning frameworks.
To break this bottleneck, this paper proposes a novel geometric evolution framework: Resistance Curvature Flow (RCF). Leveraging the concept of effective resistance from circuit physics, RCF transforms expensive curvature optimization into efficient matrix operations. This approach achieves over 100$\times$ computational acceleration while maintaining geometric optimization capabilities comparable to OCF. We provide an in-depth exploration of the theoretical foundations and dynamical principles of RCF, elucidating how it guides the redistribution of edge weights via curvature gradients to eliminate topological noise and strengthen local cluster structures.
Furthermore, we provide a mechanistic explanation of RCF’s role in manifold enhancement and noise suppression, as well as its compatibility with deep learning models. We design a graph optimization algorithm, DGSL-RCF, based on this framework. Experimental results across deep metric learning, manifold learning, and graph structure learning demonstrate that DGSL-RCF significantly improves representation quality and downstream task performance.
\end{abstract}

\begin{highlights}
\item Introduces a novel and computationally efficient curvature flow based on effective resistance from circuit theory, establishing a new paradigm for geometric graph structure evolution.
\item Formulates the dynamic evolution equation of RCF, elucidates its mechanisms for manifold enhancement and noise suppression, and highlights its differentiability and compatibility with deep learning frameworks.
\item Proposes an efficient Dynamic Graph Structure Learning method based on RCF. Extensive experiments on deep metric learning, manifold learning and graph structure learning tasks demonstrate that DGSL-RCF consistently improves representation quality and downstream performance with low runtime cost.
\end{highlights}

\begin{keyword}
Ricci Flow \sep Resistance Curvature \sep Geometric Representation Learning
\end{keyword}

\end{frontmatter}


\section{Introduction}
Geometric Representation Learning (GRL) has emerged as a fundamental paradigm in modern machine learning, aiming to extract low-dimensional embeddings from high-dimensional raw data that faithfully reflects intrinsic geometric structures and metric relationships~\cite{Bronstein2021GeometricDL}. This field is largely built upon the "manifold hypothesis"~\cite{whiteley2025statistical}, which posits that high-dimensional data observed in real-world applications typically concentrates near a low-dimensional latent manifold. Consequently, the core challenge of GRL lies in accurately characterizing and manipulating this hidden manifold structure. Since manifolds are inherently continuous geometric objects, researchers typically introduce discrete graph structures (e.g., k-nearest neighbor graphs) as approximations to represent their non-Euclidean topology for computational tractability~\cite{Costa2004}. This discretization allows us to computationally process and analyze otherwise abstract manifold geometries.

Accordingly, the quality of the graph structure directly dictates its approximation fidelity to the underlying manifold. A precise geometric representation ensures the preservation of critical structure properties, such as geodesic distances, within the embedding space, thereby providing beneficial geometric priors for downstream tasks.

However, traditional graph construction methods (e.g., Euclidean distance-based k-NN) are inherently shallow and static. They often rely on initial known metrics, specifically Euclidean distance, failing to fully account for the intrinsic non-Euclidean geometric characteristics of the data distribution. In differential geometry, such non-Euclidean properties are primarily described by curvature~\cite{JamesWCannon2018}, which governs metric and volume fluctuations on the manifold. Given that the curvature distribution of a data manifold is usually unknown and complex, initial static graphs struggle to accurately capture its dynamically evolving geometric structure.

To overcome these limitations, inspired by the concept of Ricci flow in differential geometry, our prior work~\cite{FeiAndLi2025} introduces a geometric flow mechanism. This framework establishes a variational scheme where the metric tensor evolves dynamically according to curvature. The essence of geometric flow~\cite{Ni2019CommunityDO} is to treat the graph structure as a physical system in a non-equilibrium state, utilizing curvature-driven gradient flow to continuously perform nonlinear reconstruction of local metrics. This process effectively corrects geometric distortions induced by the original Euclidean metric, allowing the discrete graph topology to gradually converge toward the ideal continuous manifold structure during evolution.

Currently, the Ollivier-Ricci Curvature Flow (OCF)~\cite{Sia2019CommunityDO, NiAndLin2018, ABATE2025111648, LI2018273} is regarded as one of the most refined geometric flow tools for characterizing the topology of discrete manifolds. This represents an evolutionary mechanism on a graph, driven by the Ollivier-Ricci curvature. Mathematically, OCF sensitively captures community structures through the distribution of positive and negative curvature in its evolution equation. Specifically, intra-community edges exhibit positive curvature condensation, while inter-community edges exhibit negative curvature divergence. This superior topological discriminative power has demonstrated immense potential across various graph representation learning scenarios~\cite{YeAndLiu2019, LIAndJun2022}. However, its sophisticated characterization capability comes at a heavy computational cost: the discretization of the underlying Ollivier-Ricci curvature relies heavily on calculating Wasserstein distances (Optimal Transport), leading to an algorithmic complexity of $O(n^4log^2n)$~\cite{pal2017efficientalternative}. This prohibitive overhead not only hinders the scalability of OCF to large-scale datasets but also makes it difficult to adapt to the real-time, high-frequency iterative optimization requirements of deep learning.

To break this deadlock, in our latest work, we proposed an efficient Resistance Curvature ~\cite{fei2025efficientcurvatureawaregraphnetwork}. Unlike Ollivier-Ricci curvature, which is based on shortest paths and optimal transport, Resistance Curvature cleverly leverages the concept of effective resistance from circuit physics to characterize the geometric connectivity of graph structures. Mathematically, Resistance Curvature transforms the expensive optimization problem of curvature calculation into a series of efficient matrix operations. This shift enables Resistance Curvature to achieve over 100x computational acceleration while maintaining geometric optimization capabilities comparable to OCF.

Building upon~\cite{fei2025efficientcurvatureawaregraphnetwork}, this paper introduces Resistance Curvature Flow (RCF). The core focus of this research is to explore the theoretical foundations, mathematical properties, and dynamical principles of RCF. We provide an in-depth analysis of how RCF guides the redistribution of edge weights via curvature gradients to eliminate topological noise and strengthen local cluster structures. Furthermore, we propose a graph structure optimization algorithm based on RCF and integrate it into neural networks using deep metric learning (DML) as a case study, providing an application paradigm for graph curvature flow within deep learning models. Additionally, we provide practical applications in manifold learning (ML) and graph structure learning (GSL) to verify the universality and robustness of RCF across multiple tasks.

The primary contributions of this paper are summarized as follows:
\begin{itemize}[topsep=0pt, parsep=0pt]
    \item Proposed the RCF theoretical framework. We formally define Resistance Curvature Flow for the first time, establishing its mathematical foundation and rigorously analyzing its geometric dynamics during graph evolution.
    \item Designed RCF-based dynamic graph learning algorithms and paradigms. We propose an RCF algorithm capable of dynamically optimizing graph structures and offer three integration paradigms, including pre-processing optimization, internal embedding regularization, and output refinement, to enhance the geometric awareness of existing models.
    \item In-depth analysis of RCF effectiveness and compatibility. We explain the mechanism of RCF in manifold enhancement and noise suppression and demonstrate its potential as a built-in deep learning operator from the perspectives of computational efficiency and differential compatibility.
    \item Extensive experimental validation. Experiments across various tasks demonstrate that RCF significantly improves representation quality and downstream performance while maintaining training efficiency superior to existing geometric methods.
\end{itemize}

The full forms of all abbreviations used in this work are provided in Appendix Table~\ref{tab:abbr}.

\section{Related Work}
GRL encompasses a variety of core tasks that share a common objective: extracting and preserving the intrinsic low-dimensional manifold structure from high-dimensional data. The RCF framework proposed in this study is validated and applied primarily across the following three representative tasks:

\subsection{Deep Metric Learning}

The objective of DML is to learn an embedding space via deep neural networks where geometric distances (e.g., Euclidean distance or cosine similarity) between data points accurately reflect their semantic similarity.

Existing research has focused primarily on loss function design and hard-sample mining strategies~\cite{sym11091066}, with representative losses including Triplet Loss~\cite{SchroffAndKalenichenko2015}, Semi-Triplet Loss~\cite{YuanAndChen2020}, N-Pairs Loss~\cite{SohnKihyuk2016}, and Multi-Similarity Loss~\cite{WangAndHan2019}. While these methods construct metric spaces through contrastive constraints, they often overlook the implicit continuous manifold structure within the embedding space. A significant challenge in DML is that the feature space may undergo geometric distortion during training, causing local metric structures to become inconsistent with the intrinsic manifold. Thus, geometric constraints are needed to enhance the stability and discriminative power of the latent space. Recent studies~\cite{LiAndFei2023} have attempted to introduce approximate geodesic distances as geometric constraints, proving that integrating manifold information can significantly bolster feature discriminability.

Nevertheless, such constraints typically rely on pre-computed static and coarse graph structures, which struggle to adapt to the high-frequency evolving feature topology in deep networks. In contrast, the RCF-based graph structure learning algorithm proposed here acts as a dynamic geometric optimizer. it refines the latent space metric structure in real-time and further mines the model’s geometric representation potential by suppressing "geometric noise".

\subsection{Manifold Learning}

As a core technique for nonlinear dimensionality reduction, manifold learning is predicated on the hypothesis that high-dimensional data actually lie on a low-dimensional manifold. The goal is to discover and "unfold" this intrinsic structure. Based on the structural properties preserved during the learning process, mainstream methods can be categorized into two groups: 1) Global Structure Preservation: These methods aim to replicate the global relationships (e.g., distances or density) between all data points in the high-dimensional space within a low-dimensional embedding. A representative algorithm is Isomap~\cite{JoshuaB2000}. 2) Local Structure Preservation: These methods focus solely on the relationships between a point and its immediate neighbors, striving to maintain this local neighborhood topology. Representative methods include LLE~\cite{SamT2000}, HLLE~\cite{DavidL2003}, LTSA~\cite{ZhangZhenyue2004}, and LEM~\cite{Belkin2003LaplacianEF}.

However, traditional manifold learning often employs "single-step" shallow mappings, making it difficult to capture complex nonlinear topological evolutions. Consequently, researchers have attempted to introduce Ricci Flow theory~\cite{FeiAndLi2025}, utilizing curvature-driven metric evolution to dynamically refine geometric structures. Although OCF excels in topological discovery, its computational complexity presents a severe scalability bottleneck for large-scale manifolds. The RCF introduced in this paper establishes a new paradigm for geometric evolution via effective resistance, substantially enhancing computational efficiency while retaining the topological sensitivity of Ricci Flow.

\subsection{Graph Structure Learning}

Graph Structure Learning (GSL) aims to address the absence of explicit graph structures in many real-world datasets (e.g., text or tabular data) or the poor quality of input graph structures~\cite{Zhu2021ASO}. GSL requires the model to learn the optimal topology dynamically or iteratively while simultaneously learning node feature representations. Existing GSL work can be broadly classified into two categories:

1) Metric-based Methods: Based on the network homophily assumption that similar nodes are more likely to be connected, these methods use differentiable kernel functions to compute similarity between node pairs as edge weights. They are simple to implement and easy to train end-to-end. Representative works include AGCN~\cite{PengAndLiu2021}, GRCN~\cite{YuAndZhang2020}, DGCNN~\cite{WangAndSun2019} and IDGL~\cite{ChenAndWu2020}.
2) Direct Methods: These methods treat the adjacency matrix itself as a set of free, learnable parameters optimized directly, often coupled with graph regularization constraints. Representative works include SLAPS~\cite{FatemiAndEl2021}, LDS-GNN~\cite{Franceschi2019LearningDS}.

In the GSL workflow, post-processing is critical to determining the final graph quality. Existing post-processing techniques (e.g., discrete sampling~\cite{JangGP17} or residual connections~\cite{LiAndWang2018}) largely fall under simple signal processing and lack deep constraints regarding geometric plausibility. The RCF optimizer proposed in this paper provides a novel geometric post-processing operator for GSL. Rather than simple edge filtering, it performs geometric rectification on structural predictions from a dynamical perspective, guided by a rigorous resistance curvature prior. RCF leverages curvature gradients to spontaneously identify and eliminate irrational "shortcut edges", ensuring the learned graph topology is more geometrically consistent with the underlying data manifold.

\section{Resistance Curvature Flow}
This chapter aims to establish a comprehensive theoretical framework and application paradigm for RCF. We begin with an in-depth analysis of the geometric essence of resistance curvature and its intrinsic connection to traditional Ricci Flow. Subsequently, we derive the dynamical evolution equations of RCF and demonstrate its compatibility and algorithmic implementation within deep learning environments. Furthermore, we propose a dynamic graph learning algorithm based on RCF and design several paradigms for embedding this algorithm into machine learning models.

\subsection{Theory of Resistance Curvature}
This section introduces the core foundations and concepts of resistance curvature, including effective resistance, vertex resistance curvature, and edge resistance curvature.
\subsubsection{Effective Resistance}
\textbf{The Essence of Effective Resistance Distance}: The effective resistance between nodes $i$ and $j$ possesses profound geometric significance. In essence, it serves as a measure of all-path connectivity. Within the context of differential geometry, it can be viewed as a discrete approximation of a "mean geodesic distance". Unlike traditional metrics that consider only the shortest path, effective resistance senses all possible paths between two points~\cite{ATIK201941, Ghosh2008}. The resulting redundancy-lowered resistance corresponds topologically to the presence of high-density local clusters.

\begin{center}
\begin{minipage}{\columnwidth}
\centering
\includegraphics[width=0.8\columnwidth]{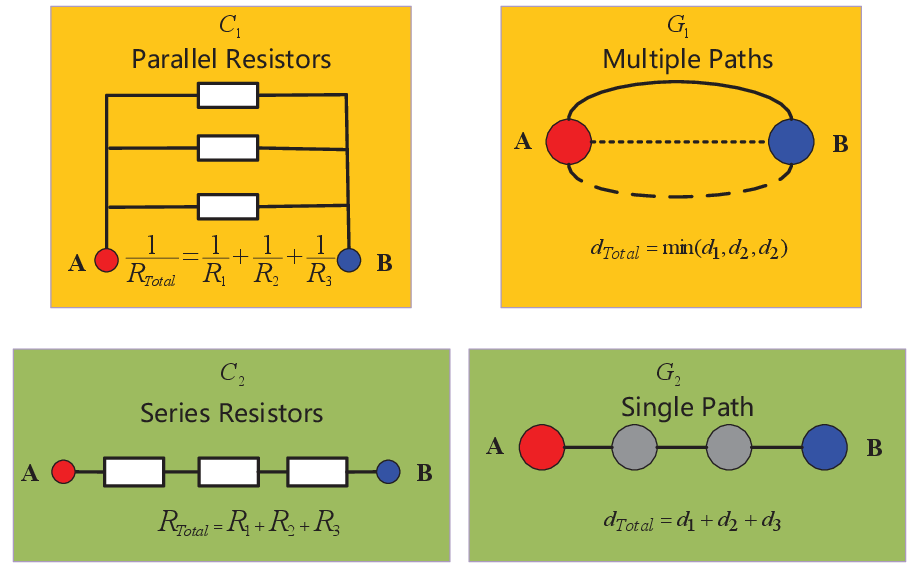}
\captionsetup{hypcap=false}
\captionof{figure}{The analogy between resistance in electrical circuits and paths in graphs.}
\label{fig:resistance_and_path}
\end{minipage}
\end{center}

Figure~\ref{fig:resistance_and_path} illustrates this intuitive analogy. In dense clusters ($C_1$ \& $G_1$), the abundance of parallel paths in a circuit significantly reduces the total effective resistance $R_{Total}$. Analogously, in a graph, multiple redundant paths between nodes indicate a tight local community, where a low effective resistance $R_{ij}$ signifies strong "intimacy" and high connectivity reliability. Conversely, in bridge structures ($C_2$ \& $G_2$), nodes connected only by series resistors exhibit increased resistance as path length grows. Similarly, graph nodes linked by a single path or situated at a "bottleneck" yield an exceptionally large $R_{ij}$, characterizing distant relationships and weak topological edges that are often susceptible to noise.

Mathematically, the effective resistance between nodes $i$ and $j$ can be efficiently computed using the Moore-Penrose pseudoinverse of the graph Laplacian matrix $\mathbf{L}$:
\begin{equation}
\label{eq:effective_resistance}
    R_{ij} = (\mathbf{e}_i - \mathbf{e}_j)^T \mathbf{L}^\dagger (\mathbf{e}_i - \mathbf{e}_j) = \mathbf{L}^\dagger_{ii} + \mathbf{L}^\dagger_{jj} - 2\mathbf{L}^\dagger_{ij},
\end{equation}
where $e_i$ denotes the unit vector. 

The conceptual foundation of effective resistance lies in interpreting the graph as a conductive network, where topological compactness is reflected in the "electrical flux" distribution. This approach is deeply embedded in spectral graph theory utilizing the graph Laplacian $\mathbf{L}$ to encode the intrinsic diffusion dynamics on the manifold. Examining charge transport resistance throughout the network yields a local structural descriptor that offers substantially stronger robustness compared to Euclidean distance alone.
\subsubsection{Resistance Curvature}
Based on the effective resistance curvature, \cite{Devriendt_2022} provides the definitions of resistance node curvature and resistance edge curvature as follows:

Resistance Node Curvature ($p_i$): Node curvature $p_i$ in the RCF framework measures the local geometric compactness around node $i$. It is defined based on the similarity weight $c_{ij}$ and the effective resistance $R_{ij}$ between a node and its neighbors. The formula for node curvature $p_i$ is:
\begin{equation}
     p_i = 1 - \frac{1}{2} \sum_{j \sim i} c_{ij} R_{ij},
     \label{eq:node_curvature}
\end{equation}
where $c_{ij}$ represents the similarity weight between nodes $i$ and $j$ (e.g., similarity based on feature distance), and $R_{ij}$ is the effective resistance between them. 

Resistance Edge Curvature ($k_{ij}$): Based on the node curvature $p_i$, RCF defines edge curvature $k_{ij}$, which reflects the geometric imbalance inherent in the edge connecting $i$ and $j$. Edge curvature $k_{ij}$ serves as the direct potential gradient driving the geometric flow. The formula for edge curvature $k_{ij}$ is:
\begin{equation}
 k_{ij} = \frac{2(p_i + p_j)}{R_{ij}}. 
 \label{eq:edge_curvature}
\end{equation}
Edge curvature $k_{ij}$ is determined by the sum of the node curvatures of its two endpoints, normalized by the effective resistance $R_{ij}$ between them. This curvature value quantifies the total local geometric information embodied by edge $(i, j)$ across its two adjacent nodes. Within the context of Ricci Flow, this curvature will be utilized to guide the adjustment of edge weights.

\subsection{Resistance Curvature Flow}
Resistance Curvature Flow (RCF) is a discrete dynamical implementation of the traditional continuous Ricci Flow within a resistance geometric metric space. It minimizes geometric energy by iteratively adjusting the metric properties of the graph structure.

Traditional Evolution Equation: The classical Ricci flow equation is $\frac{\partial g_{ij}}{\partial t} = -2k_{ij}$, and $g_{ij}$ denotes the Riemannian metric, which implies that the metric tensor evolves along the negative gradient direction of curvature. Borrowing this concept, we define the RCF evolution equation as:
\begin{equation}
 d_{ij}^{(t+1)} = d_{ij}^{(t)} - \eta \cdot d_{ij}^{(t)} \cdot k_{ij}^{(t)}, 
 \label{eq:rcf}
\end{equation}
where $d_{ij}^{(t)}$ is the distance metric between nodes $i$ and $j$ (typically inversely related to similarity $c_{ij}$), $\eta$ is a hyperparameter controlling the flow rate (learning rate), and $k_{ij}^{(t)}$ is the resistance edge curvature defined previously. This equation indicates that if the curvature $k_{ij}^{(t)}$ on edge $(i, j)$ is high (implying geometric imbalance or irrationality), the distance $d_{ij}^{(t)}$ will be contracted or expanded, driving the local geometry toward a more uniform state. This iterative process is the core mechanism by which RCF achieves graph structure geometric optimization.

Normalized Evolution Equation: To prevent global collapse or infinite expansion of the graph structure during evolution, we introduce a normalization constraint, resulting in the core RCF evolution formula:
\begin{equation}
     d_{ij}^{(t+1)} = d_{ij}^{(t)} - \eta \cdot d_{ij}^{(t)} \cdot \left(k_{ij}^{(t)} - \frac{1}{|E|
} \sum_{k,l}^{n} k_{kl}^{(t)} \right),
\label{eq:rcf_norm}
\end{equation}
where $\frac{1}{|E|} \sum_{k,l} k_{kl}^{(t)}$ represents the average curvature of all edges in the current graph. The normalization term acts like a "geometric thermostat." Its purpose is to maintain the stability of the overall graph scale while guiding the evolution via relative curvature deviations. If $k_{ij}^{(t)}$ is greater than the average curvature, the term in parentheses is positive, leading to a decrease in distance $d_{ij}^{(t)}$, meaning the compactness of that edge is reinforced. Conversely, if $k_{ij}^{(t)}$ is lower than the average, $d_{ij}^{(t)}$ increases, weakening the connection to rectify geometric imbalance. This design ensures that the evolutionary process is directed toward the goal of "curvature distribution homogenization."

\subsection{RCF-based Dynamic Graph Structure Learning Algorithm}
We propose algorithm RCF-based Dynamic Graph Structure Learning (DGSL-RCF) to demonstrate how RCF iteratively refines graph structures for downstream representation learning:

\begin{algorithm}[h]
	\caption{RCF-based Dynamic Graph Structure Learning}
	\label{alg1} 
	\begin{algorithmic}[1]
        \STATE \textbf{Input:} initial high dimensional data $\{x_1, x_2, \cdots, x_n\}$, \\
        neighbour size parameter $k$, \\
        neighbour set $N_i$ of $x_i$, \\
        Ricci flow iteration step $n\_iter$, \\
        learning rate $\eta$.
        \STATE Construct affinity graph $G=(V,E,w)$ using $k$-nearest neighbour method; 
        \STATE \textbf{if} $x_j \in N_i$ \textbf{then}
        \STATE \quad Initialize $w_{ij}^{(0)} \leftarrow exp^{-\frac{\|x_i-x_j\|^2}{t}}$, $d_{ij}^{(0)} \leftarrow 1 / w_{ij}^{(0)}$;
        \STATE \textbf{else} 
        \STATE \quad Set $w_{ij} \leftarrow 0$;
        \STATE \textbf{for} $i \leftarrow 1$ \textbf{to} $n\_iter$ \textbf{do}
         \STATE  \quad Construct perturbed Laplacian matrix $\tilde{\mathbf{L}}$ using Eq.~\eqref{eq:l_reverse}, $d_{ij}^{(t)} \leftarrow 1 / w_{ij}^{(t)}$;\\
         \STATE \quad  Approximate effective resistance $R_{ij}^{(t)}$ using Eq.~\eqref{eq:effective_resistance};\\
         \STATE \quad  Compute resistance curvature $k_{ij}^{(t)}$ using Eq.~\eqref{eq:edge_curvature};\\
         \STATE  \quad  Update $d_{ij}^{(t)}$ using normalized Ricci flow Eq.~\eqref{eq:rcf_norm};\\
         \STATE  \quad Set weight $w_{ij}^{(t)} \leftarrow 1 / d_{ij}^{(t)}$;
        \STATE Set the updated adjacency weights as $w^* \leftarrow w_{ij}^{(n\_iter)}$;
        \STATE Perform final normalization on $w^*$ to obtain output matrix $w^N$.
        \STATE \textbf{Output:} $w^N$.
	\end{algorithmic} 
\end{algorithm}

Similar to existing work~\cite{FeiAndLi2025}, DGSL-RCF introduces a dynamic evolution phase for the initial affinity graph. First, for a high-dimensional discrete dataset $\{x_1, x_2, \dots, x_n\}$ lacking an explicit graph structure, we employ the k-Nearest Neighbor (k-NN) method to construct an initial adjacency graph, denoted as $G = (V, E, w)$. During the discrete graph Ricci flow process, we treat the initial edge weight matrix $w$ as the starting metric. To ensure consistency between the adjacency matrix $w$ and the distance metric $d$, we take the reciprocal of all edge weights in $w$ to obtain the corresponding distance matrix $d$ (line 4). The evolution proceeds by utilizing the weight matrix $w$ to construct the perturbed Laplacian matrix $\tilde{\mathbf{L}}$, followed by the calculation of effective resistance $R$ and resistance curvature $k$ (line 8 to 10). We then apply the discretized normalized resistance curvature flow to $d$, allowing it to gradually evolve into a new distance matrix (line 11). We take the reciprocal of this evolved $d$ to serve as the new weight matrix $w$ (line 12). Once the geometric flow iterations are complete, we take the reciprocal of the resulting weight matrix to obtain $w^*$, which is then normalized to yield the final weight matrix $w^N$ (line 13 to 14).After evolving the adjacency matrix through this dynamic geometric flow, the newly obtained $w^N$ replaces the initial adjacency matrix $w$ for downstream tasks, such as clustering analysis or dimensionality reduction.

The algorithm can be integrated into machine learning models via three primary paradigms:
\begin{itemize}[topsep=0pt, parsep=0pt]
    \item Preprocessing Paradigm: The algorithm is applied at the input stage to purge topological noise, providing the subsequent model with more accurate geometric priors.
    \item Hidden Layer Regularization Paradigm: DGSL-RCF acts as a dynamic regularization operator. Geometric flow layers optimized by the algorithm are inserted between hidden layers to evolve the graph structure based on real-time features. The resulting curvature distribution is used as a loss term to constrain the topological evolution of the feature space.
    \item Output Refinement Paradigm: In DSL tasks, the algorithm is used for post-processing and refining the preliminary structures generated by the model. Global geometric constraints are utilized to ensure the physical plausibility of the output results.
\end{itemize}

Detailed descriptions of these three embedding methods can be found in~\ref{appendix:3Paradigms}.

\section{Theoretical Analysis}
This section analyzes the efficiency and effectiveness of the RCF and elaborates on its compatibility with deep learning models from the perspective of differentiability.
\subsection{Efficiency Analysis of RCF}
The primary bottleneck of resistance curvature is the calculation of effective resistance, which depends on the pseudoinverse of the graph Laplacian. Since $\mathbf{L}$ is positive semi-definite (PSD) and singular, its pseudoinverse is almost always a dense matrix. Direct computation has a complexity of $O(n^3)$, which is unacceptable for large-scale graphs.

To circumvent this, following our previous work~\cite{fei2025efficientcurvatureawaregraphnetwork}, we employ a diagonal perturbation method to transform the singular PSD matrix calculation into a linear system problem for a sparse positive definite matrix:
\begin{equation}
\label{eq:l_reverse}
    \tilde{\mathbf{L}} = \mathbf{L} + \epsilon \mathbf{I},
\end{equation}
where $\epsilon$ is a vanishingly small positive perturbation and $\mathbf{I}$ is the identity matrix. By adding this perturbation to the diagonal, $\tilde{\mathbf{L}}$ becomes a sparse symmetric positive definite (SPD) matrix. 
The calculation of $R_{ij}$ is reduced to:
\begin{equation}
R_{ij} = \mathbf{e}_{ij}^\top (\tilde{\mathbf{L}})^{-1} \mathbf{e}_{ij},
\end{equation}
where $\mathbf{e}_{ij}$ is the indicator vector.

For SPD matrices, the pseudoinverse and the inverse are approximately equivalent. Solving a linear system for an SPD matrix is significantly more efficient than computing the pseudoinverse of a singular matrix. By utilizing the Sparse Cholesky decomposition or iterative solvers (e.g., Conjugate Gradient), the complexity is reduced from $O(n^3)$ to nearly $O(mn^2)$ per iteration, $m$ is maximum degree of a node in the graph. This allows RCF to handle large-scale datasets while remaining over 100x faster than OT-based methods.

\subsection{Effectiveness Analysis of RCF}
The effectiveness of RCF stems from its unique structural self-healing capability, reflected in two aspects:

\textbf{Manifold Enhancement}: Within a manifold structure, points in a dense cluster are connected by multiple redundant paths. According to the properties of effective resistance $R_{ij}$, for a pair of nodes $(i, j)$ in a highly cohesive region, the presence of $k$ parallel paths implies $R_{ij} \approx (\sum_{p=1}^k r_p^{-1})^{-1}$, $r_p$ is the resistance of path $p$. As $k$ increases, $R_{ij} \to 0$.

From the definition of resistive edge curvature $k_{ij} = \frac{2(p_i + p_j)}{R_{ij}}$, as $R_{ij}$ decreases, the curvature $k_{ij}$ tends to a large positive value ($k_{ij} \gg \bar{k}$). Substituting this into the normalized evolution equation:
\begin{equation}
d_{ij}^{(t+1)}=d_{ij}^{(t)}-\eta\cdot d_{ij}^{(t)}\cdot\underbrace{\left(k_{i j}^{(t)}-\bar{k}\right)}_{>0}. 
\end{equation}
Since the term $(k_{ij} - \bar{k})$ is positive, the distance $d_{ij}$ undergoes exponential-like contraction. Mathematically, this minimizes the Dirichlet energy of the graph:
\begin{equation}
\mathcal{E}(G) = \sum_{(i,j) \in E} R_{ij} d_{ij}^2.
\end{equation}
By forcing $d_{ij} \to 0$ for high-curvature edges, RCF eliminates microscopic gaps in the feature distribution and strengthens the manifold's local cohesion.

\textbf{Noise Suppression}: Sampling noise often introduces "shortcuts" or "cross-cluster edges". While these edges may appear short in Euclidean space, they lack the path redundancy typical of the true manifold structure. For a noise edge $e_{noise}$, the effective resistance $R_{ij}^{noise} \approx d_{ij}$ (similar to a series circuit). Consequently, the absence of local connectivity around the endpoints leads to low node curvatures $p_i, p_j$, yielding $k_{ij}^{noise} < \bar{k}$. The corresponding evolution equation then takes the form:
\begin{equation}
d_{ij}^{(t+1)} = d_{ij}^{(t)} + \eta \cdot d_{ij}^{(t)} \cdot \underbrace{\left(\bar{k} - k_{ij}^{(t)}\right)}_{> 0}.
\end{equation}
This induces a repulsive force that expands the distance $d_{ij}$, effectively "pushing" noise edges out of the manifold structure. 

\textbf{Differentiable Compatibility}: For integration into deep learning frameworks (e.g., PyTorch), RCF must support backpropagation. The gradient of the RCF output with respect to the input weights $\mathbf{W}$ can be derived using the chain rule on the matrix inverse:
\begin{equation}\frac{\partial (\tilde{\mathbf{L}})^{-1}}{\partial w_{ij}} = -(\tilde{\mathbf{L}})^{-1} \frac{\partial \tilde{\mathbf{L}}}{\partial w_{ij}} (\tilde{\mathbf{L}})^{-1}.
\end{equation}
Since all operations in RCF, including the diagonal perturbation, matrix inversion, and the normalized evolution step, are composed of differentiable elementary matrix operations, the entire RCF layer is end-to-end differentiable. This allows the curvature distribution to act as a topological loss function, directly guiding the feature extraction process of the neural network.

\subsection{Connection to Ollivier-Ricci Curvature}
Ollivier-Ricci Curvature (ORC) is based on Optimal Transport theory (Wasserstein distance), measuring the ease of moving probability measures. Resistance curvature is physically consistent with ORC. In dense regions, the resistance to mass migration (or current transport) is minimal, corresponding to high positive curvature.At sparse bridges, migration resistance is maximal, corresponding to negative curvature.Research suggests that under specific regularization conditions, the gradient evolution of effective resistance effectively approximates the corrective effect of the Ricci curvature tensor on the metric tensor. 

\section{Experiments}
This chapter aims to comprehensively evaluate the performance of the DGSL-RCF across three key tasks (ML, DML, and GSL) through a systematic experimental design. The experiments are designed around the following core objectives:
\begin{itemize}[topsep=0pt, parsep=0pt]
    \item Effectiveness Validation. To assess the effectiveness of DGSL-RCF in enhancing the geometric representation capabilities of models.
    \item Efficiency Comparison. To conduct a systematic comparison with the optimizer based on OCF, quantifying the computational efficiency advantage of RCF.
    \item Sensitivity Analysis. To investigate the impact of key hyperparameters (e.g., neighborhood size $k$, iteration count $n\_iter$, and learning rate $\eta$) on algorithm performance, thereby validating its robustness.
\end{itemize}

\subsection{Experimental Setup}
\subsubsection{Datasets}
\label{sec_dataset}
To verify the generalizability of the RCF framework, we selected representative datasets across three major geometric representation learning tasks:
\begin{itemize}[topsep=0pt, parsep=0pt]
    \item DML: Standard benchmark datasets (CUB-200-2011~\cite{WahCUB_200_2011}, Cars-196~\cite{Krause2013}, and Stanford Online Products (SOP)~\cite{Song2015DeepML}) from the fine-grained image recognition domain are selected, following the commonly used dataset splitting protocols in the field, specifically adhering to the protocol in ~\cite{ZhangandLi2020NIPS}.
    \item ML: Synthetic manifold datasets (Swiss Roll, S-Curve, Truncated Sphere, Gaussian Surface) are used to visually demonstrate the algorithm's geometric repair capability in a controlled environment. Real image datasets (MNIST, USPS, Medical MNIST, KVASIR~\cite{KVASIR2017}) are utilized to verify its applicability in practical scenarios.
    \item GSL. For data lacking explicit topological structures, classic classification datasets such as Wine, Cancer, and Digits are chosen. Additionally, the large-scale text dataset 20News is used to evaluate the algorithm's scalability in large-scale scenarios.
\end{itemize}
Detailed statistics for each dataset is provided in Table~\ref{tab:dataset_stat}, with further descriptions and their download urls provided in~\ref{appendix:dataset_details}.


\begin{table*}[!htbp]
\centering
\caption{Details of datasets for three tasks.} 
\label{tab:dataset_stat}
\footnotesize
\begin{tabular}{llccc} 
\hline
Task & Dataset & \#instances & \#classes & \#features \\
\hline
\multirow{3}{*}{Deep Metric Learning} & CUB-200-2011 & 11788 & 200 & $224\times 224\times 3$ \\
& Car-196 & 16185 & 98 & $224\times 224\times 3$ \\
& SOP & 120053 & 22634 & $224\times 224\times 3$ \\
\hline
\multirow{4}{*}{Manifold learning} & USPS & 9298 & 10 & 256 \\
& MNIST & 10000 & 10 & 784 \\
& Medical MNIST & 6000 & 18 & 4096 \\
& KVASIR & 4000 & 8 & 4096 \\
\hline
\multirow{4}{*}{Graph Structure Learning} & Wine & 178 & 3 & 13 \\
& Cancer & 569 & 2 & 30 \\
& Digits & 1797 & 10 & 64 \\
& 20News & 9607 & 10 & 236 \\
\hline
\end{tabular}
\end{table*}

\subsubsection{Baselines}
To facilitate fair and in-depth comparisons, targeted baseline methods are established for each task.
\begin{itemize}[topsep=0pt, parsep=0pt]
    \item DML: Following the setup in ~\cite{ZhangandLi2020NIPS}, BN-Inception~\cite{IoffeandSergey2015} is used as the network backbone. Four representative loss functions (Triplet~\cite{SchroffAndKalenichenko2015}, Semi-Triplet~\cite{SchroffAndKalenichenko2015}, N-Pairs~\cite{SohnKihyuk2016}, and Multi-Similarity~\cite{WangAndHan2019} Loss) are chosen as baselines. Within each base model, a Geometric Flow Layer (GFL) is integrated. Performance is compared between embedding a static kNN graph (+Static) and embedding a graph dynamically optimized by DGSL-RCF (+DGSL-RCF) to validate the gains from dynamic geometric regularization. An introduction to the GFL is provided in~\ref{appendix:DML_GFL}.
    \item ML: Classical manifold learning algorithms (LLE~\cite{SamT2000}, HLLE~\cite{DavidL2003}, LTSA~\cite{ZhangZhenyue2004}, and LEM~\cite{Belkin2003LaplacianEF}) are selected as foundational comparisons. To precisely evaluate the contribution of the geometric flow itself, an ablation study is set up using the LEM algorithm as the backbone model. This involves comparing performance differences when using the original graph structure, the graph structure optimized by DGSL-OCF (based on Ollivier-Ricci Curvature Flow), and the graph structure optimized by DGSL-RCF as input.
    \item GSL: 
    Mainstream methods covering static graph construction (e.g., kNN-GCN~\cite{FatemiAndEl2021}) and differentiable graph learning (e.g., LDS~\cite{Franceschi2019LearningDS}, IDGL~\cite{ChenAndWu2020}, SLAPS~\cite{FatemiAndEl2021}) are selected as baselines. Using SLAPS as the backbone network, a DGSL-RCF module is cascaded at its output end as a post-processing refiner. Performance is compared before and after structural refinement.
\end{itemize}

\subsubsection{Evaluation Metrics}
To evaluate algorithm performance multidimensionally, well-recognized metrics in the field are adopted for different tasks.
\begin{itemize}[topsep=0pt, parsep=0pt]
    \item DML: Standard clustering metrics for the task are used, including Normalized Mutual Information (NMI), F1-Score (F1), and Recall.
    \item ML: The quality of manifold unfolding is assessed through the clustering quality in the low-dimensional embedding space, using NMI, Accuracy (ACC), Adjusted Rand Index (ARI), and F1.
    \item GSL: The effectiveness of the learned graph structure is evaluated by the ACC and its Standard Deviation (Std) in downstream node classification tasks.
\end{itemize}

\subsubsection{Implementation Details}
All experiments are implemented based on the Python 3.9 and PyTorch 1.12.0 frameworks, and are completed on a high-performance server configured with an Intel(R) Xeon(R) Gold 6248R CPU, NVIDIA RTX 3090 GPU (24GB VRAM), and 512GB RAM. 
For curvature calculation, the Ollivier-Ricci curvature is computed using the Python library\footnote{\url{https://github.com/saibalmars/GraphRicciCurvature}} provided by Ni et al.
~\cite{Ni2019CommunityDO}. The calculation of Resistance Curvature\footnote{\url{https://github.com/cqfei/resistance_curvature}} is adapted from the open-source implementation in~\cite{fei2025efficientcurvatureawaregraphnetwork}.

The key hyperparameters in the DGSL-RCF algorithm, including $\eta$, $n\_iter$, and $k$, are determined through grid search. The code for this project is open-sourced on GitHub (\url{https://github.com/cqfei/RCF}).

\subsection{Baseline Comparisons}
\subsubsection{Deep Metric Learning}
In this experiment, the model's training method and hyperparameter settings follow the protocol in ~\cite{ZhangandLi2020NIPS}. The geometric flow layer is inserted between the convolutional and activation layers of the deep neural network. Experimental hyperparameter settings are detailed in Appendix Table~\ref{tab:dml_parameter}.

The results, shown in Table~\ref{tab:dml_cmp}, indicate that DGSL-RCF brings significant improvements across all datasets and loss functions.

For instance, on the CUB-200-2011 dataset using Triplet Loss, the +DGSL-RCF method improved NMI, F1, and Recall by 44.61\%, 131.53\%, and 56.61\%, respectively, compared to the baseline model. Compared to the static graph method (+Static), it further improved NMI, F1, and Recall by 1.73, 12.77, and 1.34 points, demonstrating robust gains and validating its ability to enhance inter-class discriminability by dynamically optimizing the geometric distribution of the feature space.

Furthermore, on the large-scale long-tailed dataset SOP, DGSL-RCF exhibited strong scalability. Using Triplet Loss, compared to the base method, the F1 score surged from 30.26\% to 56.59\%, an increase of over 87\%, and was 14.65\% higher than the +Static method. This confirms the universality of DGSL-RCF's structural enhancement capability in complex semantic spaces.

\begin{table*}
	\caption{\small{Experimental results of deep metric learning on CUB-200-2011 and Cars196 datasets. NMI, F1, and Recall are reported (\%).}}
	\label{tab:dml_cmp}
    \footnotesize
	\setlength\tabcolsep{6pt} 
	\begin{center}
				\begin{tabular}{lcccccccccccc}
					\toprule
					\multirow{2}{2cm}{method} & \multicolumn{3}{c}{CUB-200-2011} & \multicolumn{3}{c}{Cars-196} & \multicolumn{3}{c}{SOP} \\\cmidrule(lr){2-4}\cmidrule(lr){5-7}\cmidrule(lr){8-10} 
					& NMI & F1 & Recall & NMI & F1 & Recall & NMI & F1 & Recall	\\
					\midrule
        				CGML(Triplet) & 60.20 & 27.00 & 53.30 &63.50 & 32.90 & 75.80 & 87.10 & 23.20 &  64.10\\
					 CGML(N-pair) &  60.40 &  28.50 & 52.10  & 62.60 &  31.00 & 75.80 & 88.10 & 27.00 & 68.40\\
    					 IBCDML & 74.00 & - &  70.30  & 74.80 & - &  88.10 & 92.60 & -& 81.40\\
					\midrule
					Triplet   & 59.34 & 23.12 & 52.98 & 56.03 & 24.94 & 61.09 & 88.72 & 30.26 & 63.19 \\
					+Static & 84.35 & 47.47 & 81.87 & 80.34 & 40.13 & 87.82 & 93.12 & 49.36 & 77.06 \\
					\textbf{+DGSL-RCF}    &  \textbf{85.81} & \textbf{53.53} & \textbf{82.97} & \textbf{81.83} & \textbf{46.73} &\textbf{ 89.28} & \textbf{94.33} & \textbf{56.59} & \textbf{77.71} \\
					\midrule
					Semi-Triplet  & 70.36 & 41.02 & 65.89 & 68.58 & 37.47 & 81.08 & 91.21 & 42.03 & 74.97 \\
					+Static  & 84.33 & 48.12 & 82.33 & \textbf{78.95} & 37.60 & 87.66 & 92.25 & 44.90 & 76.34 \\
					\textbf{+DGSL-RCF}  & \textbf{85.42} & \textbf{52.51} & \textbf{82.71} & 78.68 & \textbf{40.80} & \textbf{88.82} & \textbf{93.56} & \textbf{53.09} & \textbf{78.97}\\
					\midrule
					N-Pairs & 70.48 & 40.67 & 62.06 & 68.43 & 38.14 & 79.09 & 91.04 & 41.57 & 74.69\\
					+Static  & 84.54 & 49.50 & 80.27 & 81.28 & 44.17 & 85.97 & 92.43 & 45.91 & 75.18\\
					\textbf{+DGSL-RCF} & \textbf{85.06} & \textbf{50.53} & \textbf{81.36} & \textbf{81.58} & \textbf{44.77} & \textbf{86.81} & \textbf{93.01} & \textbf{49.40} & \textbf{76.33}\\
					\midrule
					Multi-Similarity  & 70.63 & 41.17 & 66.58 & 70.87 & 42.20 & 84.43 & 91.35 & 43.67 & 76.56\\
					+Static   & \textbf{84.23} & \textbf{47.86} & 81.06 & \textbf{80.77} & \textbf{42.96} & 86.42 & 92.68 & 46.90 & 77.53 \\
					\textbf{+DGSL-RCF} & 82.81 & 46.99 & \textbf{82.11} & 80.13 & 41.58 & \textbf{86.79} & \textbf{93.63} & \textbf{53.24} & \textbf{79.94}\\
					\bottomrule
				\end{tabular}
	\end{center}
\end{table*}

\subsubsection{Manifold Learning}
This set of experiments covers both synthetic manifold datasets and real image datasets. The hyperparameter settings for DGSL-OCF and DGSL-RCF are detailed in Appendix Table~\ref{tab:ml_parameters}.

Results in Table~\ref{tab:ml_cmp} show that introducing curvature flow optimization (either DGSL-OCF or DGSL-RCF) significantly outperforms the original baseline algorithms on almost all datasets. On the Medical MNIST dataset, combining LEM with DGSL-RCF increased the ACC metric by approximately 24.28\%, fully demonstrating its powerful capability to repair the initial topological structure of complex manifolds. On synthetic datasets like S-Curve and Gaussian Surface, although the base LEM model has inherent limitations, its performance was enhanced after integration with the curvature flow. Aggregating metrics across all datasets, RCF outperformed OCF in 62.5\% of the comparisons, indicating that DGSL-RCF not only matches OCF's ability to capture intrinsic manifold geometric features but also shows better adaptability in most tasks.
\begin{table*}[htbp]
\centering
\caption{\small{Results for real-word and synthetic datasets on ML task (\%).}}
\label{tab:ml_cmp}
\footnotesize
\begin{tabular}{@{}lllccccccc@{}}
\toprule
\textbf{\shortstack{Dataset\\Type}} & \textbf{Dataset} & \textbf{Metric} & \multicolumn{6}{c}{\textbf{Method}} \\
\cmidrule(l){4-9}
& & & LLE & HLLE & LTSA & LEM & \textbf{\shortstack{LEM+\\DGSL-OCF}} & \textbf{\shortstack{LEM+\\DGSL-RCF}} \\
\midrule

\multirow{16}{*}{\rotatebox[origin=c]{90}{Real-world Dataset}} & 
\multirow{4}{*}{USPS} & NMI & 58.32 & 35.61 & 36.08 & 74.80 & \textbf{75.57} & 74.91 \\
& & ACC & 42.13 & 18.54 & 21.73 & 53.35 & \textbf{63.91} & 53.46 \\
& & ARI & 42.70 & 15.29 & 19.32 & 54.52 & \textbf{61.23} & 54.61 \\
& & F1 & 50.02 & 30.05 & 32.16 & 60.02 & \textbf{65.50} & 60.10 \\
\addlinespace[0.2em]
\cline{4-9}
\addlinespace[0.2em]

& 
\multirow{4}{*}{MNIST} & NMI & 55.43 & 0.89 & 0.97 & 58.53 & 58.58 & \textbf{59.05} \\
& & ACC & 37.38 & 9.88 & 9.99 & 41.50 & 43.00 &\textbf{ 43.47} \\
& & ARI & 35.77 & 0.01 & 0.01 & 39.58 & 41.66 &\textbf{ 42.58} \\
& & F1 & 43.19 & 18.09 & 18.10 & 46.32 & 48.19 & \textbf{49.06} \\
\addlinespace[0.2em]
\cline{4-9}
\addlinespace[0.2em]

& 
\multirow{4}{*}{\shortstack{Medical\\MNIST}} & NMI & 82.43 & 0.25 & 0.22 & 77.33 & \textbf{87.75} & 84.43 \\
& & ACC & 67.37 & 16.65 & 16.65 & 56.21 & \textbf{73.15} & 69.97 \\
& & ARI & 70.49 & 0.00 & 0.00 & 59.05 & \textbf{77.32} & 73.78 \\
& & F1 & 76.03 & 28.53 & 28.54 & 67.20 & \textbf{81.53} & 78.69 \\
\addlinespace[0.2em]
\cline{4-9}
\addlinespace[0.2em]

& 
\multirow{4}{*}{KVASIR} & NMI & 33.13 & 1.27 & 0.83 & 34.66 & 34.77 & \textbf{38.54} \\
& & ACC & 27.21 & 12.48 & 12.48 & 27.02 & \textbf{27.29} & 22.25 \\
& & ARI & 20.65 & 0.01 & 0.00 & 18.35 & \textbf{18.64} & 17.25 \\
& & F1 & 32.47 & 22.15 & 22.16 & 29.45 & 29.68 &\textbf{ 32.19} \\
\midrule

\multirow{12}{*}{\rotatebox[origin=c]{90}{Synthetic Dataset}} & 
\multirow{4}{*}{S Curve} & NMI & 41.32 & 61.30 & \textbf{61.30} & 44.55 & 45.20 & 45.15 \\
& & ACC & 68.28 & 92.46 & \textbf{92.46} & 73.04 & 74.09 & 73.84 \\
& & ARI & 27.58 & 59.63 & \textbf{59.63} & 32.74 & 33.71 & 33.81 \\
& & F1 & 58.52 & 76.22 & \textbf{76.22} & 60.52 & 60.85 & 61.12 \\
\addlinespace[0.2em]
\cline{4-9}
\addlinespace[0.2em]

& 
\multirow{4}{*}{\shortstack{Swiss\\Roll}} & NMI & 56.69 & 31.83 & 31.83 & 72.05 & 73.75 & \textbf{78.22} \\
& & ACC & 68.53 & 41.77 & 41.77 & 78.38 & 80.91 & \textbf{85.40} \\
& & ARI & 50.99 & 19.41 & 19.41 & 64.10 & 67.01 & \textbf{73.70} \\
& & F1 & 62.08 & 37.46 & 37.46 & 72.38 & 74.61 & \textbf{79.84} \\
\addlinespace[0.2em]
\cline{4-9}
\addlinespace[0.1em]

& 
\multirow{4}{*}{\shortstack{Truncated\\Sphere}} & NMI & 43.34 & 0.38 & 0.38 & 43.11 & 43.11 & \textbf{44.04} \\
& & ACC & 61.86 & 33.31 & 33.31 & 61.33 & 61.33 &\textbf{ 62.38} \\
& & ARI & 43.76 & -0.02 & -0.02 & 42.55 & 42.55 & \textbf{44.46} \\
& & F1 & 62.78 & 49.92 & 49.92 & 61.86 & 61.86 & \textbf{63.21} \\
\addlinespace[0.2em]
\cline{4-9}
\addlinespace[0.2em]
& 
\multirow{4}{*}{\shortstack{Gaussian\\Surface}} & NMI & 45.99 & 46.56 & \textbf{46.56} & 43.48 & 44.63 & 44.53 \\
& & ACC & 49.99 & 50.25 & \textbf{50.25} & 48.76 & 49.42 & 49.36 \\
& & ARI & 32.13 & 32.46 & \textbf{32.46} & 30.38 & 31.21 & 31.14 \\
& & F1 & 49.39 & 49.62 & \textbf{49.62} & 48.02 & 48.63 & 48.58 \\

\bottomrule
\end{tabular}
\end{table*}

\subsubsection{Graph Structure Learning}
The hyperparameter settings for this part of the experiment are provided in~\ref{appendix:parameter_setting_ml_dml}.

\begin{table}[htbp]
\centering
\caption{GSL Results on classification datasets without explict graph, the results of the baseline models in this table are taken from~\cite{FatemiAndEl2021} (\%). }
\label{tab:gsl_performance_comparison}
\footnotesize
\begin{tabular}{@{}lcccc@{}}
\toprule
\textbf{Model} & \textbf{Wine} & \textbf{Cancer} & \textbf{Digits} & \textbf{20News} \\
\midrule
MLP            & 96.1 $\pm$ 1.0 & 95.3 $\pm$ 0.9 & 81.9 $\pm$ 1.0 & 30.4 $\pm$ 0.1 \\
kNN-GCN        & 93.5 $\pm$ 0.7 & 95.3 $\pm$ 0.4 & 95.4 $\pm$ 0.4 & 46.3 $\pm$ 0.3 \\
LDS            & 97.3 $\pm$ 0.4 & 94.4 $\pm$ 1.9 & 92.5 $\pm$ 0.7 & 46.4 $\pm$ 1.6 \\
IDGL           & 97.0 $\pm$ 0.7 & 94.2 $\pm$ 2.3 & 92.5 $\pm$ 1.3 & 48.5 $\pm$ 0.6 \\
SLAPS   & 96.5 $\pm$ 0.8 & 96.6 $\pm$ 0.2 & 94.2 $\pm$ 0.1 & 49.8 $\pm$ 0.9 \\
\textbf{SLAPS+DGSL-RCF} & \textbf{98.2 $\pm$ 0.1} & \textbf{97.5 $\pm$ 0.3} & \textbf{97.4 $\pm$ 0.4} & \textbf{50.8 $\pm$ 0.7}\\
\bottomrule
\end{tabular}
\end{table}
Results in Table~\ref{tab:gsl_performance_comparison} show that the method integrating the DGSL-RCF module outperformed the original SLAPS model on almost all datasets. For example, classification accuracy improved by 1.76\%, 0.93\%, 3.40\%, and 2.01\% on the Wine, Cancer, Digits, and 20News datasets, respectively. This indicates that even for learned latent graph structures, introducing the Resistance Curvature Flow for post-processing refinement can effectively eliminate noisy edges and enhance geometric consistency, thereby improving the robustness of downstream tasks.

\subsection{Efficiency Comparison: RCF vs. OCF}
To quantitatively evaluate the computational efficiency of RCF, we conducted comparative experiments with OCF on four synthetic datasets. The experiments measured the computation time required for both algorithms to complete 5 iterations ($n\_
iter=5$) under varying data scales (1,000 to 10,000 sample points) and different neighborhood sizes $k$ ($k \in \{10, 30, 50\}$). It is noteworthy that OCF runs in a CPU parallel environment (96 threads), while RCF uses only a single GPU for computation. The results are shown in Figure~\ref{fig:ml_cost}, with detailed statistics provided in Appendix Tables~\ref{tab:s_curve_swiss_roll_cost} and~\ref{tab:sphere_gaussian_cost}.

The results indicate:
\begin{itemize}[topsep=0pt, parsep=0pt]
    \item Under the same data scale, the computation time of RCF is significantly lower than OCF. For example, on the S-Curve dataset with k=50, RCF achieved a speed-up ratio of up to 283.61 times.
    \item As the sample size increases, the time cost of OCF exhibits polynomial growth and is extremely sensitive to graph connectivity (the value of $k$). In contrast, the computational efficiency of RCF shows high robustness to changes in $k$, as its core is based on matrix operations, with overhead primarily determined by the number of nodes rather than graph density.
\end{itemize}

\begin{center}
\begin{minipage}{\columnwidth}
\centering
	\includegraphics[width=0.24\columnwidth]{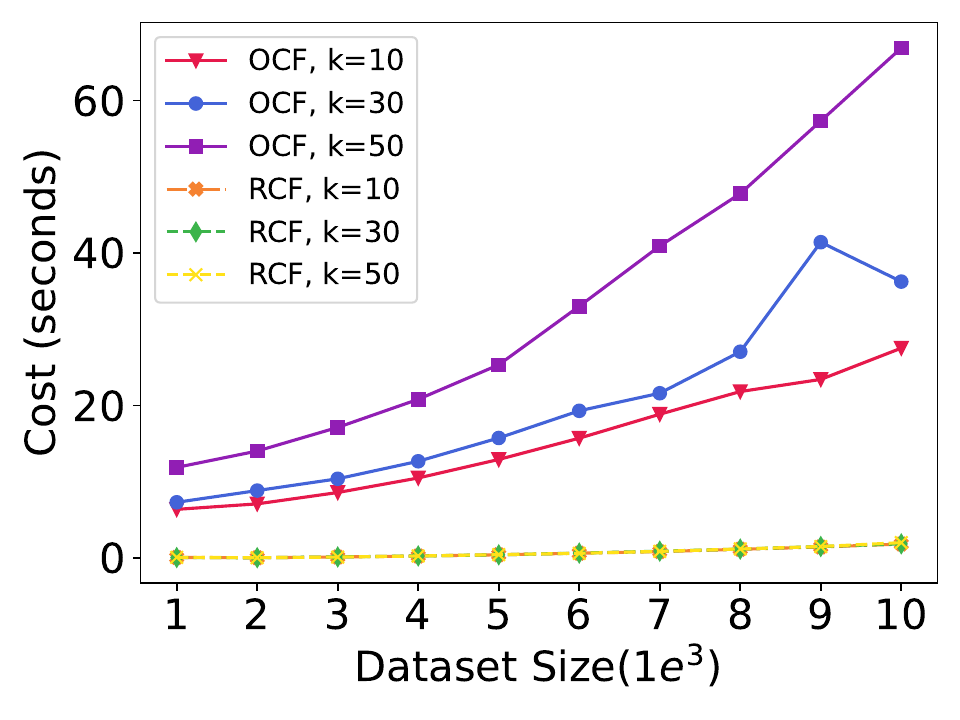}
    \includegraphics[width=0.24\columnwidth]{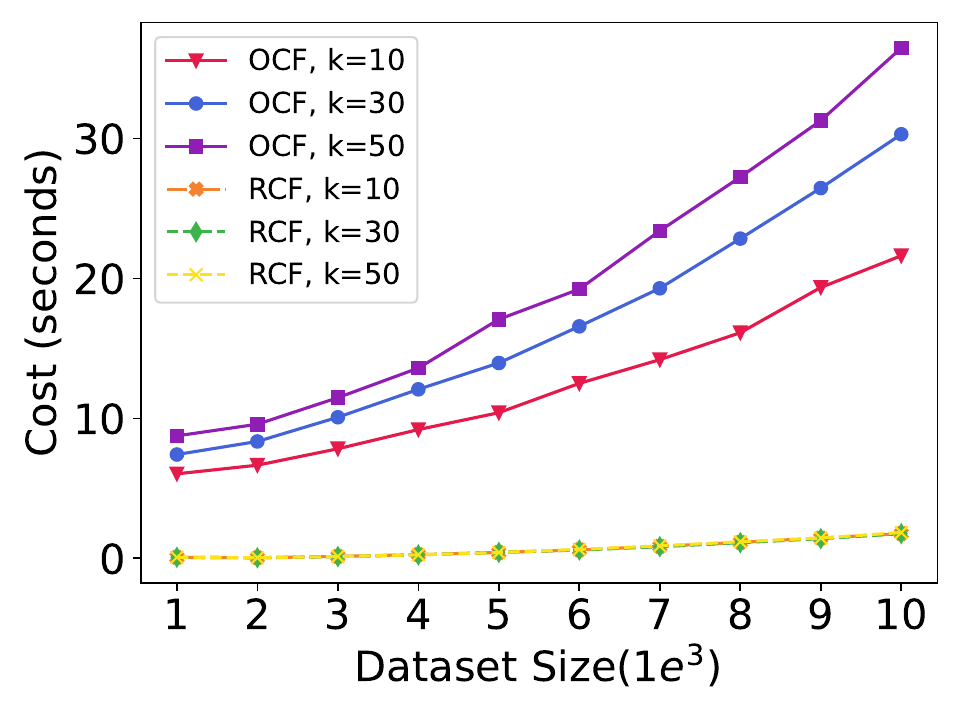}
    \includegraphics[width=0.24\columnwidth]{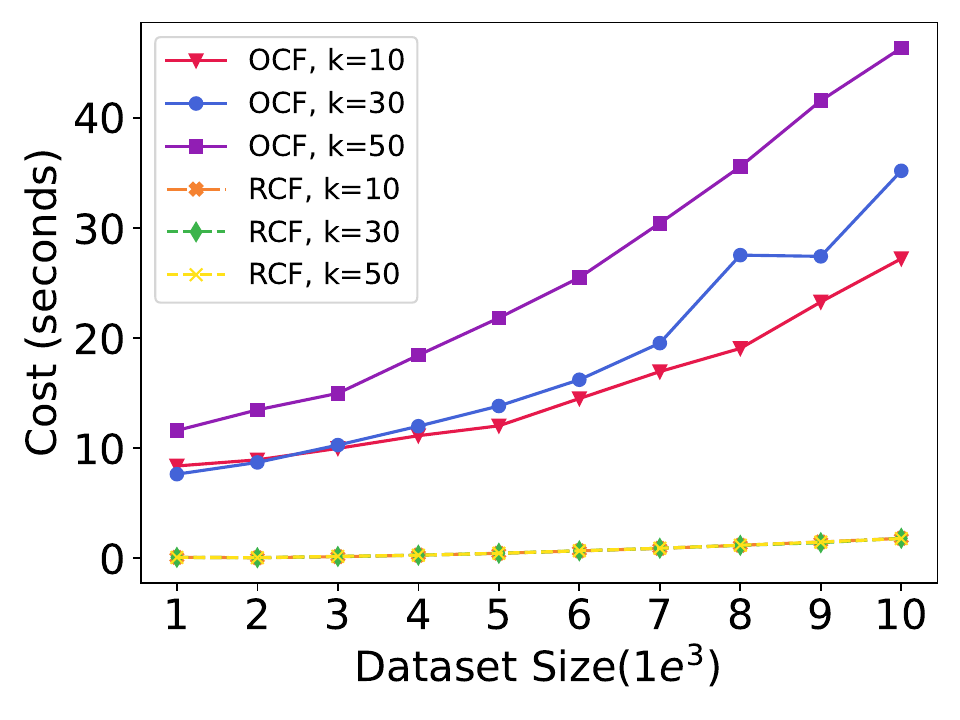}
    \includegraphics[width=0.24\columnwidth]{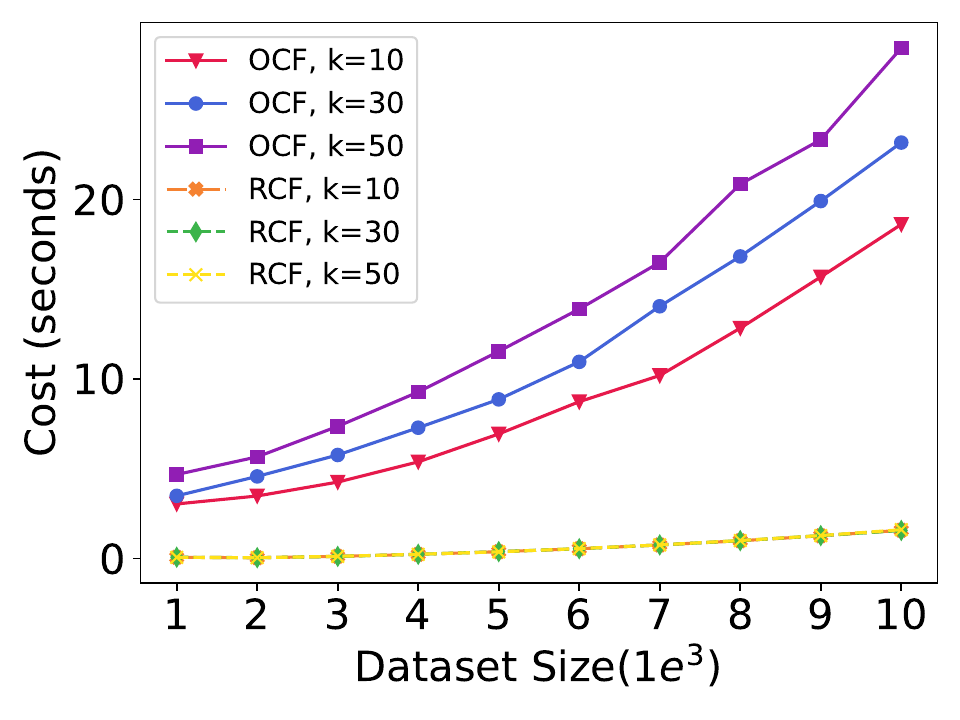}
\captionsetup{hypcap=false}
\captionof{figure}{\small{Cost comparison for RCF and OCF under different $k$ (top-left: S Curve, top-right: Swiss Roll, bottom-left: Truncated Sphere, bottom-right: Gaussian Surface).}}
\label{fig:ml_cost}
\end{minipage}
\end{center}

In summary, while maintaining excellent geometric representation capability, RCF's computational efficiency far surpasses that of the traditional OCF, demonstrating its potential for handling large-scale graph data.

\subsection{Convergence Analysis}
We take the DML task as the subject of study to compare the convergence trajectories of models incorporating DGSL-RCF against the baseline models (Figure~\ref{fig:dml_convergence}).
Experiments were conducted on the CUB-200-2011 and Cars-196 datasets using Triplet Loss as the loss function, with the maximum number of iterations set to 8,000. We recorded the change curves of key metrics (NMI, F1, Recall) with respect to the number of training epochs in real-time, as shown in Figure~\ref{fig:dml_convergence}. 

We observe two main benefits of the DGSL-RCF algorithm:
\begin{itemize}[topsep=0pt, parsep=0pt]
    \item Rapid Early Stabilization. As shown in Fig.~\ref{fig:dml_convergence}, the +DGSL-RCF method exhibits a distinct and efficient optimization trajectory. Metrics evaluating clustering purity and overall consistency (NMI, F1) converge and remain stable within the initial $<$100 epochs, while the Recall metric which reflecting comprehensive sample coverage, steadily improves thereafter, reaching full convergence at around 1000 epochs (taking the Cars‑196 dataset as an example). In stark contrast, all baseline methods show a slow and synchronized increase across all three metrics, requiring approximately 3000 epochs to converge. This indicates that DGSL‑RCF can quickly capture and solidify the primary clustering structures of the data.
    \item Adaptive Efficiency Across Datasets. This advantage is further validated on datasets with different levels of complexity. On the more structurally well‑defined Cars‑196 dataset, the Recall of +DGSL‑RCF converges in only 1000 epochs—twice as fast as the baseline methods. On the more challenging CUB‑200‑2011 dataset, although the final convergence epoch for Recall is similar (around 2000 epochs), our method consistently outperforms others at every intermediate training stage. This demonstrates that our optimization follows a consistently superior path, yielding better models under the same computational cost.
\end{itemize}

\begin{center}
\begin{minipage}{\columnwidth}
\centering
\includegraphics[width=0.24\columnwidth]{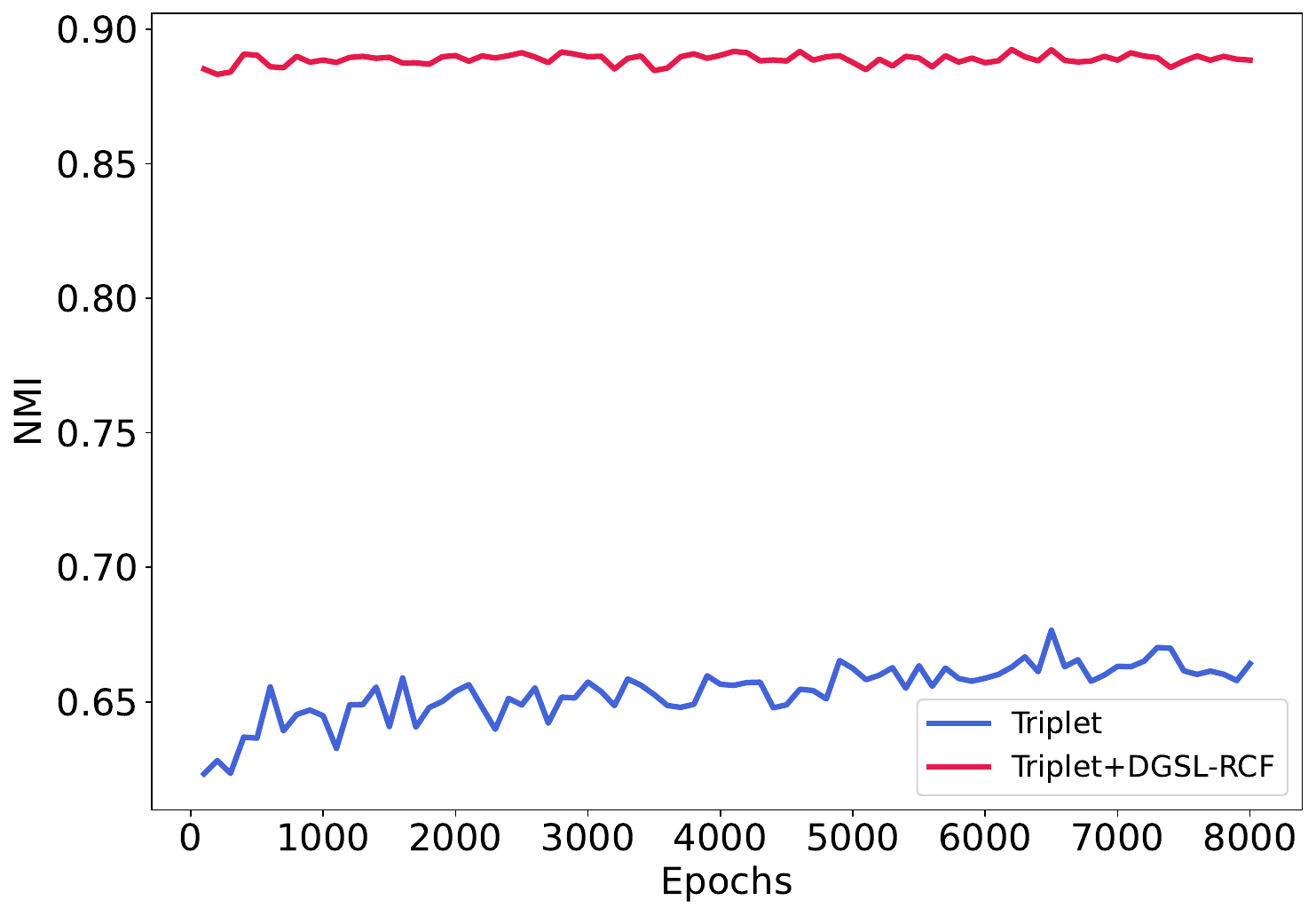}
\includegraphics[width=0.24\columnwidth]{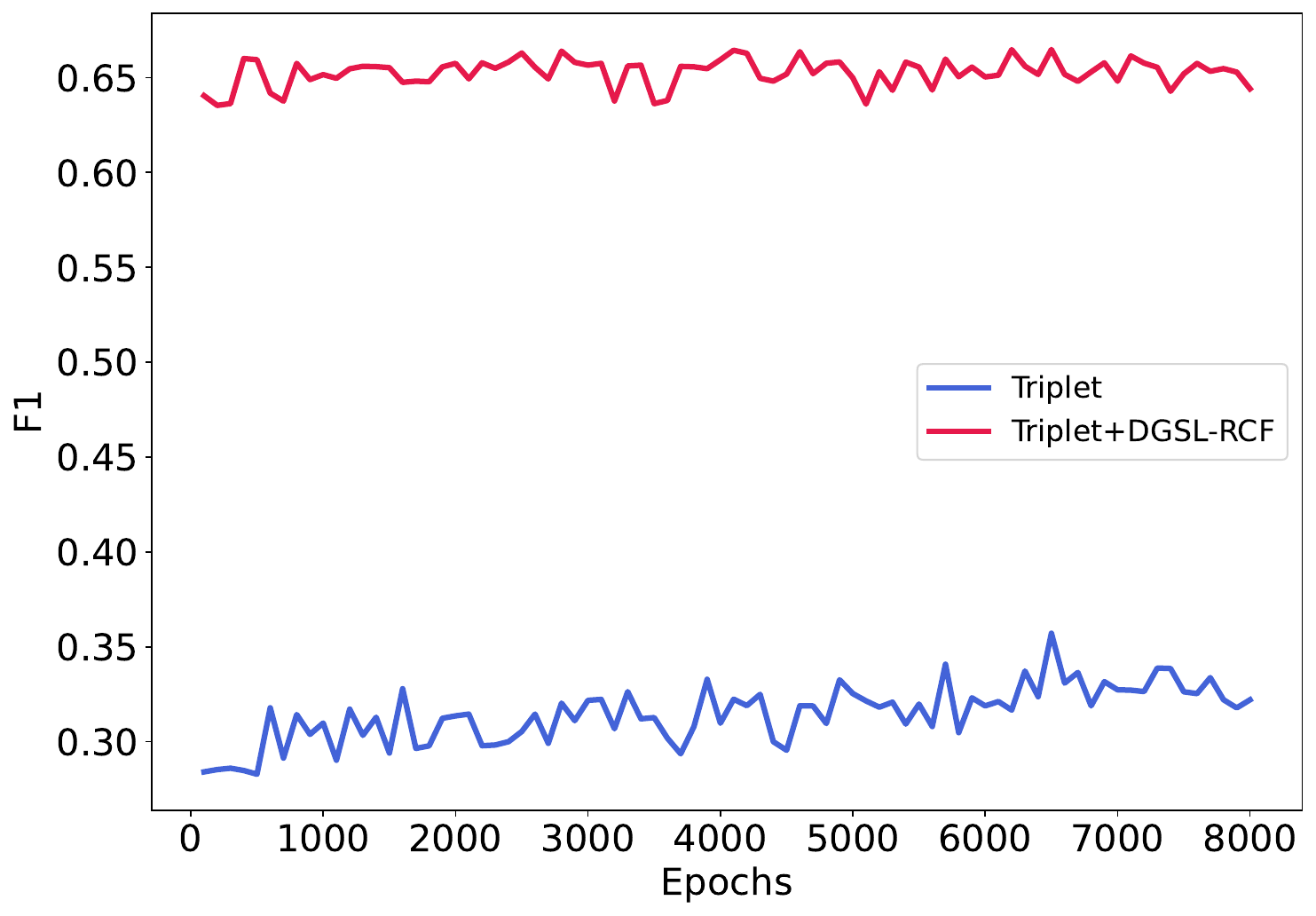}
\includegraphics[width=0.24\columnwidth]{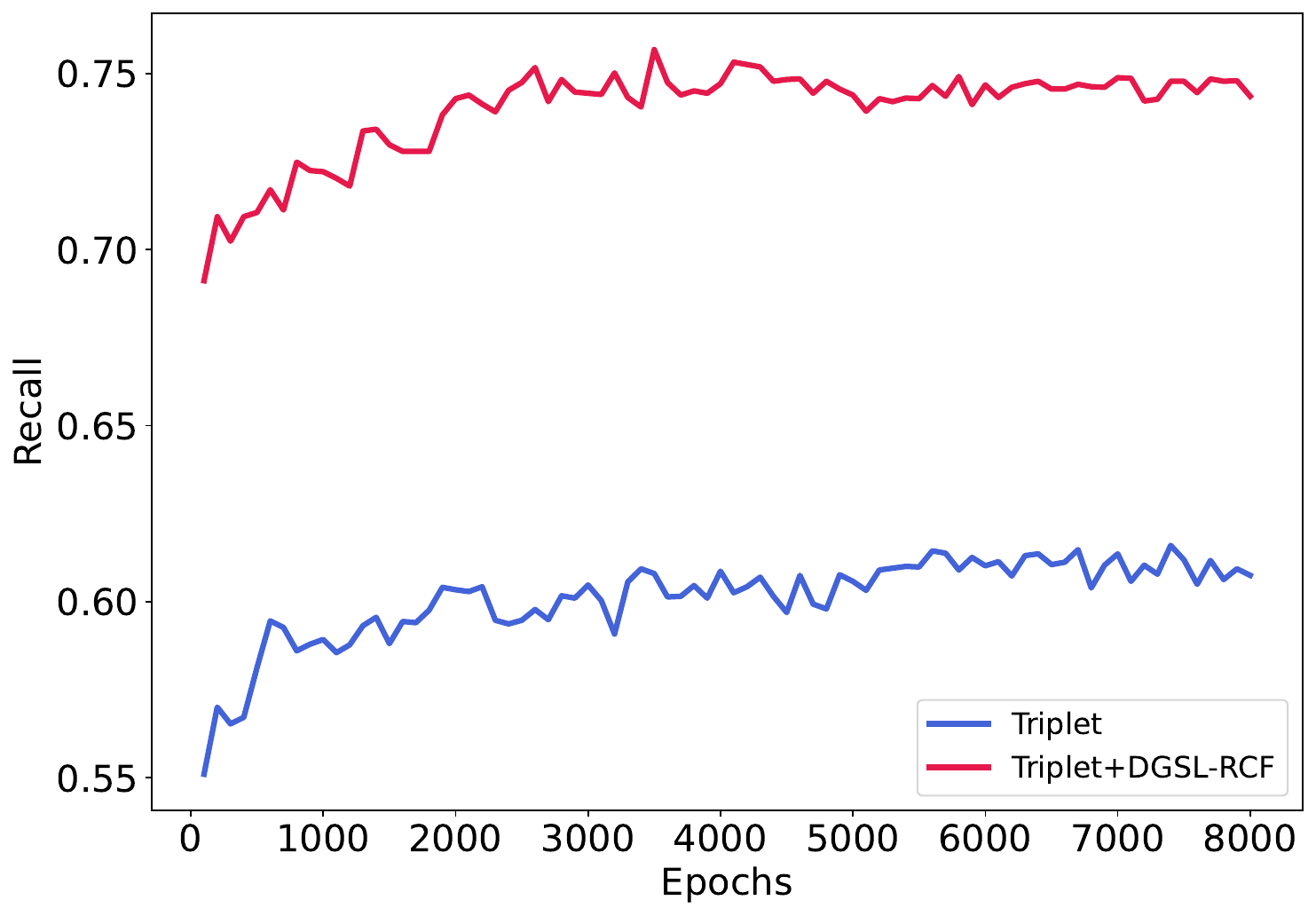}
\includegraphics[width=0.24\columnwidth]{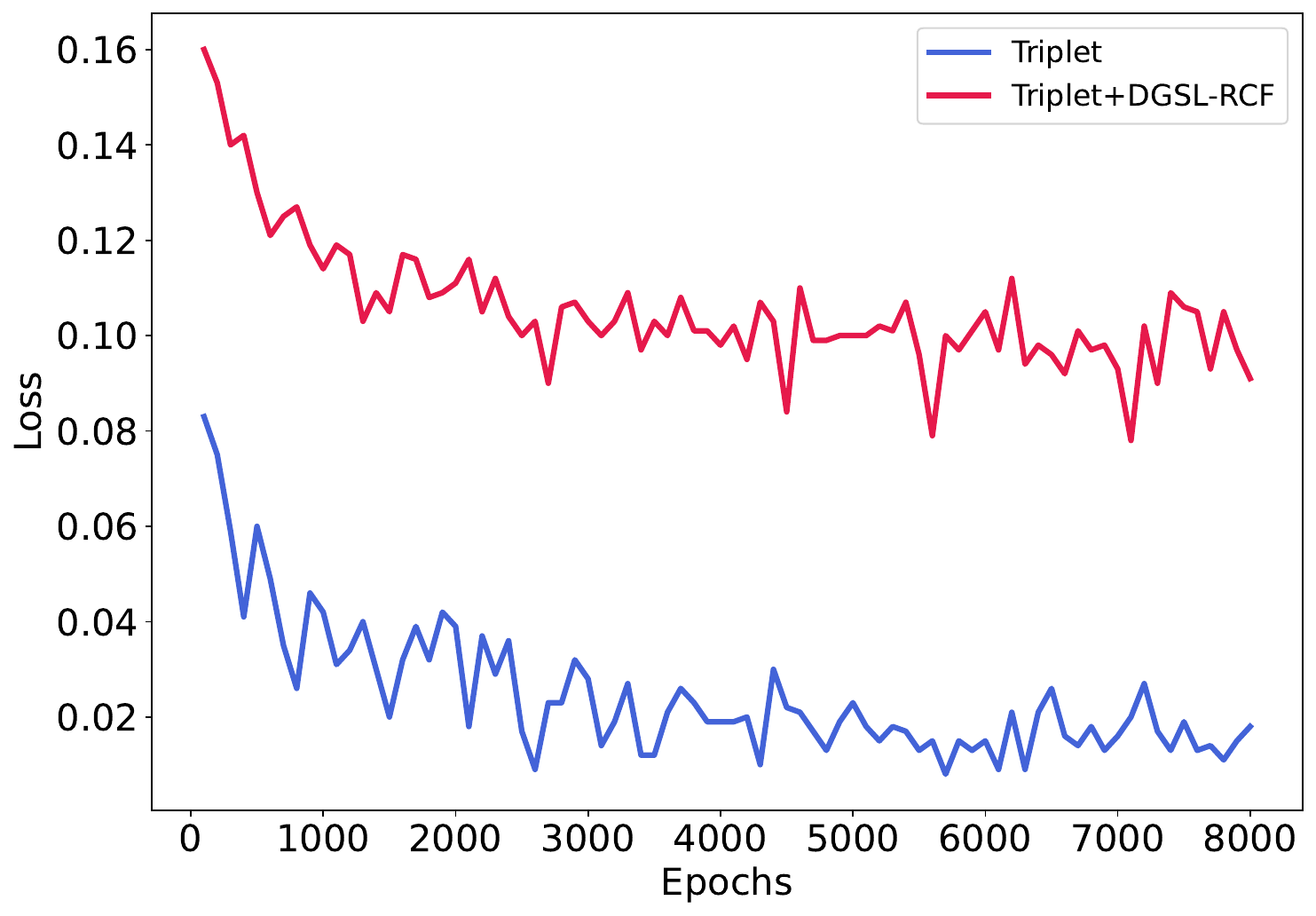}
\\
\includegraphics[width=0.24\columnwidth]{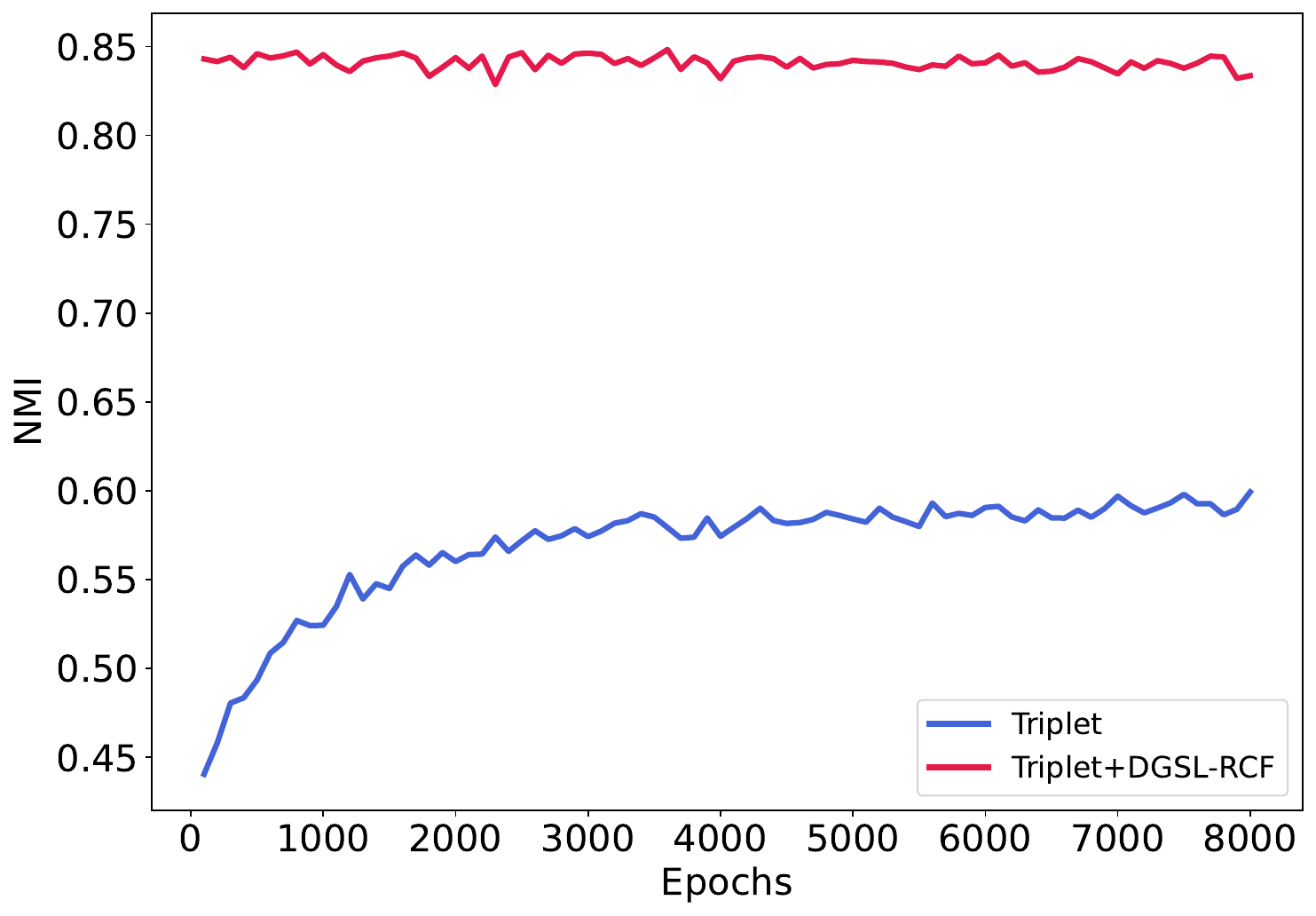}
\includegraphics[width=0.24\columnwidth]{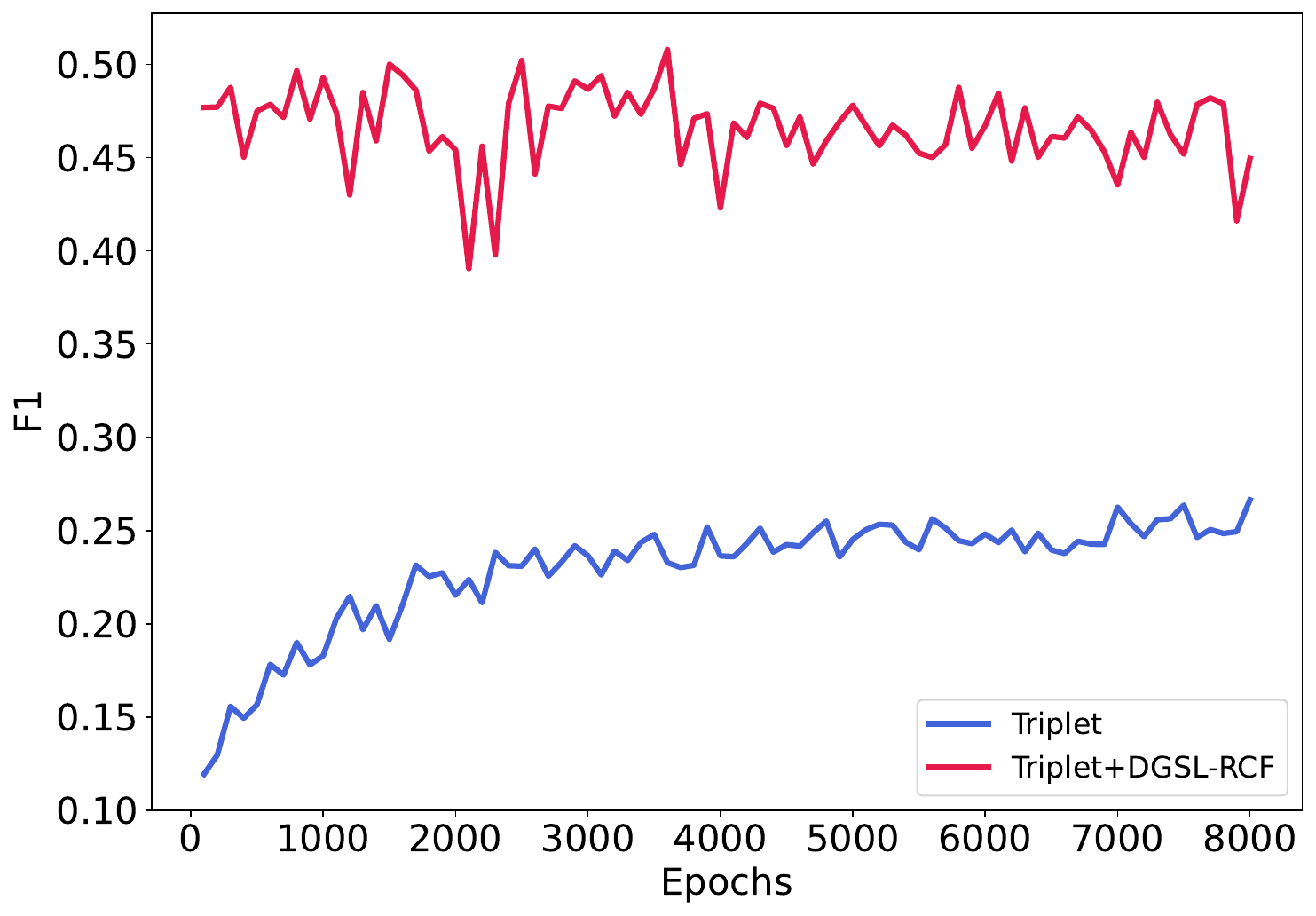}
\includegraphics[width=0.24\columnwidth]{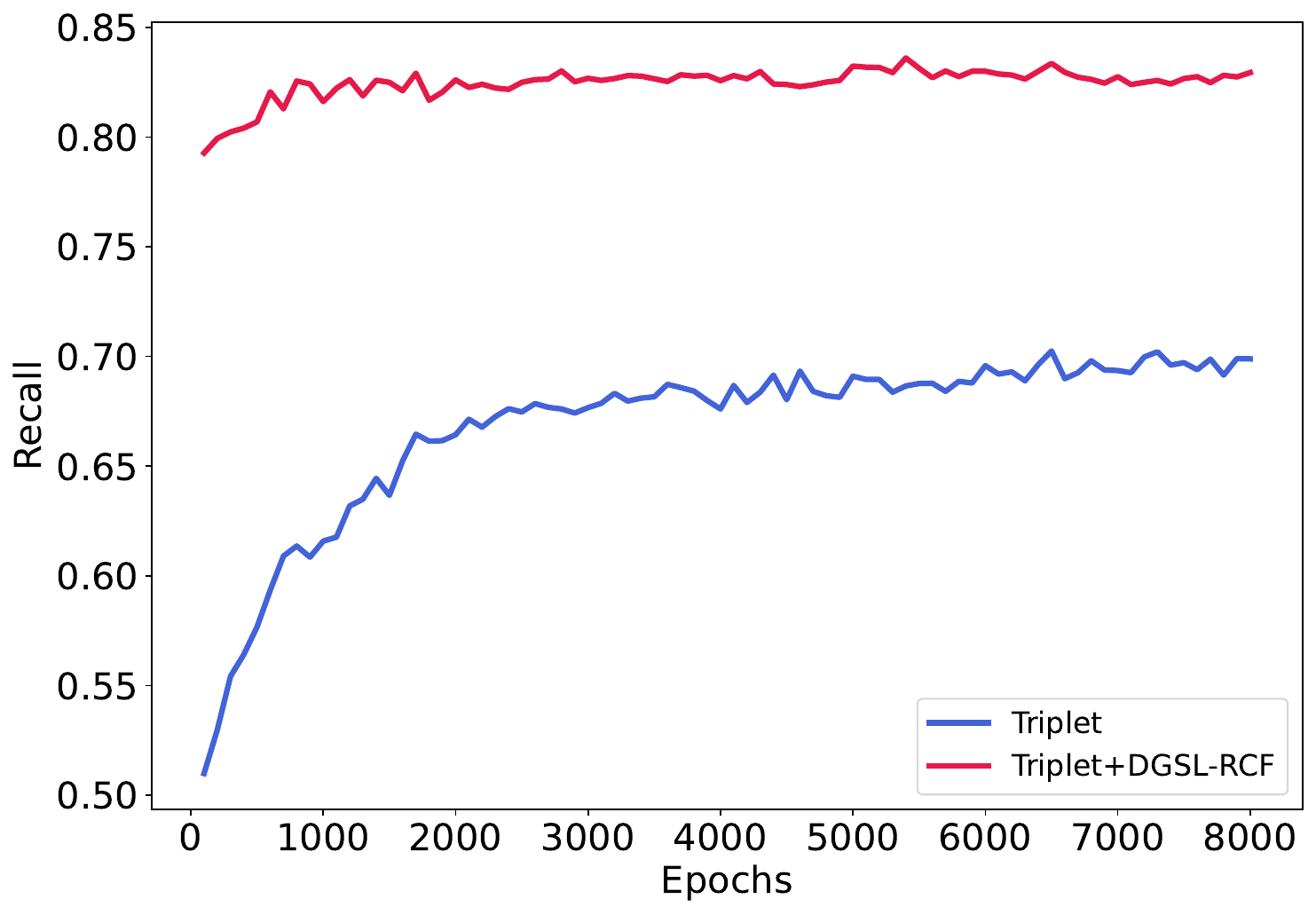}
\includegraphics[width=0.24\columnwidth]{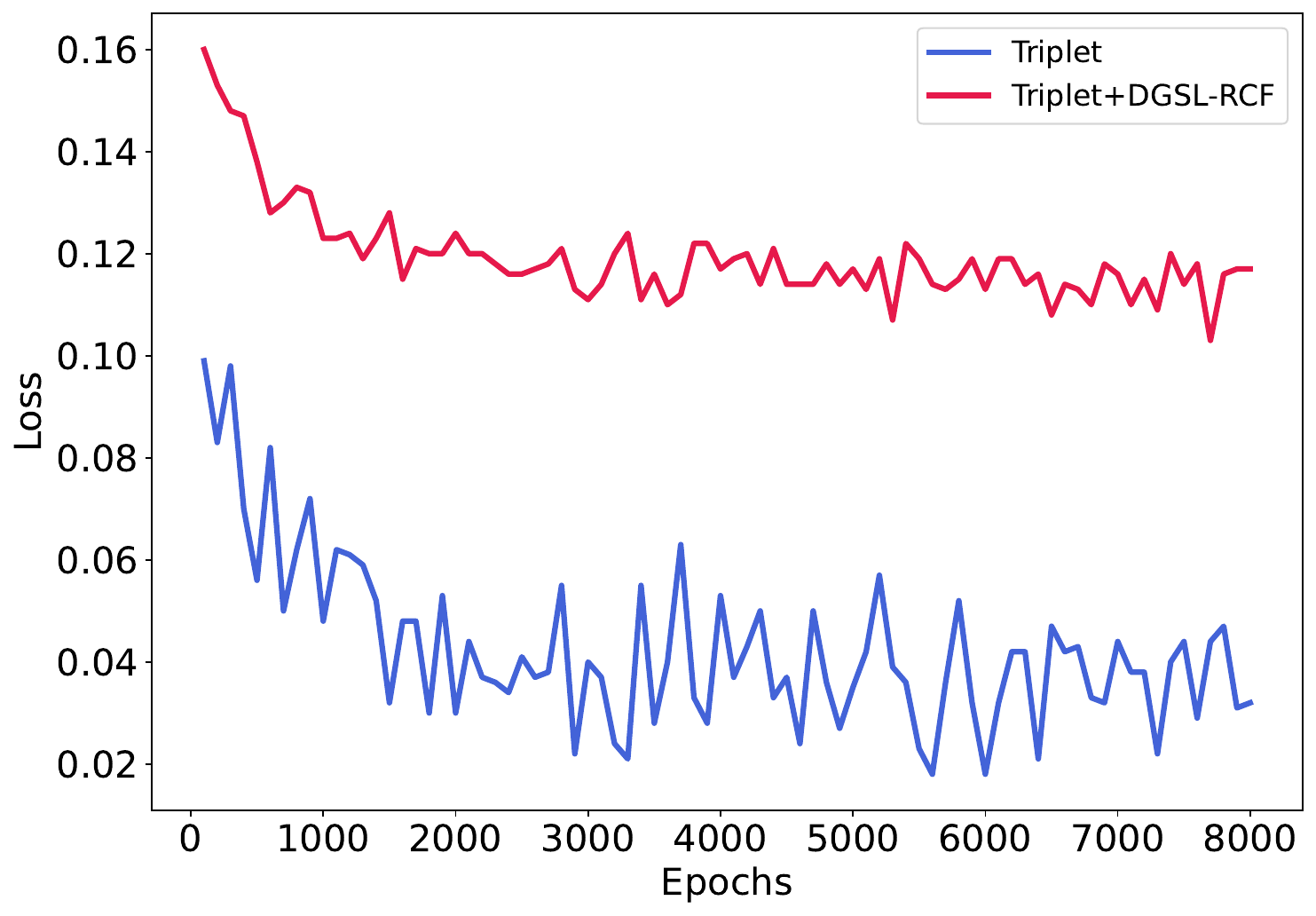}
\captionsetup{hypcap=false}
\captionof{figure}{Comparison of convergence on NMI, F1 and Recall metric between the +DGSL-RCF method and the baseline method on the CUB-200-2011 (Top) and Cars-196 (Bottom) datasets for DML task.}
\label{fig:dml_convergence}
\end{minipage}
\end{center}

\subsection{Hyperparameter Sensitivity Analysis}
To evaluate the stability and robustness of the DGSL-RCF algorithm, we focus on the DML and ML tasks, analyzing the impact of key hyperparameters:$k$, $n\_iter$, and $\eta$. In the hyperparameter experiments, for the DML task, the CUB-200-2011 and Cars-196 datasets are used with Triplet Loss, and the observed metrics are ACC and NMI. For the ML experiments, the four synthetic manifold datasets mentioned in Section \ref{sec_dataset} are used, and the observed metrics are NMI, F1, and Recall.

\subsubsection{Impact of Neighborhood Size $k$}
For DML Task, we analyze the influence of $k$ in terms of graph density (density = $k$/batch size). The experimental results are shown in Figure~\ref{fig:dml_k}.

Key observations are as follows:
\begin{itemize}[topsep=0pt, parsep=0pt]
\item When batch size = 20, as density increases, recall improves significantly, while NMI and F1 remain almost unchanged. This indicates that under a small batch size with a relatively sparse graph structure, increasing density helps capture more marginal or hard positive samples, thereby substantially enhancing retrieval completeness.

\item When batch size = 40, as density increases, NMI and F1 increase, whereas recall remains largely stable. A moderate increase in density strengthens local semantic aggregation and optimizes intra-class structure. However, excessively high density may introduce heterophilic noise, leading to saturation in the improvement of NMI and F1.
\end{itemize}

For ML task, We evaluate the performance with $k \in \{10, 15, \dots, 60\}$ on four synthetic datasets (scale N=1,000). The ACC results are shown in Figure~\ref{fig:ml_k} (NMI exhibits a consistent trend; see~\ref{appendix:ml_other_exp} for details). The experiments reveal that the sensitivity of $k$ strongly depends on the underlying manifold structure.

For instance, on regular manifolds such as Gaussian Surface, performance improvement tends to saturate after $k>35$, indicating that its geometric structure has been sufficiently captured. Conversely, on complex manifolds like Swiss Roll, performance begins to decline after $k>30$. This is because an excessively large neighborhood disrupts the local linearity of the Swiss Roll, causing over-smoothing and structural distortion during manifold unfolding.

These findings confirm that selecting $k$ is essentially a trade-off between preserving local linearity and maintaining global structural coherence.

\begin{center}
\begin{minipage}{\columnwidth}
\centering
	\includegraphics[width=0.24\columnwidth]{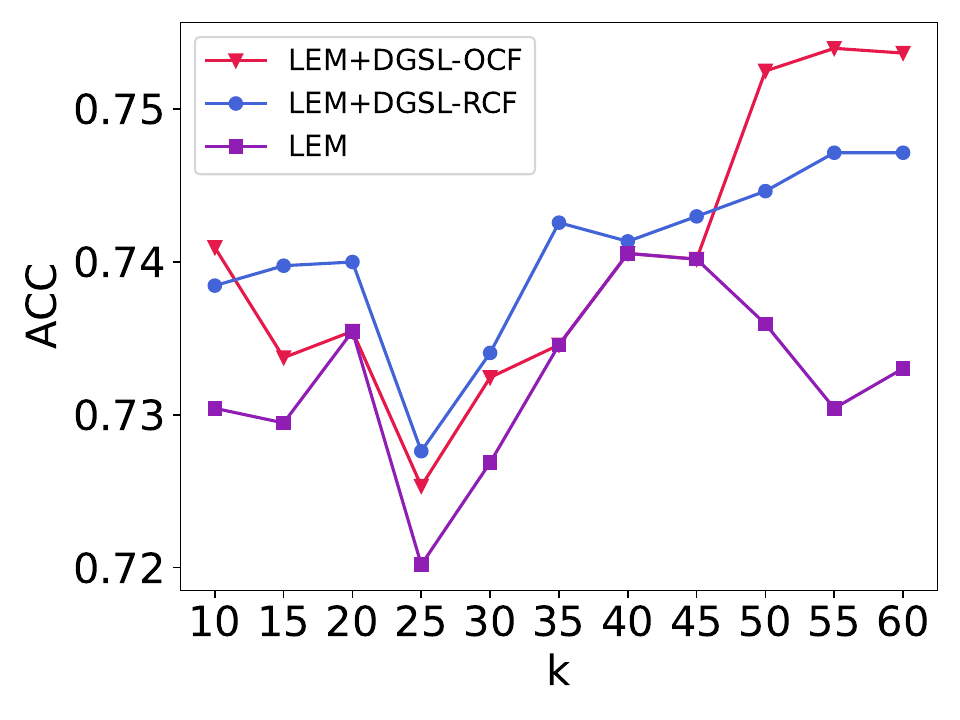}
    \includegraphics[width=0.24\columnwidth]{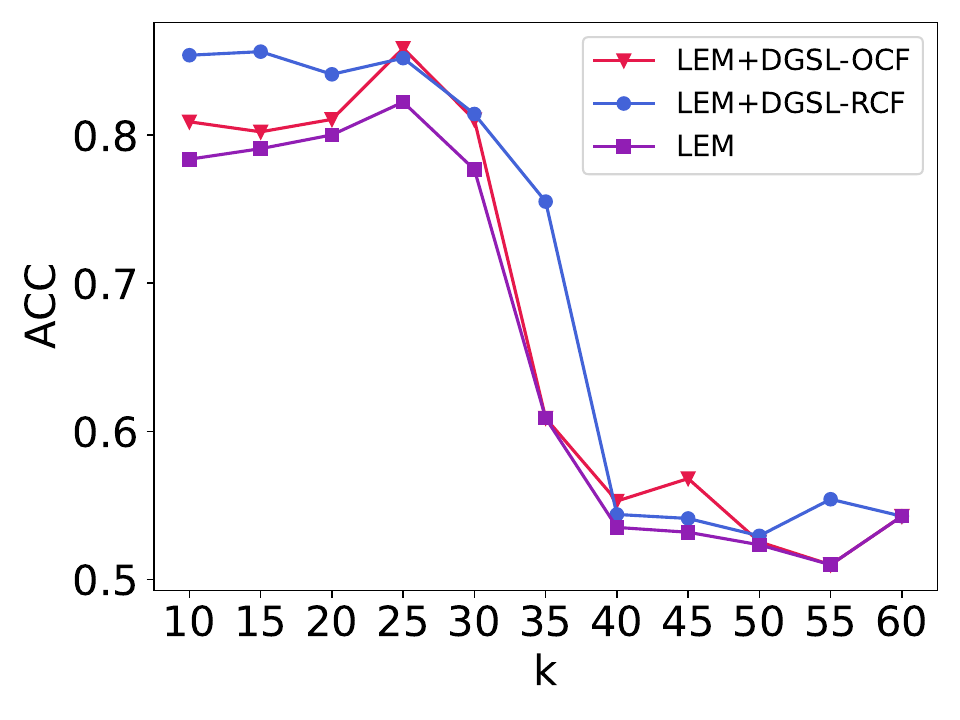}
    \includegraphics[width=0.24\columnwidth]{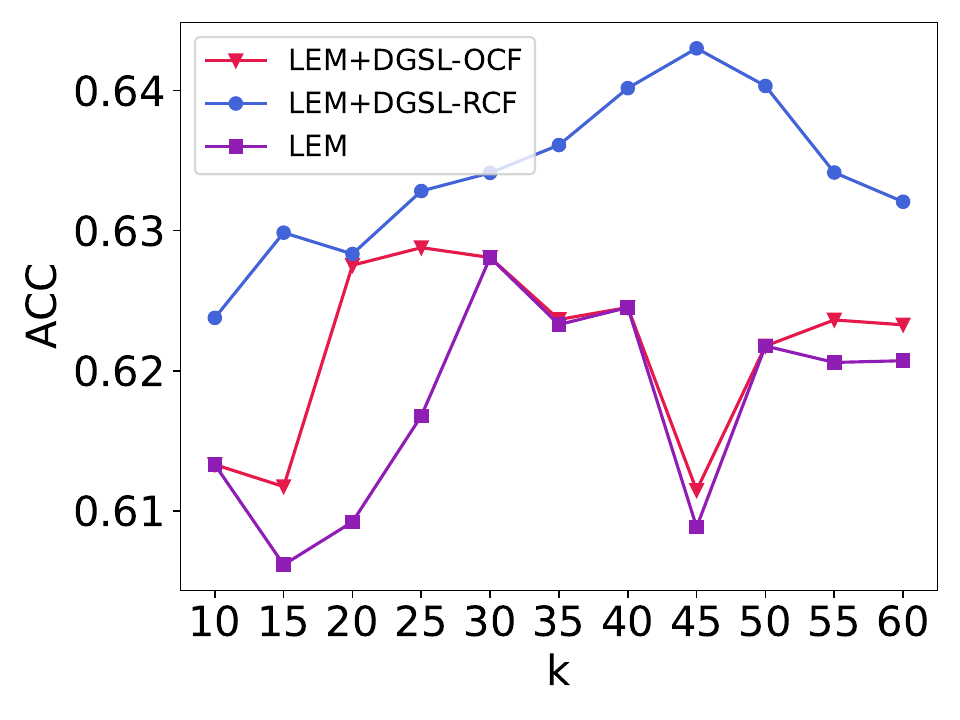}
    \includegraphics[width=0.24\columnwidth]{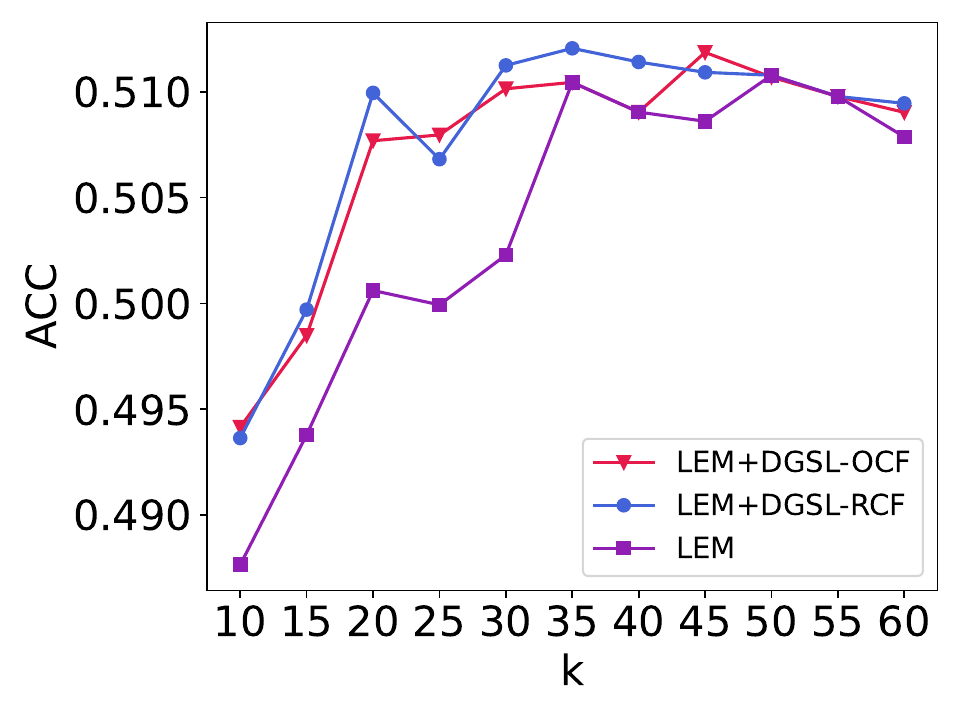}
\captionsetup{hypcap=false}
\captionof{figure}{Impact of hyperparameter $k$ on the ACC metric of LEM, LEM+OCF, and LEM+RCF in ML task (top-left: S Curve, top-right: Swiss Roll, bottom-left: Truncated Sphere, bottom-right: Gaussian Surface).}
\label{fig:ml_k}
\end{minipage}
\end{center}

\begin{center}
\begin{minipage}{\columnwidth}
\centering
	\includegraphics[width=0.32\columnwidth]{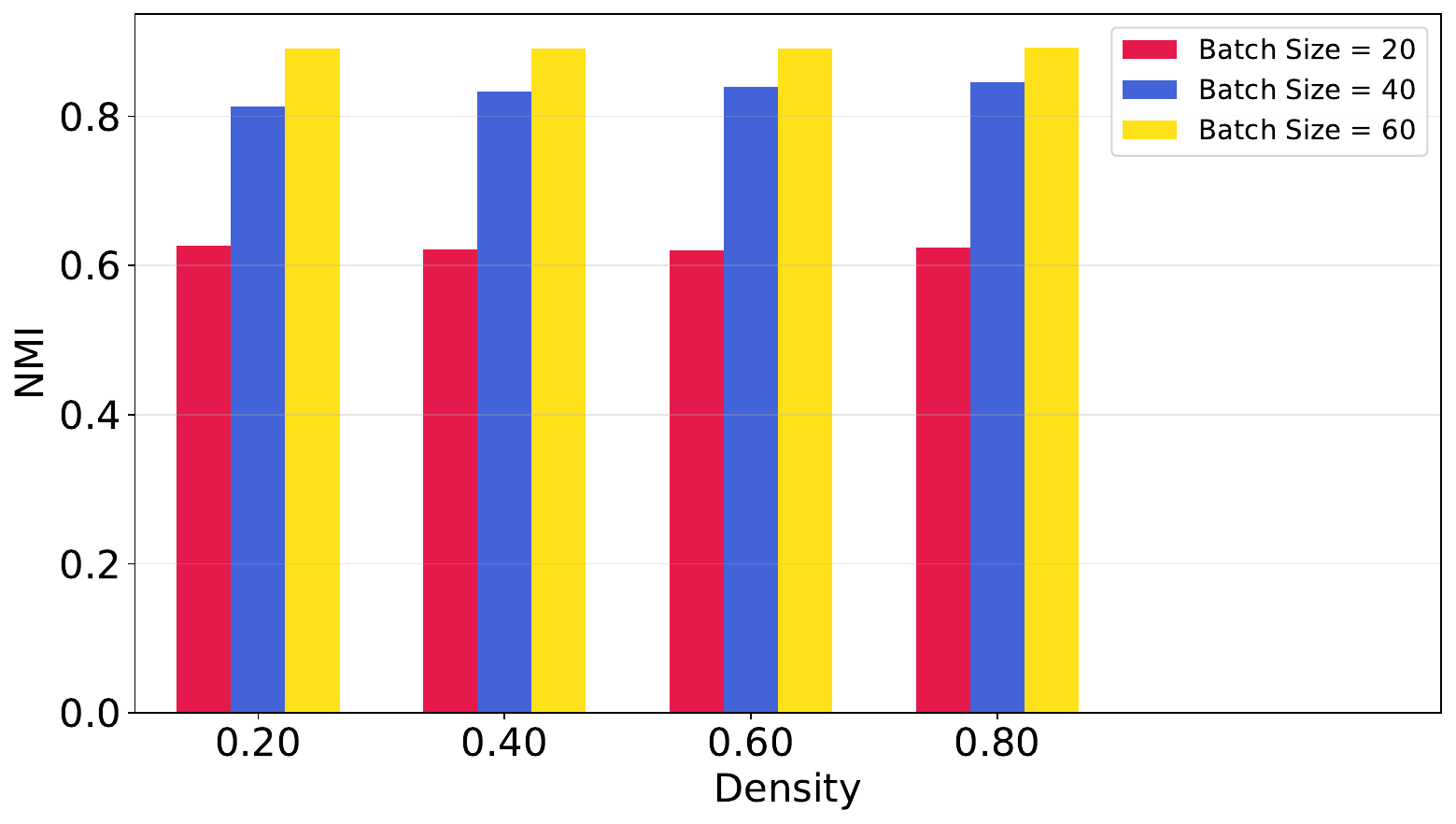}
    \includegraphics[width=0.32\columnwidth]{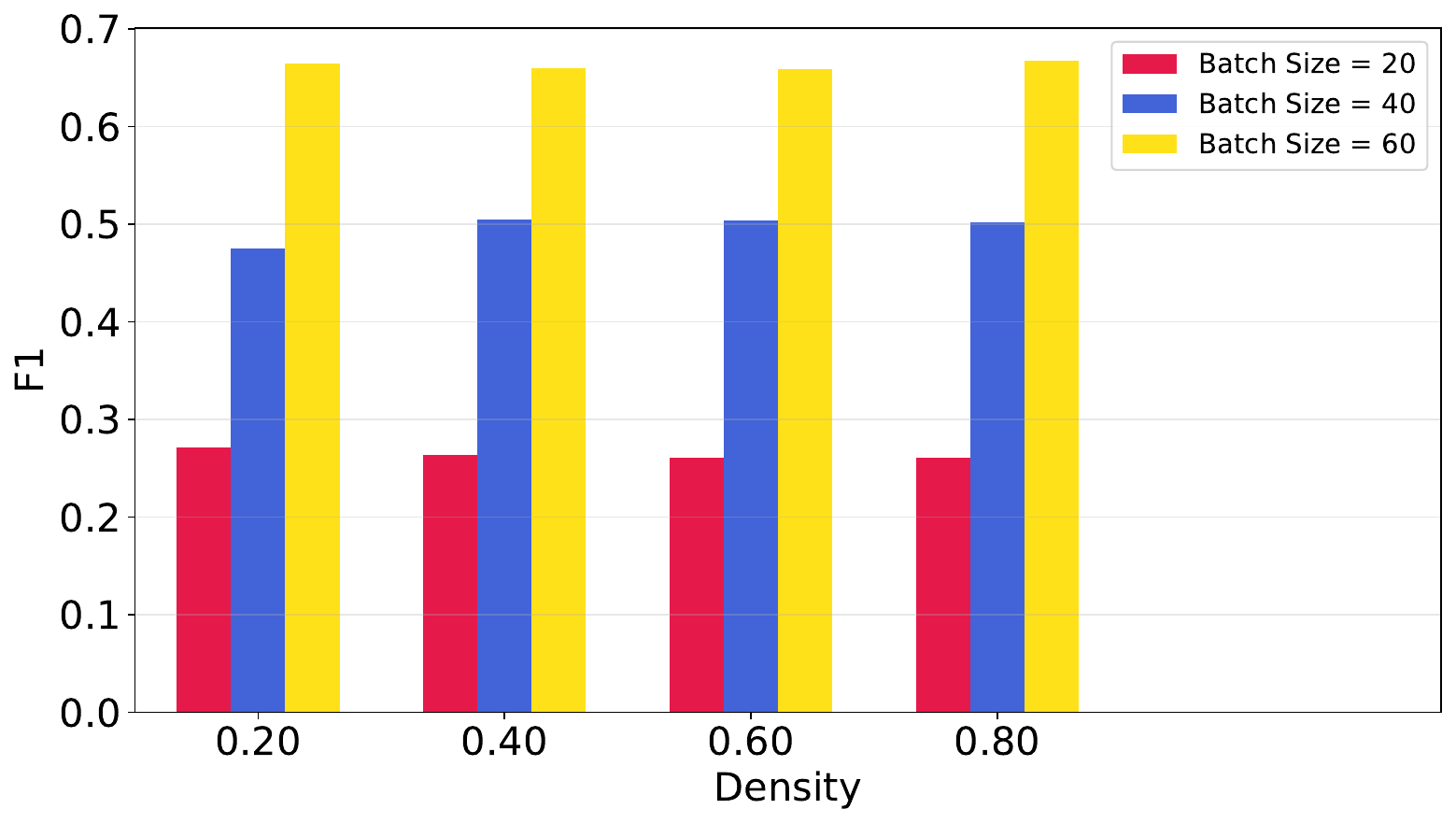}
    \includegraphics[width=0.32\columnwidth]{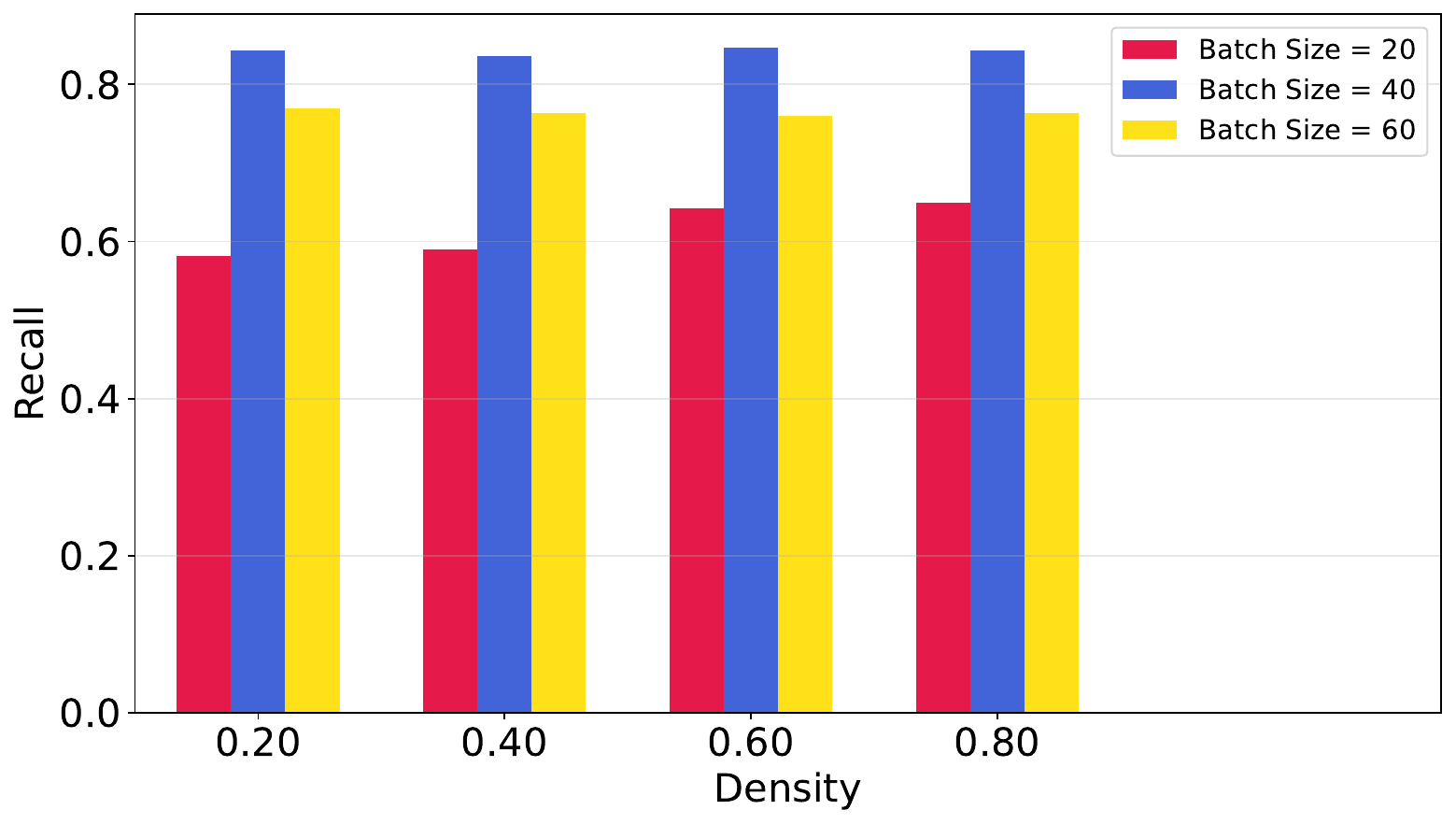}
    \\
    \includegraphics[width=0.32\columnwidth]{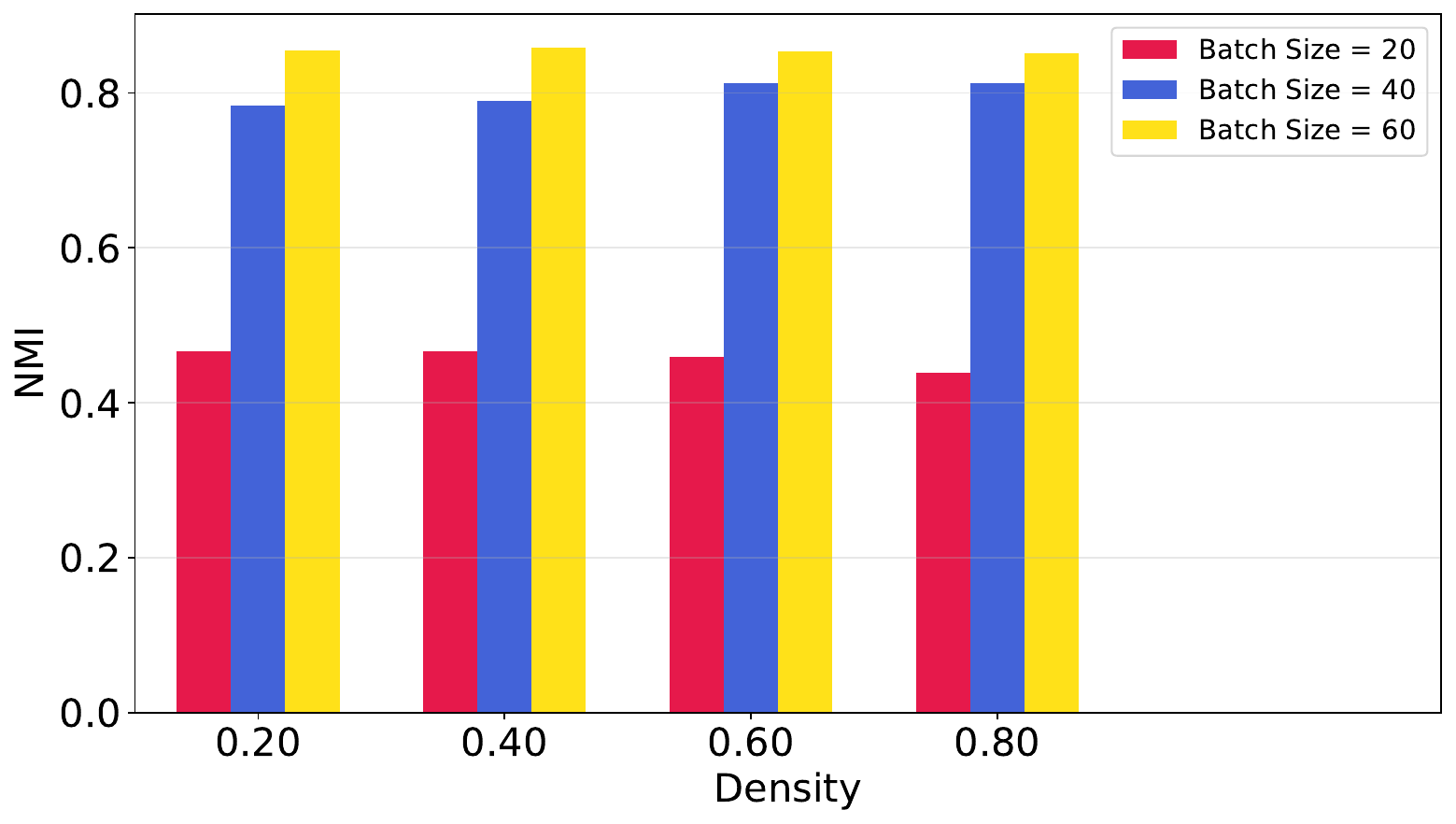}
    \includegraphics[width=0.32\columnwidth]{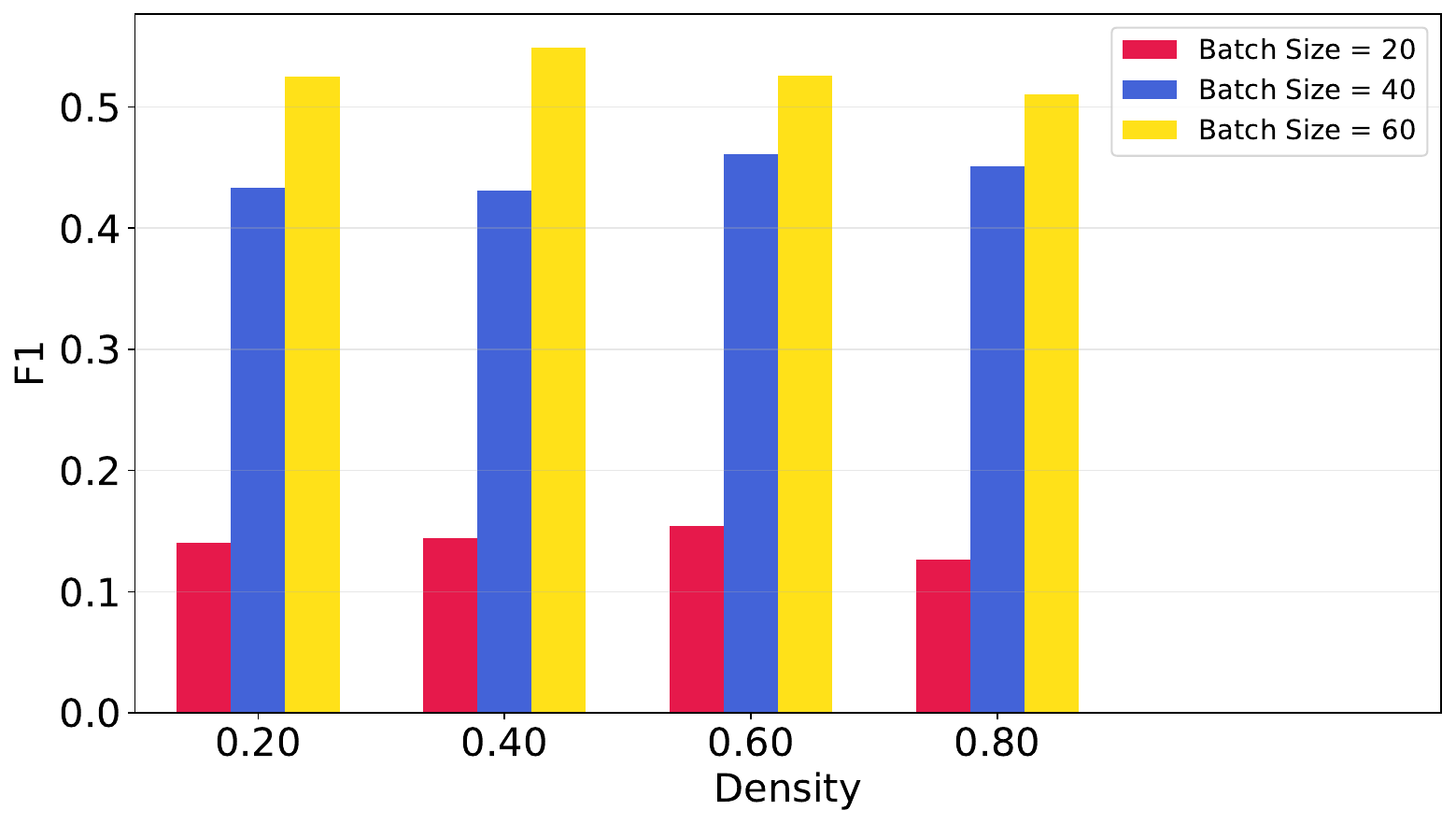}
    \includegraphics[width=0.32\columnwidth]{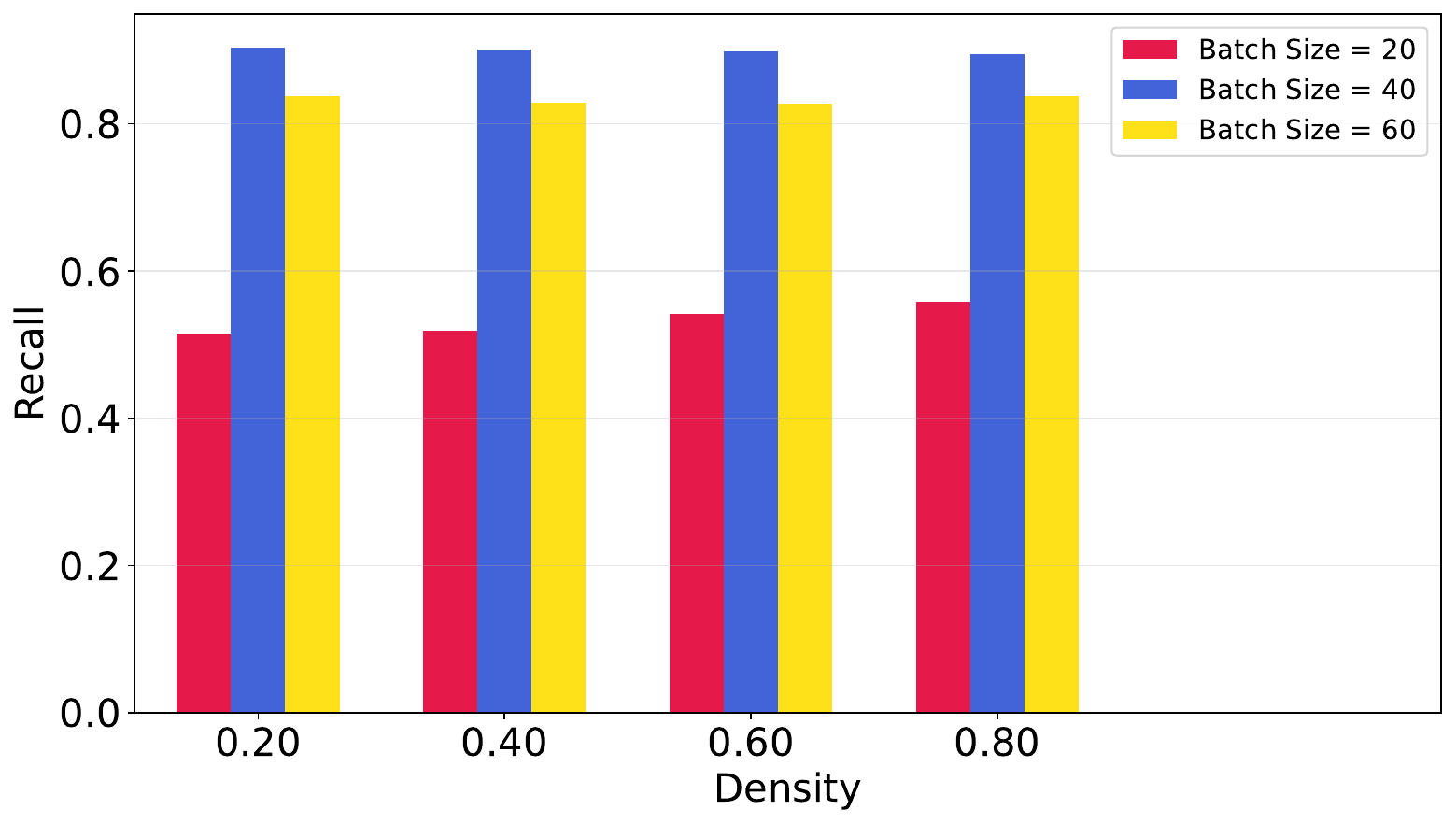}
\captionsetup{hypcap=false}
\captionof{figure}{Impact of hyperparameter $k$ (density) on the NMI (left), F1(middle) and Recall (right) under different batch size in DML task (top: CUB-200-2011, bottom: Cars-196).}
\label{fig:dml_k}
\end{minipage}
\end{center}

\subsubsection{Impact of Iteration Number $n\_iter$}
The iteration count $n\_iter$ determines the depth of the curvature flow optimization process. The observed values for $n\_iter$ are $\{1, 2, \dots ,7\}$. Experiments on both DML and ML tasks reveal that DGSL-RCF exhibits rapid convergence in most tasks, with optimal performance typically achieved within the first or second iteration (Figure~\ref{fig:dml_n_iter}, ~\ref{fig:ml_n_iter}).

This phenomenon indicates that geometric repair based on Resistance Curvature is a highly efficient process. In the first iteration, the algorithm can quickly identify and correct the most unreasonable connections in the initial topology. The diminishing marginal returns of subsequent iterations suggest that major geometric structural defects are resolved early on. This highlights the computational efficiency advantage of RCF. This characteristic is crucial for scenarios requiring integration into deep learning training loops, as it implies that significant geometric structure optimization can be achieved with negligible additional computational overhead.
\begin{center}
\begin{minipage}{\columnwidth}
\centering
	\includegraphics[width=0.24\columnwidth]{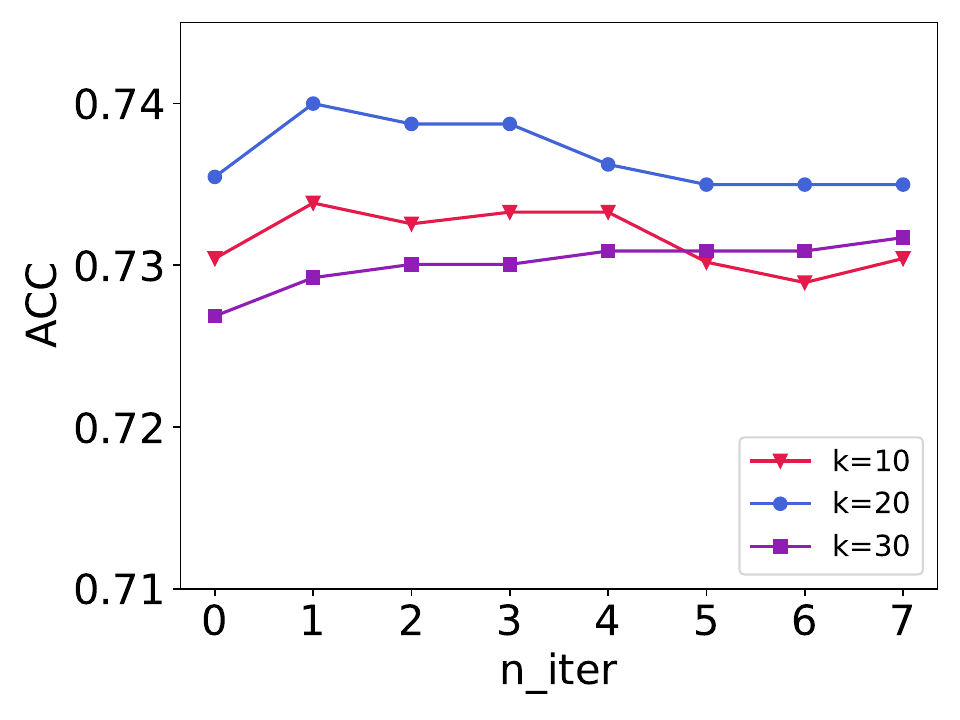}
    \includegraphics[width=0.24\columnwidth]{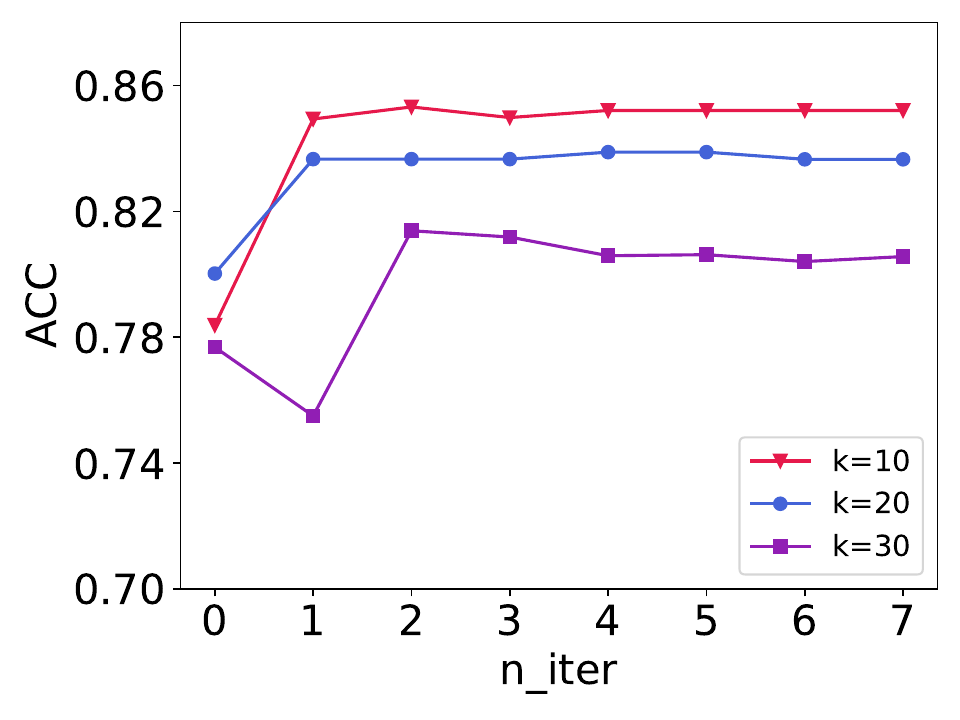}
    \includegraphics[width=0.24\columnwidth]{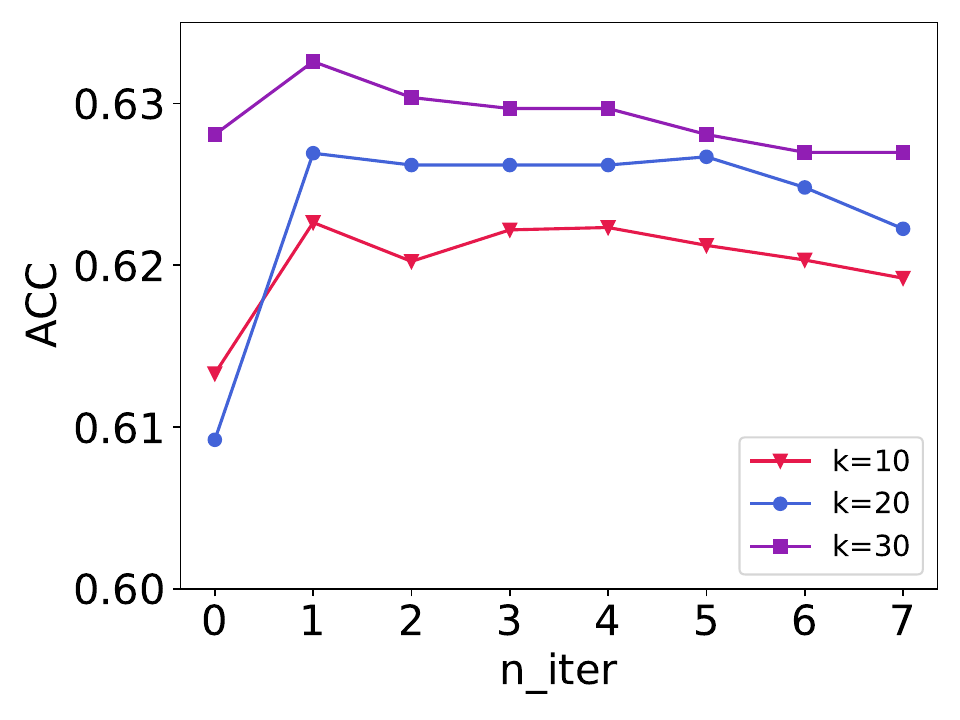}
    \includegraphics[width=0.24\columnwidth]{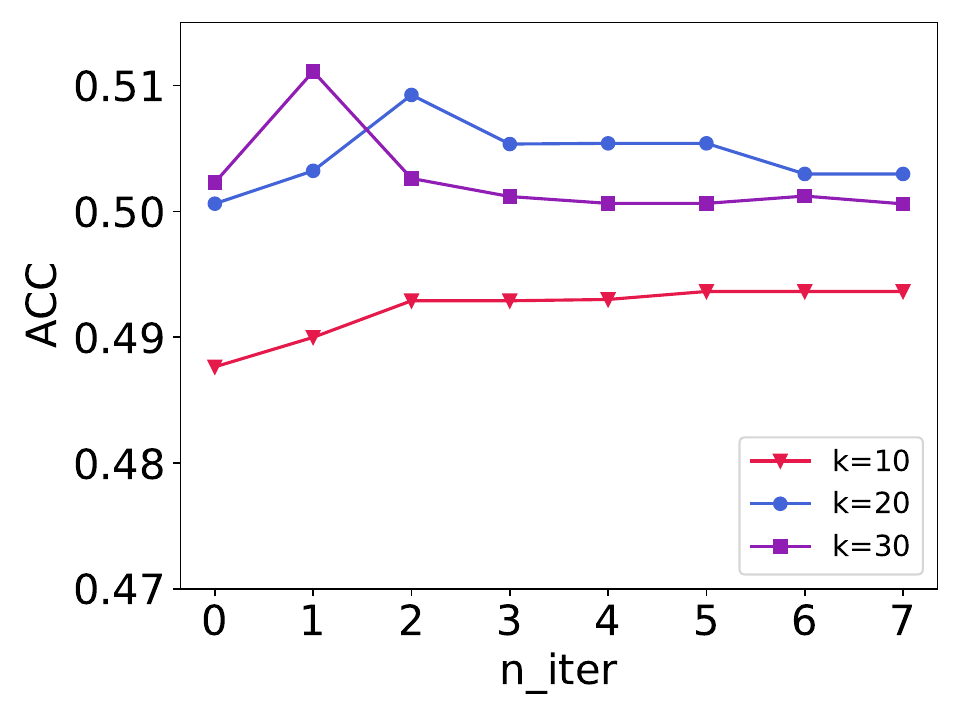}
\captionsetup{hypcap=false}
\captionof{figure}{Impact of hyperparameter $n\_iter$ on the ACC metric in ML task under different $k$ (top-left: S Curve, top-right: Swiss Roll, bottom-left: Truncated Sphere, bottom-right: Gaussian Surface).}
\label{fig:ml_n_iter}
\end{minipage}
\end{center}

\begin{center}
\begin{minipage}{\columnwidth}
\centering
	\includegraphics[width=0.4\columnwidth]{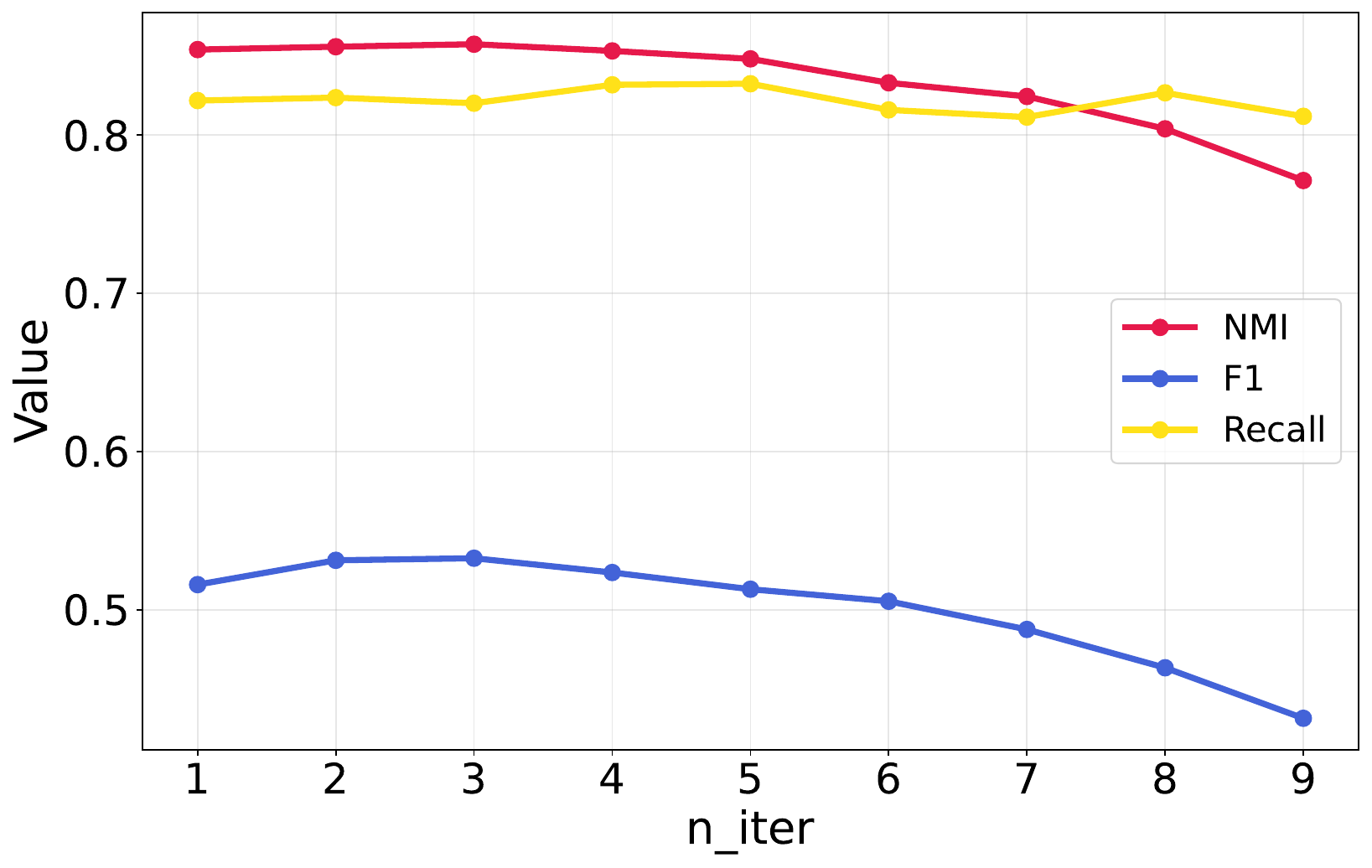}
    \includegraphics[width=0.4\columnwidth]{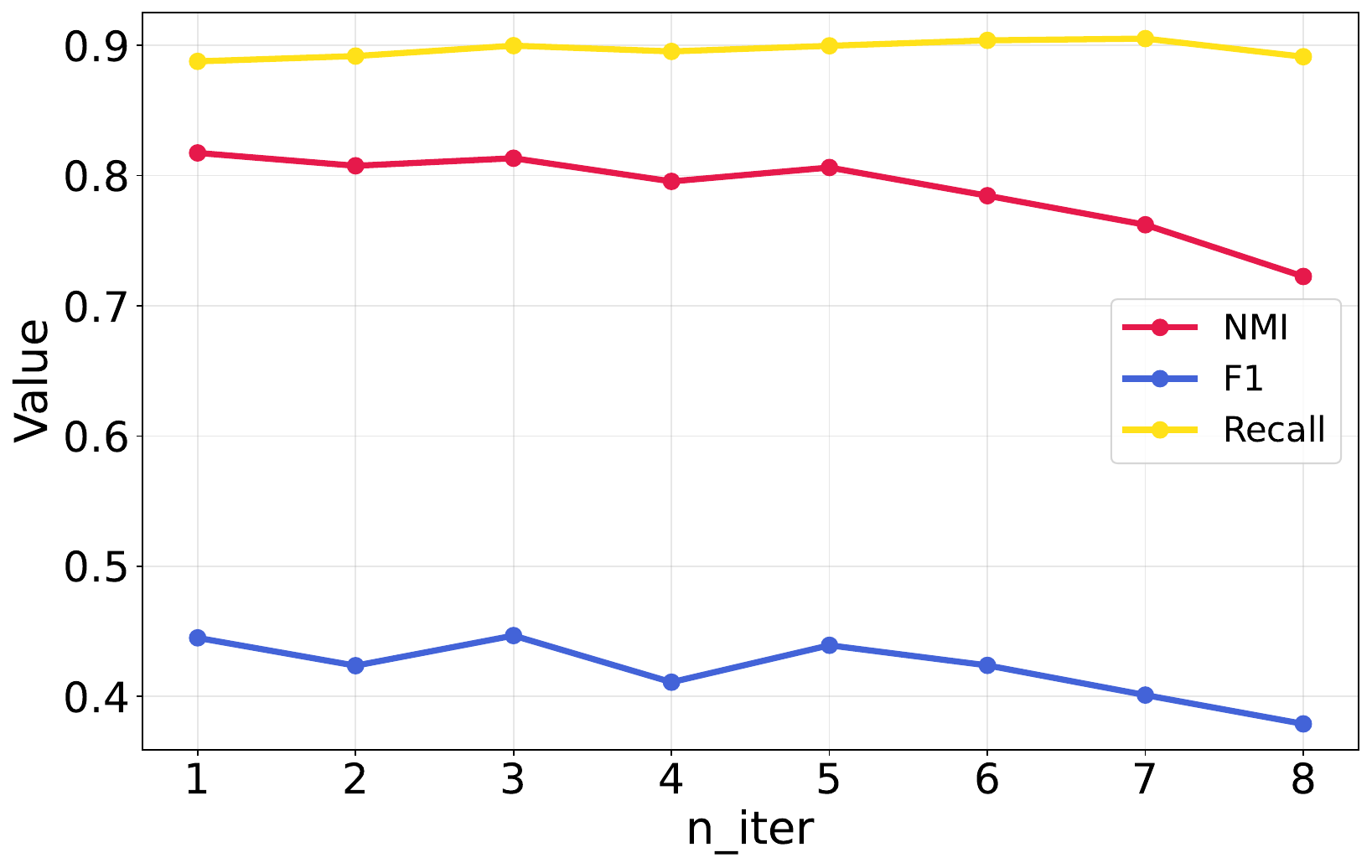}
\captionsetup{hypcap=false}
\captionof{figure}{Impact of hyperparameter $n\_iter$ on the NMI, F1 and Recall metrics in DML task (left: CUB-200-2011, right: Cars-196).}
\label{fig:dml_n_iter}
\end{minipage}
\end{center}

\subsubsection{Impact of the RCF Learning Rate $\eta$}
The curvature flow step length $\eta$ controls the intensity of geometric structure updates in a single iteration. The observed values for $\eta$ are $\{0.1, 0.2, \dots, 1.0\}$. Experiments show that its influence exhibits significant structural dependency and parameter interaction.

Detailed observations are as follows:
\begin{itemize}[topsep=0pt, parsep=0pt]
    \item Robustness Interval: In the DML task, when $\eta<0.6$, model performance remains highly stable with minimal fluctuation (Figure~\ref{fig:dml_eta}). In the ML task, within a reasonable value range, the impact of DGSL-RCF on model performance is relatively stable. For example, on the S-Curve dataset with $k=10$, the ACC results differ negligibly when $\eta$ ranges from 0.1 to 0.5. This indicates that DGSL-RCF is insensitive to the choice of learning rate within a range, demonstrating good robustness and reducing the difficulty of parameter tuning.
    \item Interaction with neighborhood size $k$: In ML, the influence of $\eta$ is closely related to the density of the current graph (Figure~\ref{fig:ml_eta}). On sparse graphs (e.g., $k=10$), where geometric information is limited, the value of $\eta$ significantly affects the optimization path and final results. In contrast, on dense graphs (e.g., $k=50$), abundant connections provide sufficient structural redundancy, making the optimization process less sensitive to the step size of a single update. This illustrates that different hyperparameters are not independent but exhibit synergistic interaction effects.
\end{itemize}

\begin{center}
\begin{minipage}{\columnwidth}
\centering
	\includegraphics[width=0.24\columnwidth]{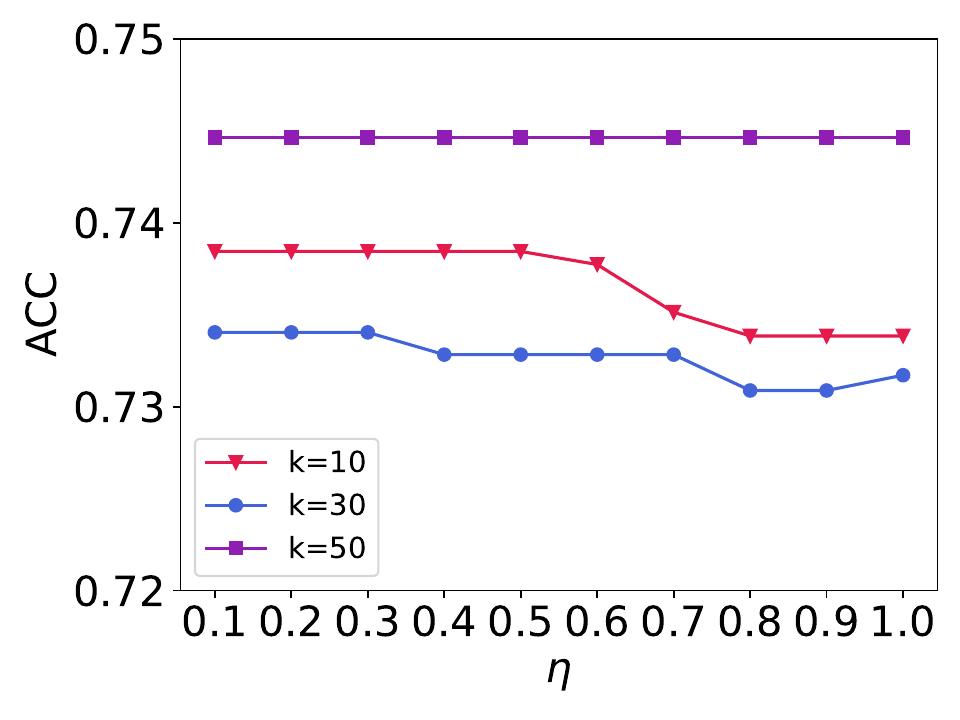}
    \includegraphics[width=0.24\columnwidth]{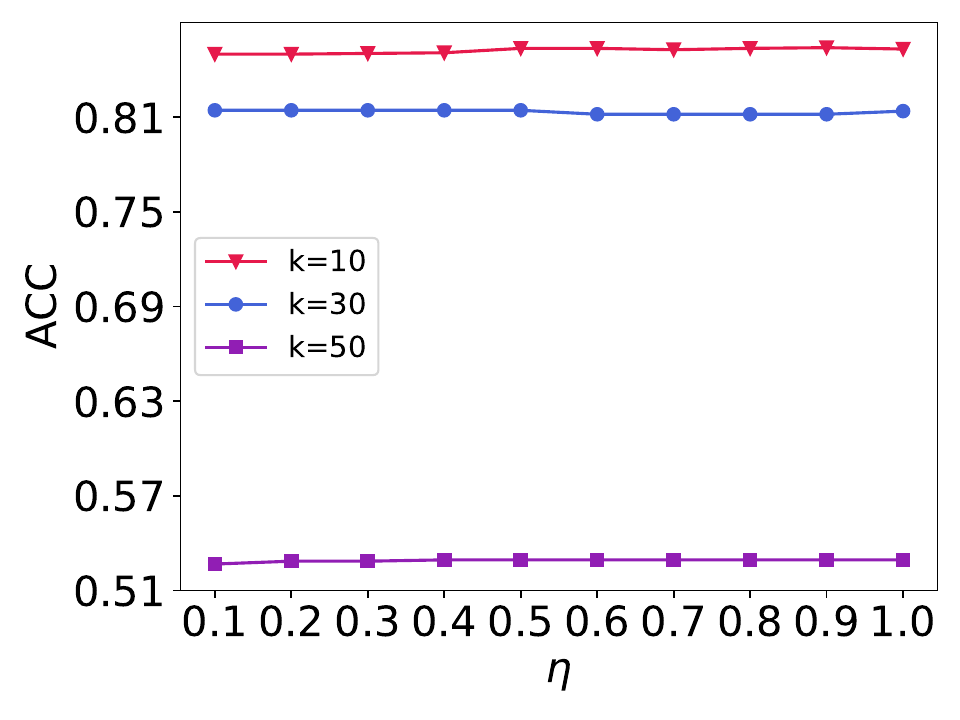}
    \includegraphics[width=0.24\columnwidth]{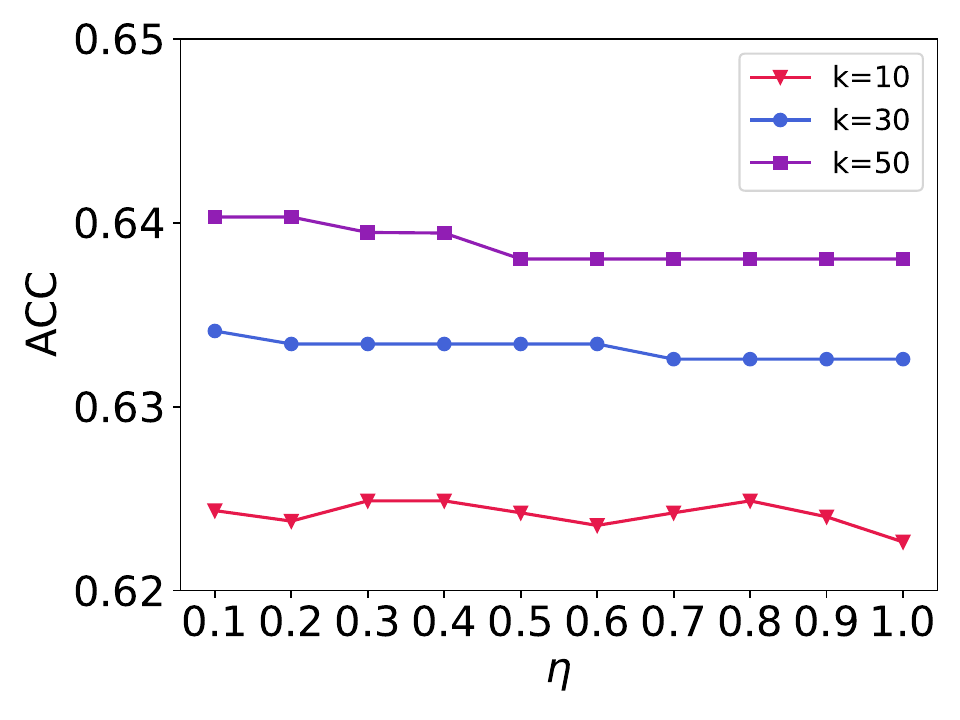}
    \includegraphics[width=0.24\columnwidth]{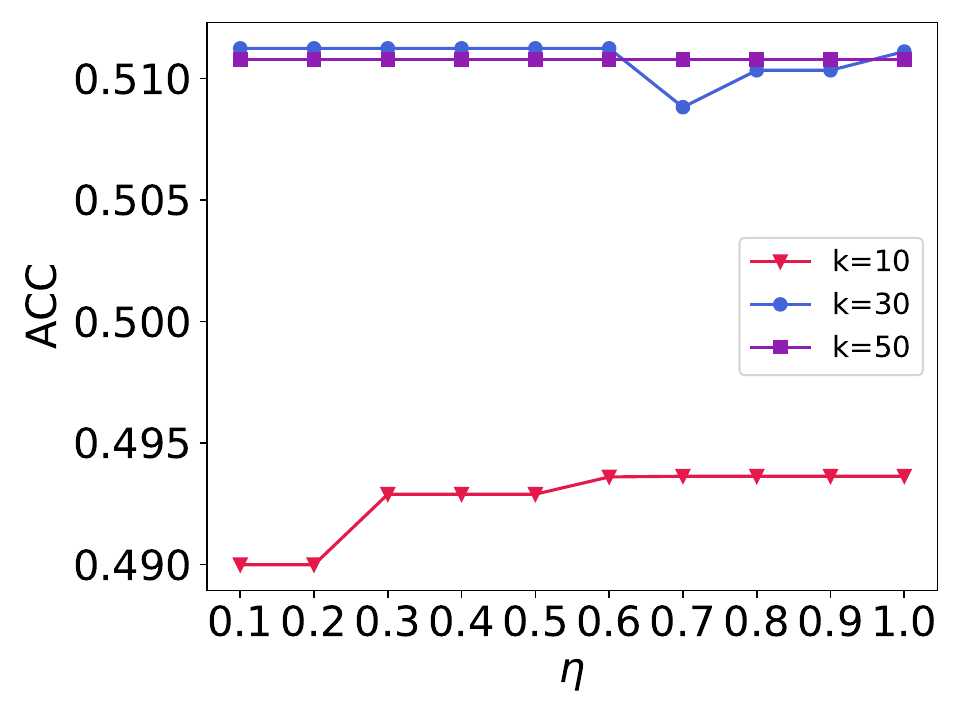}
\captionsetup{hypcap=false}
\captionof{figure}{Impact of hyperparameter $\eta$ on the ACC metric in ML task under different $k$ (top-left: S Curve, top-right: Swiss Roll, bottom-left: Truncated Sphere, bottom-right: Gaussian Surface).}
\label{fig:ml_eta}
\end{minipage}
\end{center}
\begin{center}
\begin{minipage}{\columnwidth}
\centering
	\includegraphics[width=0.47\columnwidth]{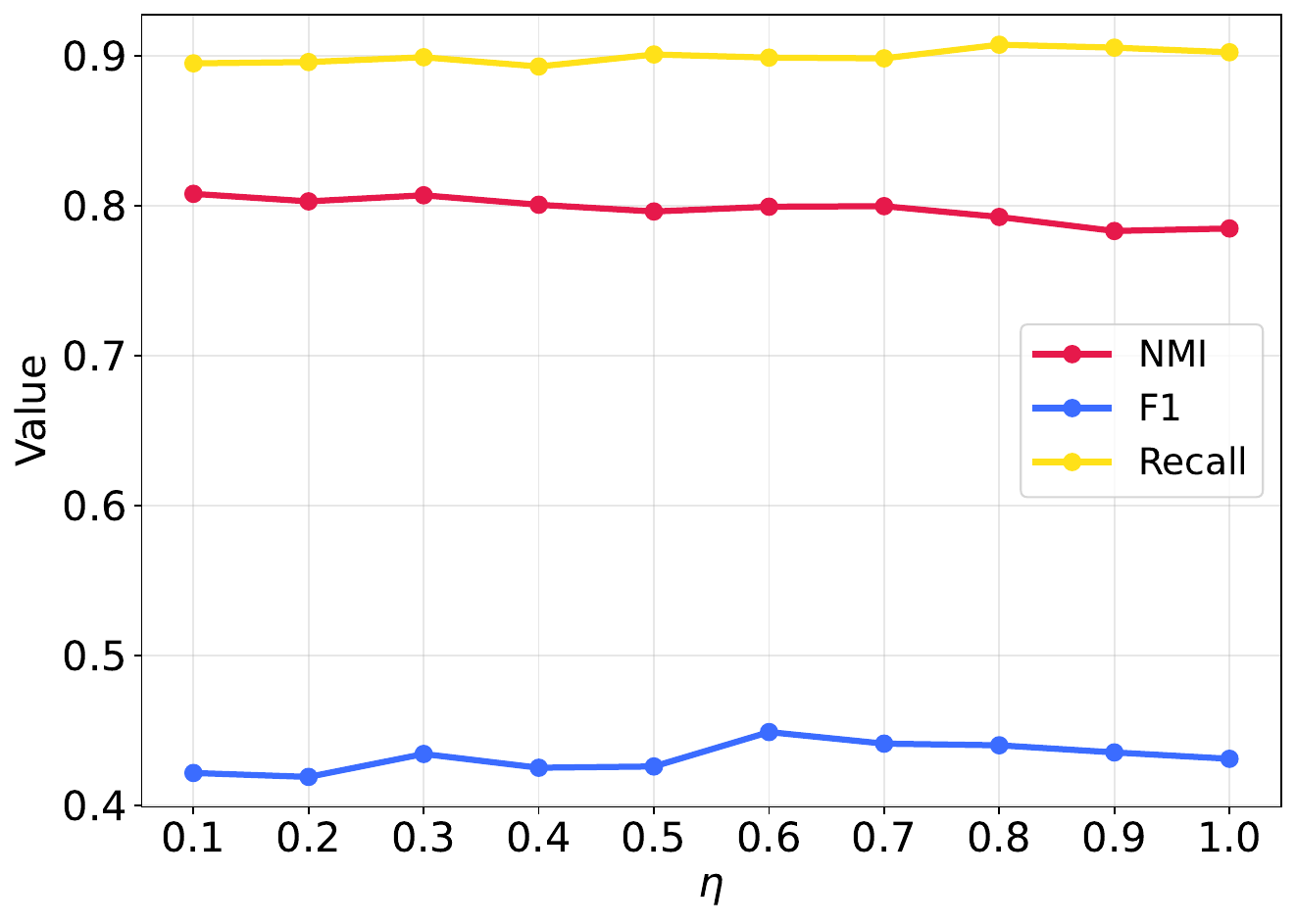}
    \includegraphics[width=0.47\columnwidth]{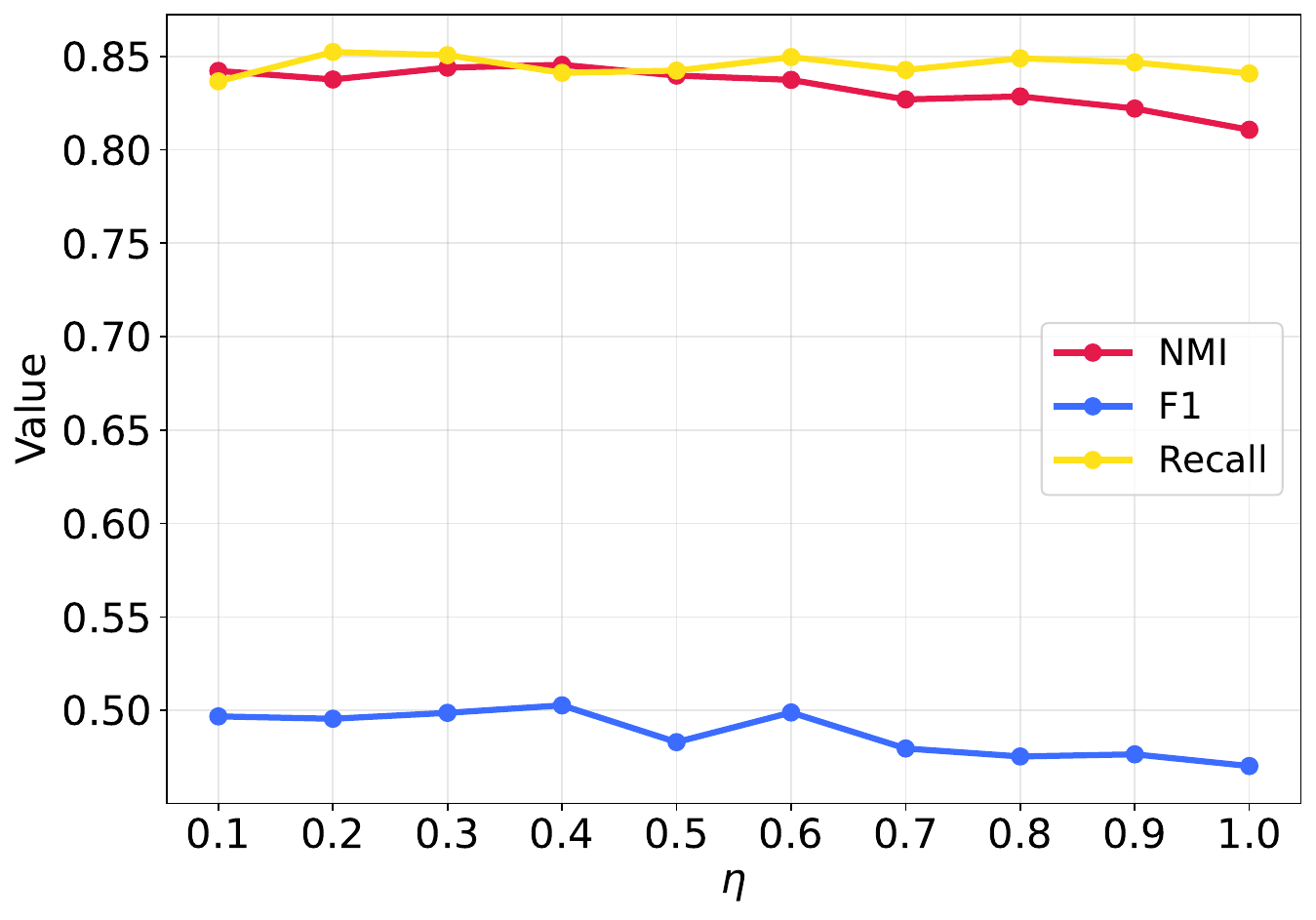}
\captionsetup{hypcap=false}
\captionof{figure}{Impact of hyperparameter $\eta$ on the  NMI, F1 and Recall metrics in DML task (left: CUB-200-2011, right: Cars-196).}
\label{fig:dml_eta}
\end{minipage}
\end{center}

\section{Conclusions}
This paper proposes an efficient and dynamic graph structure learning algorithm DGSL-RCF). By constructing a novel resistive curvature measure and a geometric flow learning paradigm, this research effectively overcomes the computational complexity bottlenecks associated with traditional Ricci curvature flows (such as OCF, OCF), opening up new theoretical pathways for large-scale graph geometric optimization.

The main contributions and findings of this study are summarized as follows:
1) We formally defined the Resistive Curvature Flow and derived its dynamic evolution equation. Theoretical analysis demonstrates that the RCF utilizes curvature gradients to drive nonlinear redistribution of edge weights, facilitating "geometric self-healing" and manifold recovery within graph structures.
2) Through theoretical arguments and experimental validation, we have proven that the RCF maintains excellent topological discriminative power while achieving high computational efficiency and differential compatibility. This allows it to be seamlessly integrated into deep neural networks requiring high-frequency iterations.
3) Experimental results confirm that the RCF can acutely identify and suppress "shortcut edges" generated by sampling noise, while simultaneously strengthening the local cluster structures within the manifold. This significantly enhances the discrete graph's approximation accuracy to the underlying continuous manifold.
4) We designed and validated three integration modes: preprocessing, hidden-layer regularization, and output refinement. Across various tasks, including manifold learning, deep metric learning, and graph structure learning, the RCF demonstrated outstanding universality, robustness, and generalization capability.

Future research will focus on extending the application of RCF to ultra-large-scale graph structures and specific industrial scenarios, exploring adaptive flow-rate scheduling strategies to further optimize convergence performance on massive datasets. Furthermore, integrating RCF with generative models (such as diffusion models) to explore its potential in geometry-aware generative tasks represents a highly promising research direction.

\section{Acknowledgements}
This work was supported in part by the Young Elite Scientists Sponsorship Program by China Association for Science and Technology (2022QNRC001), the National Natural Science Foundation of China (62406315), China Postdoctoral Science Foundation (2025M771504), GuangDong Basic and Applied Basic Research Foundation (2024A1515110108),  Shaanxi Provincial Key Research and Development Program (2025SF-YBXM-023), and Shaanxi Provincial Public Health Scientific Research Innovation Team Project (202511).

\bibliographystyle{elsarticle-num} 
\bibliography{refs.bib}

\appendix

\section{Details of the Datasets}
\label{appendix:dataset_details}
\begin{itemize}[topsep=0pt, parsep=0pt]
 \item \textbf{MNIST:} This dataset consists of Zalando's article images contains a total of $70,000$ images, $60,000$ images in the training set and $10,000$ images in the test set. Each instance is a $28\times 28$ grayscale image associated with a label from 10 classes. Limited by the performance of the experimental hardware, we select the first $4$ categories, with a total of $28,000$ images, as the experimental data in this work.
 \item \textbf{USPS:} This dataset is obtained from the scanning of handwritten digits from envelopes by the U.S. Postal Service. Each example is a $16\times 16$ grayscale image associated with a label from $10$ classes. There are $7291$ training images and $2007$ test images. We use the training set as our experimental data.
 \item \textbf{Medical MNIST}. A dataset containing multiple modalities of data, including MRI, CT, and X-ray. The dataset is divided into six categories, which are AbdomenCT, BreastMRI, CXR, ChestCT, Hand, and HeadCT. We selected 1000 images from each category, in total 6000 images, as our experimental data.
 \item  \textbf{Kvasir}. A dataset containing images from inside the gastrointestinal tract. The dataset is classified into three important anatomical landmarks, three clinically significant findings, and two categories of images related to endoscopic polyp removal. Each category consists 500 images.
 \item \textbf{CUB-200-2011}. There are $11,788$ bird images in this dataset, including $200$ bird subsets, $5864$ images in the training set and $5924$ in the test set. Each image provides image class marking information, including birds bounding box, key part information, and bird attribute information.
 \item \textbf{Cars-196}.  This dataset contains a total of $16,185$ images of different models of vehicles. When it is used for classification tasks, there are $8054$ images in the training set and $8131$ images in the test set. When used to metric learning tasks, the first $98$ classes are typically used as training sets and the last $98$ classes as test sets.
 \item \textbf{SOP}. Stanford Online Products (SOP) dataset has $22,634$ classes with $120,053$ product images. Each product has approximately $5.3$ images.
 \item \textbf{Wine}. The dataset is a classic multiclass classification dataset containing the results of a chemical analysis of wines grown in the same region in Italy. The dataset includes 178 samples with 13 chemical features such as alcohol content, malic acid, ash, flavanoids, and proline. 
 \item \textbf{Cancer}. The Breast Cancer Wisconsin Diagnostic Dataset (BCWD) contains biopsy features for classifying breast masses as malignant or benign. The dataset includes 569 instances with 30 features extracted from digital images of fine needle aspirate biopsy slides. 
 \item \textbf{Digits}. The dataset consists of 8×8 pixel images of hand-written digits from 0 to 9. It contains 1,797 grayscale images where each image is represented as an 8×8 array of pixel values. The dataset is commonly used for image classification and pattern recognition tasks. 
 \item \textbf{20News}. The 20 Newsgroups dataset is a collection of approximately 20,000 newsgroup documents partitioned evenly across 20 different newsgroups. The newsgroups cover various topics including computer graphics, operating systems, hardware, sports, science, politics, and religion. 
\end{itemize}

The download links of all the abovementioned datasets used in our experiments are presented in Table~\ref{tab:dataset_url}.

\begin{table}[!ht]
\centering
\footnotesize
\caption{Dataset download links}
\label{tab:dataset_url}
\setlength{\tabcolsep}{0.5mm}{
\begin{tabular}{ll}
\hline
  \textbf{Dataset} & \textbf{Download link} \\ \hline
    MNIST & \href{http://yann.lecun.com/exdb/mnist/}{http://yann.lecun.com/exdb/mnist/}\\
 USPS & \href{https://www.kaggle.com/datasets/bistaumanga/usps-dataset}{https://www.kaggle.com/datasets/bistaumanga/usps-dataset}\\
 Medical MNIST & \href{https://www.kaggle.com/datasets/alincijov/medical-mnist-qbits}{https://www.kaggle.com/datasets/alincijov/medical-mnist-qbits}\\
KVASIR & \href{https://www.kaggle.com/datasets/meetnagadia/kvasir-dataset}{https://www.kaggle.com/datasets/meetnagadia/kvasir-dataset}\\
Swiss Roll & \href{https://scikit-learn.org/stable/modules/generated/sklearn.datasets.make_swiss_roll.html}{https://scikit-learn.org/stable/modules/generated/sklearn.datasets.
make\_swiss\_roll.html}\\
S Curve & \href{https://scikit-learn.org/stable/modules/generated/sklearn.datasets.make_s_curve.html}{https://scikit-learn.org/stable/modules/generated/sklearn.datasets.make\_s\_curve.html}\\
Truncated Sphere & \href{https://scikit-learn.org/stable/modules/generated/sklearn.datasets.make_blobs.html}{https://scikit-learn.org/stable/modules/generated/sklearn.datasets.make\_blobs.html}\\
Gaussian Surface & \href{https://scikit-learn.org/stable/modules/generated/sklearn.datasets.make_gaussian_quantiles.html}{https://scikit-learn.org/stable/modules/generated/sklearn.datasets.make\_gaussian\_quantiles.html}\\
\hline
Wine & \href{https://archive.ics.uci.edu/ml/datasets/Wine} {https://archive.ics.uci.edu/ml/datasets/Wine}\\
Cancer & \href{https://archive.ics.uci.edu/ml/datasets/Breast+Cancer+Wisconsin+(Diagnostic)}{https://archive.ics.uci.edu/ml/datasets/Breast+Cancer+Wisconsin+(Diagnostic)}\\ 
Digits & \href{https://scikit-learn.org/stable/modules/generated/sklearn.datasets.load\_digits.html}{https://scikit-learn.org/stable/modules/generated/sklearn.datasets.load\_digits.html}\\
 20News & \href{http://qwone.com/\~jason/20Newsgroups/}{http://qwone.com/\~jason/20Newsgroups/}\\
\hline
 CUB-200-2011 & \href{https://www.vision.caltech.edu/datasets/cub\_200\_2011/}{https://www.vision.caltech.edu/datasets/cub\_200\_2011/} \\
 Cars-196 & \href{https://huggingface.co/datasets/pawlo2013/Cars196}{https://huggingface.co/datasets/pawlo2013/Cars196} \\
 SOP & \href{https://opendatalab.com/OpenDataLab/Stanford\_online\_Products}{https://opendatalab.com/OpenDataLab/Stanford\_online\_Products} \\
\hline
\end{tabular}}
\end{table}

\section{Ollivier-Ricci curvature}
Ollivier-Ricci curvature (ORC) is a discrete notion of Ricci curvature based on optimal transport theory, introduced by Ollivier and refined by Lin and Yau. It quantifies curvature by comparing probability distributions on node neighborhoods using the Wasserstein distance.

For an edge $e$ connecting nodes $u$ and $v$, the Ollivier-Ricci curvature is defined as:

\begin{equation}
\kappa(u, v) = 1 - \frac{W_1(\mu_u, \mu_v)}{d(u, v)},
\end{equation}
where $W_1(\mu_u, \mu_v)$ is the Wasserstein distance (also called Earth Mover's Distance) between the probability distributions at nodes $u$ and $v$, $d(u, v)$ is the shortest-path distance between $u$ and $v$.

The Wasserstein distance is computed by solving an optimal transport problem, measuring the minimal “cost” required to align the two distributions. This approach integrates geometric intuition with probability and optimal transport, offering a powerful tool for analyzing network structure and dynamics.

\section{Gemetric Flow Layer for DML}
\label{appendix:DML_GFL}
Building upon the framework in [18], the geometric flow layer (GFL) is designed to re-represent hidden-layer features through the dynamic guidance of curvature signals. This layer serves as a structural regularizer that injects local geometric priors into the deep learning pipeline.

Let the input features of the $i$-th layer be $X^i = \{X^i_1,X^i_2,\ldots,\\X^i_n\}$, where $n$ is the mini-batch size. The intermediate activation is denoted as $\sigma_j^i =\sigma\left(W^iX^i_j+b^i\right)$. The geometric flow layer then generates the input for the $(i+1)$-th layer via the following curvature-guided aggregation:

\begin{align}\label{18}
	X^{i+1}_j = \frac{\sigma^i_j+\lambda \sum_{l\neq j, l=1}^n\sigma^i_lw^{i+1}_{jl}}{1+\lambda \sum_{l\neq j, l=1}^nw^{i+1}_{jl}},
\end{align}
where $\lambda \ge 0$ is a balancing hyperparameter, and the weights $w^{i+1}_{jl}$ are dynamically evolved according to the RCF defined in Eq.~(5).

\section{DGSL-RCF Integration Paradigms}
\label{appendix:3Paradigms}
We provide three independent integration paradigms, each targeting specific deficiencies in geometric representation learning at different stages of the pipeline to implement precise geometric interventions and enhance the overall geometric fidelity and robustness of the model.

\textbf{Paradigm I: Preprocessing Paradigm}

The raw features $X_{\text{raw}}$ are used to construct a static initial graph $A_{\text{init}}$ via a kNN module. The static $A_{\text{init}}$ is then input into the RCF optimizer, which performs iterative optimization based on the DGSL-RCF algorithm to eliminate noise and unreasonable topological connections, outputting a geometrically refined structure $A_{\text{opt}}$. The optimized $A_{\text{opt}}$ serves as the graph input, which, together with $X_{\text{raw}}$, is fed into the machine learning model. This method is offline and static. It corrects inherent topological defects in the data manifold introduced during the sampling phase, thereby enhancing the noise robustness of the machine learning model.

\textbf{Paradigm II: In-Optimization (Dynamic Regularization Paradigm)}

This paradigm integrates the RCF as a dynamic regularizer within the internal layers of a deep CNN encoder, aiming to optimize the local geometry of the feature space in real-time. This paradigm first dynamically generates an initial adjacency matrix $A_{\text{init}}^{(L)}$ based on the hidden layer features $h^{(L)}$. The DGSL-RCF algorithm is then applied to produce an instantaneously optimized geometric constraint $A_{\text{opt}}$. This $A_{\text{opt}}$ is input to a GFL. The GFL layer applies curvature constraints via a geometric reconstruction operation $A_{\text{opt}} \otimes h^{(L)}$, outputting the regularized features $h_{\text{recon}}^{(L)}$ to the next layer. This mechanism ensures local compactness and geometric stability of the features during propagation through curvature-driven regularization.

\textbf{Paradigm III: Post-Optimization (Joint Refinement Paradigm)}

This paradigm incorporating DGSL-RCF into the model's optimization objective to ensure the learned topological structure is geometrically reasonable. Taking GSL as an example, the GSL model first predicts a graph structure $A_{\text{pre}}$. The DGSL-RCF algorithm then refines it to obtain $A_{\text{opt}}$. The geometric loss $\mathcal{L}_{\text{RCF}}$, calculated based on $A_{\text{opt}}$, is fed back (as indicated by the dashed line) into the total loss function of the GSL model and the downstream specific task model for joint optimization.

Joint Optimization Objective: The overall training objective for the GSL model is defined as:
 \begin{equation}
 \min_{\Theta} \mathcal{L}_{\text{total}} = \mathcal{L}_{\text{task}}(X_{\text{raw}}, A_{\text{opt}}) + \lambda \cdot \mathcal{L}_{\text{RCF}}(A_{\text{opt}}),
 \end{equation} 
where, $\Theta$ represents the parameters of the GSL model, $\mathcal{L}_{\text{task}}$ is the downstream task loss, and $\lambda$ is a hyperparameter balancing the contribution of the geometric loss. The geometric loss $\mathcal{L}_{\text{RCF}}$ encourages the edge curvatures $k_{ij}$ in $A_{\text{opt}}$ to be uniformly distributed or close to a preset target, thereby preventing the learning of geometrically unreasonable connections and enhancing the structure's physical plausibility.

\section{Supplementary Hyperparameter Settings}
\label{appendix:parameter_setting_ml_dml}
Table~\ref{tab:dml_parameter} provides the description of the hyperparameters for Table~2 presented in main text. Table~\ref{tab:ml_parameters} provides the description of the hyperparameters for Table~3 presented in main text. OCF and RCF correspond to the hyperparameter settings of the methods based on DGSL-OCF and DGSL-RCF, respectively. For real-world and synthetic datasets, we set $k$ to 45 and 10, and the target dimensionality $d$ to 8 and 2, respectively. For Table~4 presented in main text, the SLAPS component within the SLAPS+DGSL-RCF model adheres to the configuration in [29]; the hyperparameters for the DGSL-RCF algorithm across all four experimental datasets are uniformly configured as: $k=10$, $\eta=0.2$, and $n\_iter=1$.

\begin{table}[htbp]
\centering
\setlength{\tabcolsep}{2pt}
\small
\caption{Hyperparameter settings for DML Task}
\label{tab:dml_parameter}
\begin{tabular}{lccccccc}
\hline
Dataset & Loss & batch size & r & n\_iter & $\eta$ & k & instances \\
\hline
\multirow{4}{*}{CUB} & Triplet & 40 & 0.1 & 3 & 0.1 & 20 & 20 \\
& Semi-Triplet & 44 & 0.1 & 3 & 0.9 & 10 & 22 \\
& N-Npair & 40 & 0.1 & 3 & 0.8 & 20 & 20 \\
& Multi-Simi & 40 & 0.1 & 3 & 0.8 & 20 & 20 \\ \hline
\multirow{4}{*}{Cars} & Triplet & 40 & 0.1 & 3 & 0.8 & 20 & 20 \\
& Semi-Triplet & 40 & 0.1 & 3 & 0.8 & 20 & 20 \\
& N-Npair & 40 & 0.1 & 3 & 0.8 & 20 & 20 \\
& Multi-Simi & 40 & 0.1 & 3 & 0.8 & 20 & 20 \\ \hline
\multirow{4}{*}{SOP} & Triplet & 12 & 0.2 & 3 & 0.1 & 6 & 6 \\
& Semi-Triplet & 20 & 0.1 & 4 & 0.8 & 10 & 10 \\
& N-Npair & 30 & 0.1 & 1 & 0.8 & 15 & 15 \\
& Multi-Simi & 20 & 0.1 & 2 & 0.8 & 10 & 10 \\ \hline
\end{tabular}
\end{table}

\begin{table}[htbp]
    \centering
    \small
    \caption{Hyper-paramter setting for Manifold Learning Task}
    \label{tab:ml_parameters}
    \begin{tabular}{lccccc}
        \toprule
        \multirow{2}{*}{\textbf{Dataset}} & 
        \multicolumn{3}{c}{\textbf{DGSL-OCF}} & 
        \multicolumn{2}{c}{\textbf{DGSL-RCF}} \\
        \cmidrule(lr){2-4} \cmidrule(lr){5-6}
        & $\mathbf{n\_iter}$ & $\mathbf{\eta}$ & $\mathbf{\alpha}$ & 
        $\mathbf{n\_iter}$ & $\mathbf{\eta}$ \\
        \midrule
        USPS           & 1 & 1.0 & 0.5 & 4 & 0.1 \\
        MNIST          & 1 & 0.6 & 0.1 & 1 & 0.1 \\
        Medical MNIST  & 3 & 1.0 & 0.1 & 1 & 1.0 \\
        KVASIR         & 3 & 0.7 & 0.3 & 1 & 0.2 \\
        S Curve        & 6 & 1.0 & 0.1 & 6 & 0.7 \\
        Swiss Roll     & 6 & 1.0 & 0.1 & 6 & 0.7 \\
        Sphere         & 6 & 0.1 & 0.3 & 7 & 1.0 \\
        Gaussian Surface & 1 & 1.0 & 0.1 & 6 & 0.7 \\
        \bottomrule
    \end{tabular}
\end{table}

\section{Supplementary DML Experiments for Impact of Batch Size on DML}
\label{appendix:dml_batch_size_exp}
\begin{figure*}[htbp]
  \centering
  \begin{minipage}[b]{\textwidth}
    \centering
    	\includegraphics[width=0.32\columnwidth]{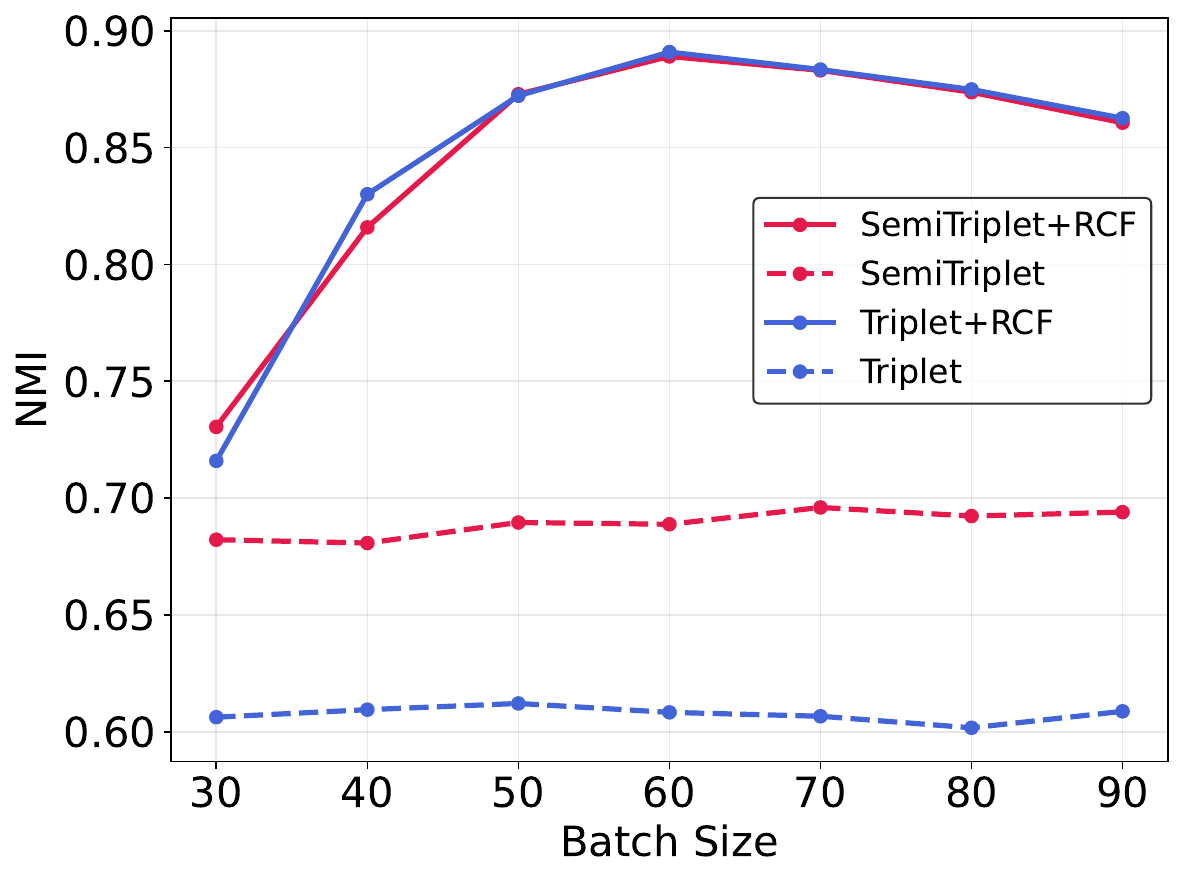}
    \hfill
    \includegraphics[width=0.32\columnwidth]{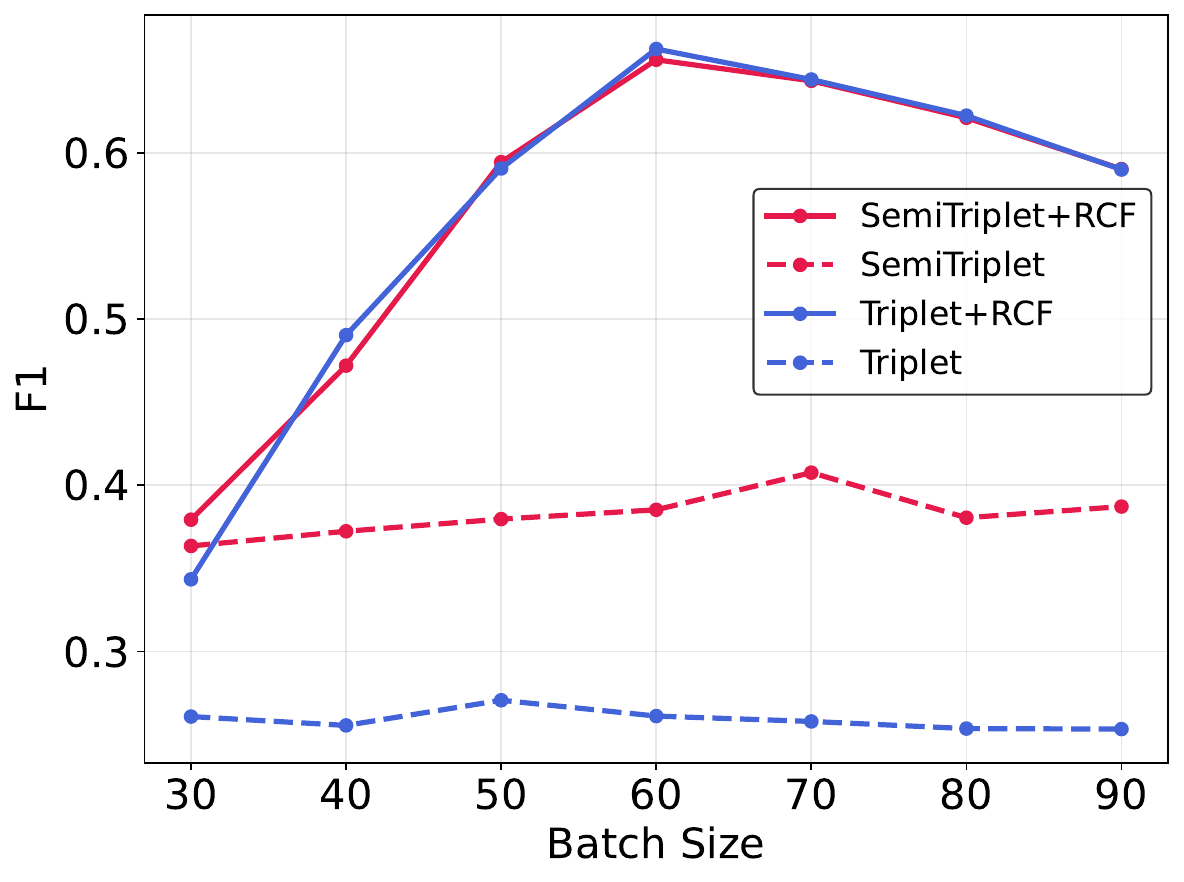}
    \hfill
    \includegraphics[width=0.32\columnwidth]{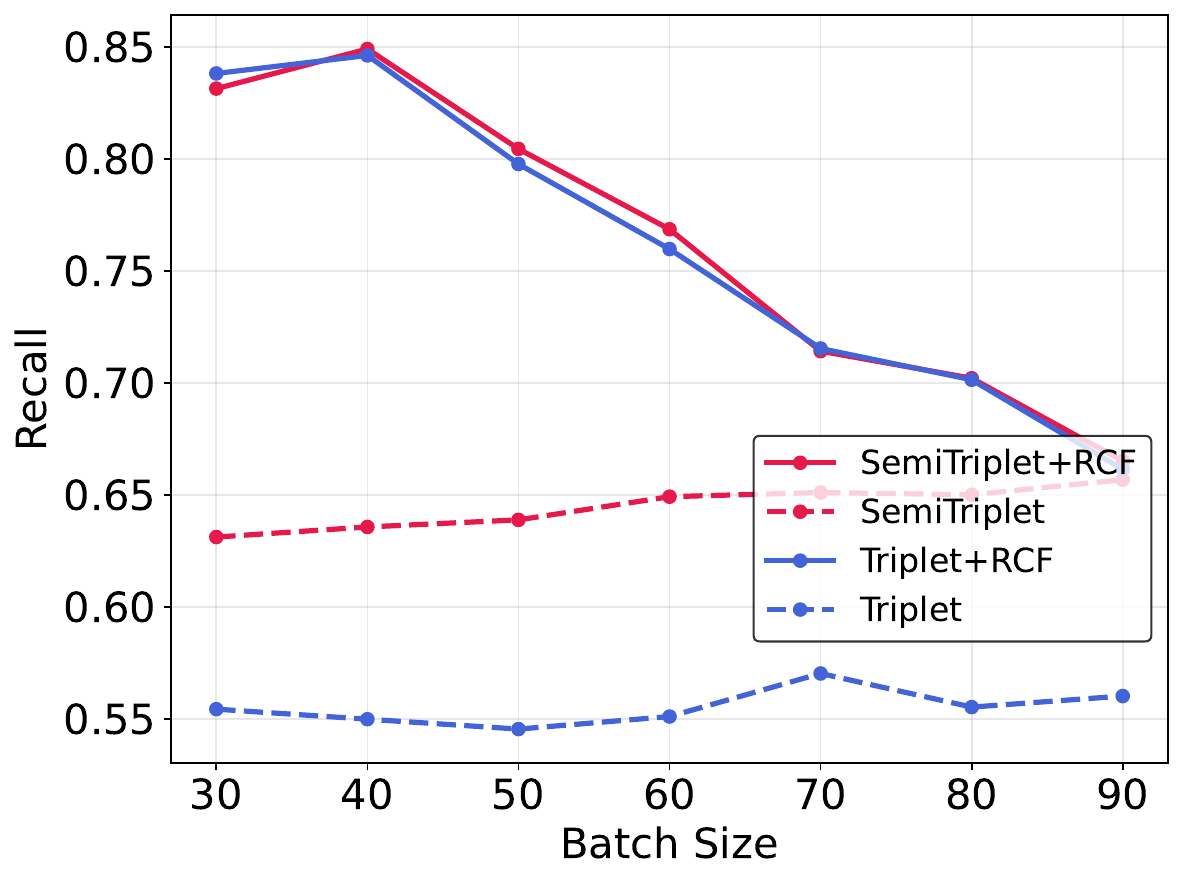}
  \end{minipage}%
  \hfill 
  \\
  \begin{minipage}[b]{\textwidth}
    \centering
    	\includegraphics[width=0.32\columnwidth]{figs/cub_BatchSize_NMI.pdf}
        \hfill
    \includegraphics[width=0.32\columnwidth]{figs/cub_BatchSize_F1.pdf}
    \hfill
    \includegraphics[width=0.32\columnwidth]{figs/cub_BatchSize_Recall.pdf}
  \end{minipage}
  \caption{Batch size for DML}
  \label{fig:dml_batch_size}
\end{figure*}

To investigate how Batch Size affects the performance of DML tasks under the RCF framework, we conducted experiments on the CUB-200-2011 and Cars-196 datasets. Batch Sizes were varied from $30$ to $80$ using Triplet and Semi-hard Triplet losses. The results are illustrated in Figure~\ref{fig:dml_batch_size}.

Positive Gains in Clustering Metrics: The results demonstrate that batch size significantly influences the +RCF method. A moderate increase in batch size leads to a substantial improvement in NMI and F1 scores. This suggests that larger batches provide RCF with a more comprehensive view of the local geometry, enabling more precise reinforcement of the "similar samples closer, dissimilar samples farther" paradigm.

The "Recall-Inversion" Phenomenon: Interestingly, while Recall initially increases with batch size at lower scales ($<40$), it begins to gradually decline as the batch size continues to expand.

Mechanistic Analysis: This divergence (increasing NMI/F1 vs. decreasing Recall) suggests that in larger batches, RCF directs the model's focus toward learning the most prominent and universal manifold patterns. While this optimizes the global discriminative structure, which leading to higher clustering accuracy, it may lead to the neglect of specific "difficult" or "outlier" positive samples, thereby reducing the retrieval recall for fringe cases. This indicates that RCF further intensifies the structural refinement, prioritizing high-density geometric clusters.

\section{Supplementary ML experiments}
\subsection{Supplementary Details for Figure.~2}
\label{appendix:rcf_ocf_cost_cmp}
Table~\ref{tab:s_curve_swiss_roll_cost} serves as a data supplement to Figure~2, detailing the computational costs of OCF (abbreviated as O) and RCF (abbreviated as R), as well as their speed-up ratio (O/R). Based on these detailed results, we derive the following key observations:
\begin{itemize}[topsep=0pt, parsep=0pt]
    \item Optimal Graph Size for Speed-Up Ratio: The speed-up ratio of RCF relative to OCF reaches its maximum when the graph size is 2000. This finding provides crucial insights for adjusting the Batch Size in deep learning models, suggesting that an optimal efficiency point may exist at specific computational scales.
    \item Speed-Up Ratio Increases with $k$: Under the same graph size, an increase in the value of $k$ significantly enhances the speed-up ratio of RCF compared to OCF. For instance, on the S Curve dataset with a size of 5000, as $k$ increases from 10 to 30 and then to 50, the speed-up ratio (O/R) rises from 29.70 to 36.51 and further to 55.19. This indicates that the efficiency advantage of the RCF method becomes more pronounced when handling higher graph densities or more complex neighborhood relationships.
    \item RCF Insensitivity to Graph Density: Across all tested graph sizes, different $k$ values have a minimal impact on the computation time of RCF, demonstrating that its computational efficiency is largely insensitive to changes in graph density. In contrast, the computation time of OCF increases significantly with higher $k$ values. For example, on the S Curve dataset with a size of 5000, as $k$ increases from 10 to 30 and then to 50, the computation time of RCF shows only minor fluctuations between 0.44, 0.43, and 0.46. Meanwhile, the computation time of OCF rises markedly from 12.94 to 15.75 and 25.35. This confirms that the RCF method exhibits enhanced computational stability and scalability.
\end{itemize}

\begin{table*}[htbp]
\centering
\scriptsize
\caption{
Computational cost and speed-up ratio of OCF and RCF under varying graph sizes and $k$ values for S Curve and Swiss Roll datasets.}
\label{tab:s_curve_swiss_roll_cost}
\setlength{\tabcolsep}{1.2pt}
\begin{tabular}{c|ccc|ccc|ccc|ccc|ccc|ccc}
\toprule
\multirow{2}{*}{\textbf{Size}} & \multicolumn{9}{c|}{\textbf{S Curve}} & \multicolumn{9}{c}{\textbf{Swiss Roll}} \\
\cmidrule(lr){2-10} \cmidrule(lr){11-19}
& \multicolumn{3}{c}{$k=10$} & \multicolumn{3}{c}{$k=30$} & \multicolumn{3}{c}{$k=50$} & \multicolumn{3}{c}{$k=10$} & \multicolumn{3}{c}{$k=30$} & \multicolumn{3}{c}{$k=50$} \\
\cmidrule(lr){2-4} \cmidrule(lr){5-7} \cmidrule(lr){8-10} \cmidrule(lr){11-13} \cmidrule(lr){14-16} \cmidrule(lr){17-19}
& O & R & O/R & O & R & O/R & O & R & O/R & O & R & O/R & O & R & O/R & O & R & O/R \\
\midrule
1000 & 6.39 & 0.09 & 69.15 & 7.32 & 0.08 & 97.08 & 11.90 & 0.11 & 109.86 & 6.04 & 0.07 & 88.99 & 7.42 & 0.07 & 102.88 & 8.75 & 0.07 & 129.69 \\
2000 & 7.11 & 0.05 & \textbf{145.02} & 8.86 & 0.05 & \textbf{177.32} & 14.04 & 0.05 & \textbf{283.61} & 6.66 & 0.04 & \textbf{162.79} & 8.35 & 0.04 & \textbf{202.55} & 9.58 & 0.04 & \textbf{231.37} \\
3000 & 8.60 & 0.15 & 57.60 & 10.40 & 0.15 & 67.69 & 17.14 & 0.14 & 121.91 & 7.83 & 0.14 & 57.76 & 10.08 & 0.12 & 83.59 & 11.48 & 0.13 & 90.61 \\
4000 & 10.51 & 0.27 & 38.46 & 12.70 & 0.28 & 44.61 & 20.82 & 0.28 & 74.72 & 9.20 & 0.27 & 33.82 & 12.08 & 0.24 & 49.43 & 13.60 & 0.25 & 54.29 \\
5000 & 12.94 & 0.44 & 29.70 & 15.75 & 0.43 & 36.51 & 25.35 & 0.46 & 55.19 & 10.41 & 0.42 & 25.00 & 13.96 & 0.42 & 32.93 & 17.08 & 0.41 & 41.27 \\
6000 & 15.73 & 0.65 & 24.32 & 19.32 & 0.66 & 29.29 & 33.02 & 0.63 & 52.17 & 12.51 & 0.60 & 20.77 & 16.58 & 0.59 & 27.93 & 19.24 & 0.64 & 30.18 \\
7000 & 18.87 & 0.85 & 22.15 & 21.62 & 0.91 & 23.81 & 40.92 & 0.89 & 46.18 & 14.20 & 0.86 & 16.58 & 19.30 & 0.83 & 23.26 & 23.41 & 0.89 & 26.30 \\
8000 & 21.84 & 1.16 & 18.85 & 27.06 & 1.18 & 22.87 & 47.79 & 1.20 & 39.84 & 16.13 & 1.14 & 14.14 & 22.85 & 1.12 & 20.42 & 27.25 & 1.19 & 22.99 \\
9000 & 23.43 & 1.46 & 16.06 & 41.42 & 1.52 & 27.29 & 57.32 & 1.52 & 37.67 & 19.36 & 1.45 & 13.32 & 26.47 & 1.40 & 18.86 & 31.29 & 1.45 & 21.56 \\
10000 & 27.53 & 1.84 & 14.93 & 36.26 & 1.87 & 19.39 & 66.89 & 2.03 & 32.98 & 21.63 & 1.77 & 12.25 & 30.31 & 1.76 & 17.20 & 36.48 & 1.84 & 19.78 \\
\bottomrule
\end{tabular}
\end{table*}

\begin{table*}[htbp]
\scriptsize
\centering
\caption{Cost comparison on Sphere and Gaussian}
\label{tab:sphere_gaussian_cost}
\setlength{\tabcolsep}{1.2pt}
\begin{tabular}{c|ccc|ccc|ccc|ccc|ccc|ccc}
\toprule
\multirow{2}{*}{\textbf{Size}} & \multicolumn{9}{c|}{\textbf{Sphere}} & \multicolumn{9}{c}{\textbf{Gaussian}} \\
\cmidrule(lr){2-10} \cmidrule(lr){11-19}
& \multicolumn{3}{c}{$k=10$} & \multicolumn{3}{c}{$k=30$} & \multicolumn{3}{c}{$k=50$} & \multicolumn{3}{c}{$k=10$} & \multicolumn{3}{c}{$k=30$} & \multicolumn{3}{c}{$k=50$} \\
\cmidrule(lr){2-4} \cmidrule(lr){5-7} \cmidrule(lr){8-10} \cmidrule(lr){11-13} \cmidrule(lr){14-16} \cmidrule(lr){17-19}
& O & R & O/R & O & R & O/R & O & R & O/R & O & R & O/R & O & R & O/R & O & R & O/R \\
\midrule
1000 & 8.39 & 0.09 & 89.06 & 7.65 & 0.09 & 87.56 & 11.62 & 0.08 & 136.74 & 3.03 & 0.06 & 51.21 & 3.49 & 0.06 & 54.20 & 4.67 & 0.06 & 80.08 \\
2000 & 8.95 & 0.05 & \textbf{168.68} & 8.71 & 0.06 & \textbf{150.68} & 13.49 & 0.06 & \textbf{220.10} & 3.48 & 0.03 & \textbf{100.19} & 4.57 & 0.04 & \textbf{126.29} & 5.66 & 0.04 & \textbf{155.55} \\
3000 & 9.99 & 0.15 & 67.78 & 10.29 & 0.17 & 61.08 & 14.99 & 0.18 & 83.93 & 4.25 & 0.11 & 38.07 & 5.76 & 0.11 & 52.80 & 7.36 & 0.11 & 64.94 \\
4000 & 11.14 & 0.29 & 39.01 & 12.01 & 0.29 & 41.87 & 18.46 & 0.28 & 65.10 & 5.38 & 0.22 & 24.26 & 7.28 & 0.23 & 32.13 & 9.27 & 0.23 & 40.67 \\
5000 & 12.04 & 0.46 & 26.33 & 13.84 & 0.46 & 30.00 & 21.84 & 0.46 & 47.41 & 6.94 & 0.37 & 18.67 & 8.86 & 0.37 & 23.69 & 11.54 & 0.37 & 31.56 \\
6000 & 14.52 & 0.67 & 21.55 & 16.23 & 0.68 & 23.90 & 25.51 & 0.69 & 37.19 & 8.73 & 0.54 & 16.31 & 10.96 & 0.54 & 20.35 & 13.90 & 0.55 & 25.48 \\
7000 & 16.97 & 0.91 & 18.69 & 19.55 & 0.92 & 21.31 & 30.45 & 0.92 & 33.22 & 10.19 & 0.75 & 13.62 & 14.05 & 0.75 & 18.75 & 16.50 & 0.75 & 21.99 \\
8000 & 19.07 & 1.18 & 16.15 & 27.55 & 1.18 & 23.27 & 35.59 & 1.19 & 29.88 & 12.83 & 0.98 & 13.08 & 16.83 & 0.99 & 17.05 & 20.86 & 0.99 & 20.96 \\
9000 & 23.29 & 1.49 & 15.63 & 27.44 & 1.42 & 19.30 & 41.58 & 1.48 & 28.17 & 15.69 & 1.26 & 12.48 & 19.92 & 1.27 & 15.73 & 23.34 & 1.27 & 18.36 \\
10000 & 27.23 & 1.82 & 14.98 & 35.21 & 1.82 & 19.32 & 46.37 & 1.80 & 25.80 & 18.62 & 1.57 & 11.89 & 23.18 & 1.55 & 14.91 & 28.45 & 1.60 & 17.81 \\
\bottomrule
\end{tabular}
\end{table*}

\subsection{Other Supplementary Experimental Results for ML Task}
\label{appendix:ml_other_exp}
This section supplements the presentation of the results regarding the hyperparameters $k$, $n\_iter$, and $\eta$ on the NMI metric within the ML tasks (corresponding to Figure~\ref{fig:ml_k_nmi}, Figure~\ref{fig:ml_n_iter_nmi}, and Figure~\ref{fig:ml_eta_nmi}, respectively). The results indicate that the variation trends of these hyperparameters on the NMI metric are consistent with those on the ACC metric (Figure~4, Figure~6, and Figure~8). This consistency demonstrates that the impact of these hyperparameters on model performance remains stable across different evaluation dimensions.

\begin{center}
\begin{minipage}{\columnwidth}
\centering
	\includegraphics[width=0.24\columnwidth]{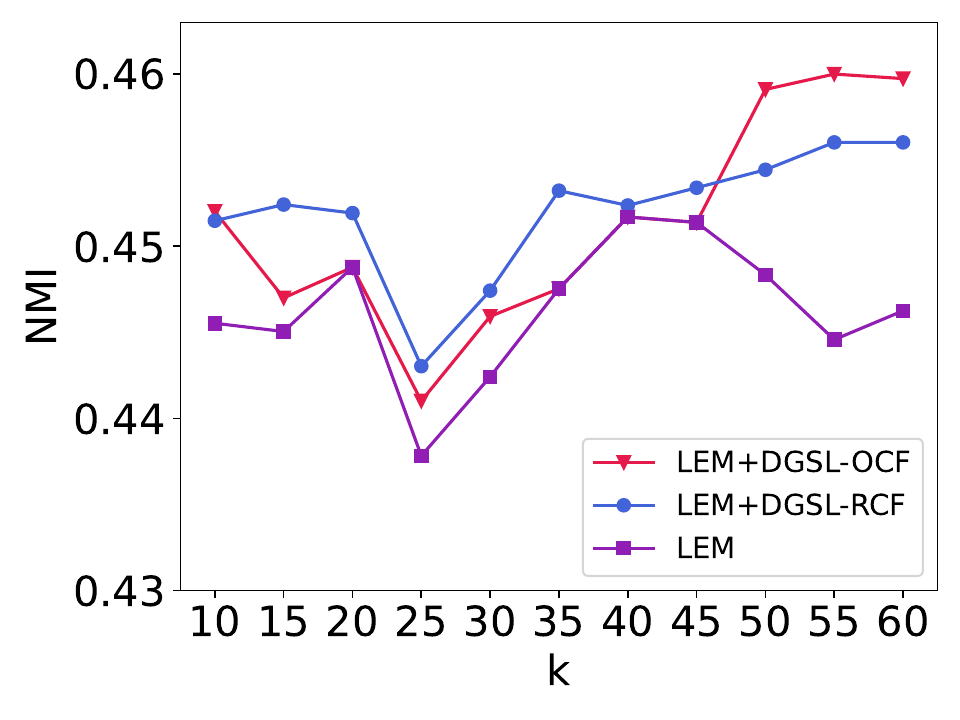}
    \includegraphics[width=0.24\columnwidth]{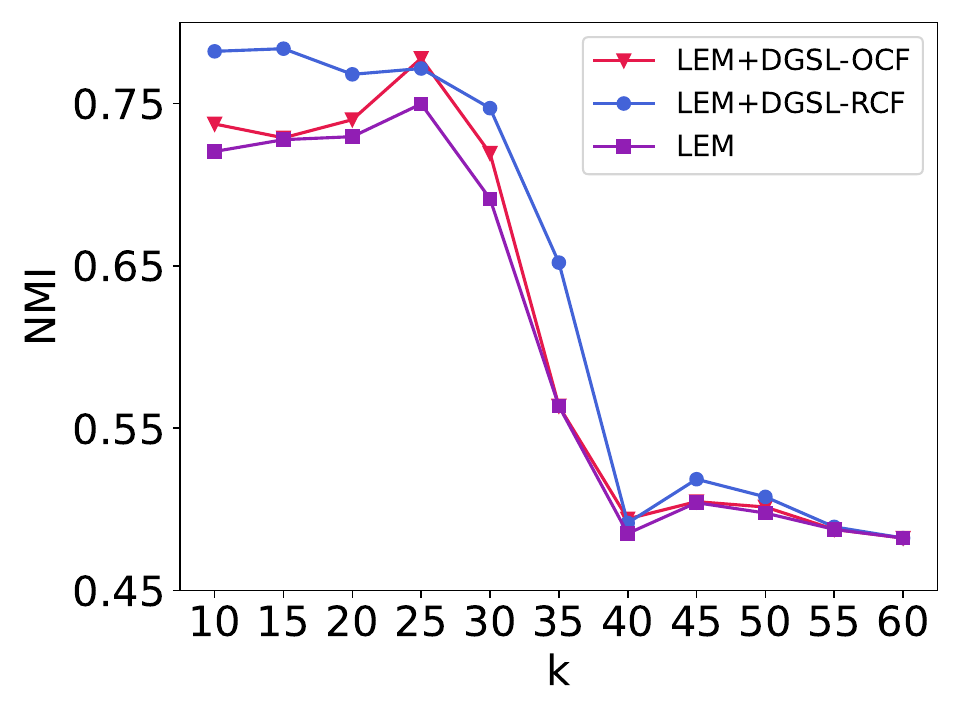}
    \includegraphics[width=0.24\columnwidth]{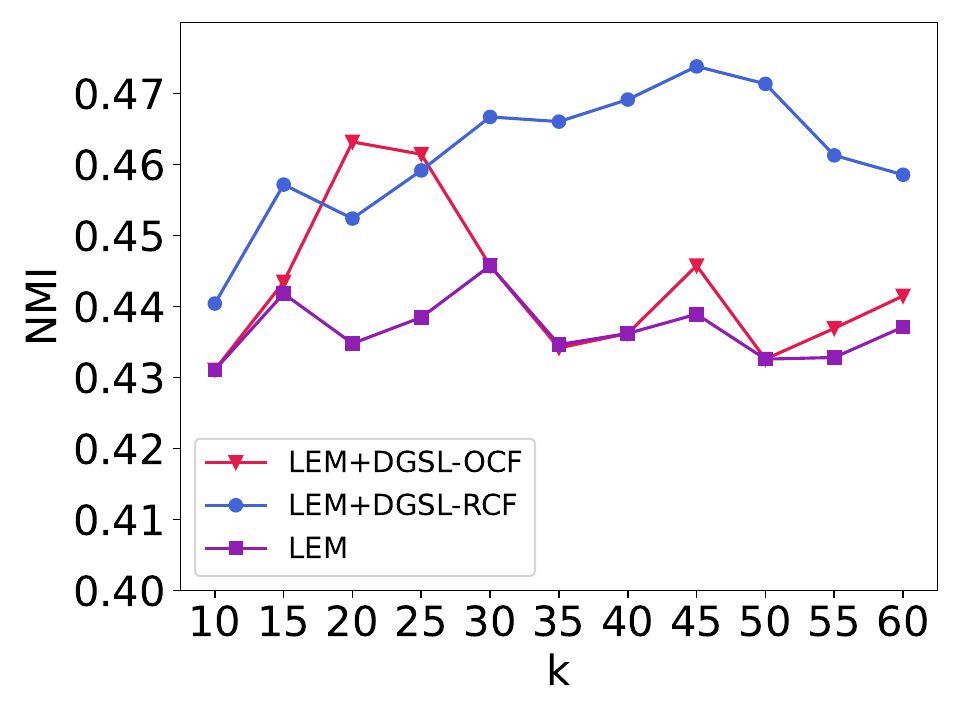}
    \includegraphics[width=0.24\columnwidth]{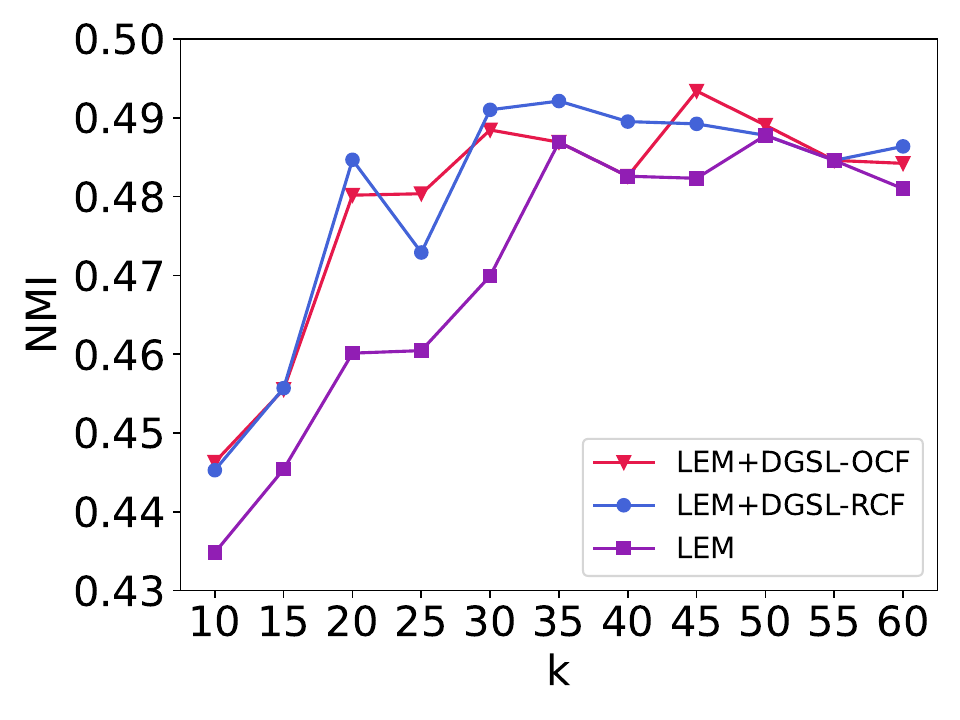}
\captionsetup{hypcap=false}
\captionof{figure}{Impact of hyperparameter $k$ on the NMI metric of LEM, LEM+OCF, and LEM+RCF in ML task (top-left: S Curve, top-right: Swiss Roll, bottom-left: Truncated Sphere, bottom-right: Gaussian Surface).}
\label{fig:ml_k_nmi}
\end{minipage}
\end{center}

\begin{center}
\begin{minipage}{\columnwidth}
\centering
	\includegraphics[width=0.24\columnwidth]{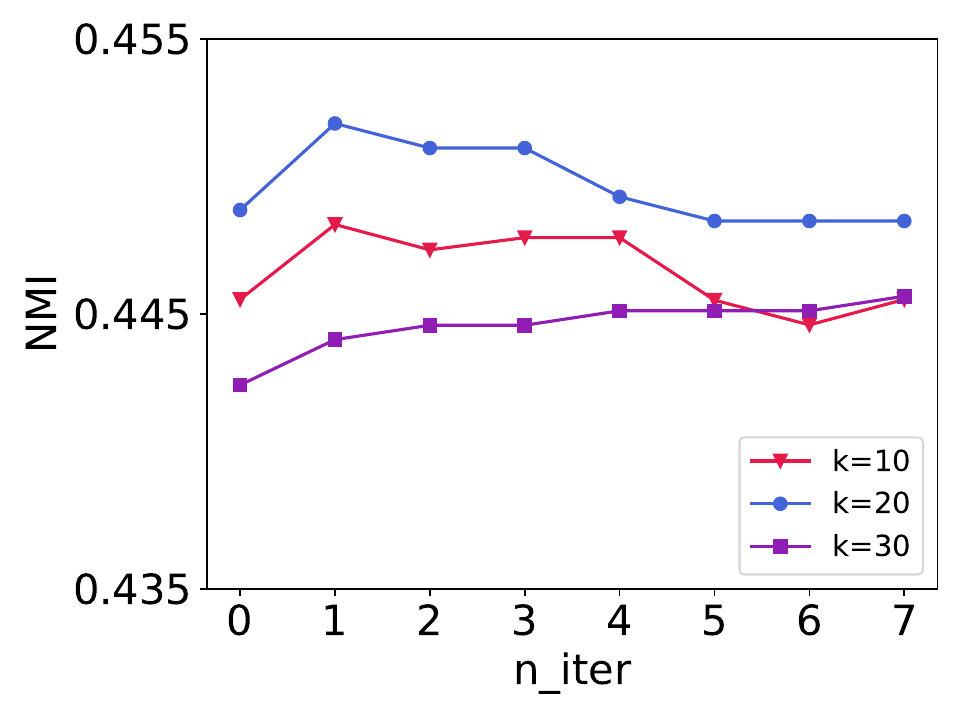}
    \includegraphics[width=0.24\columnwidth]{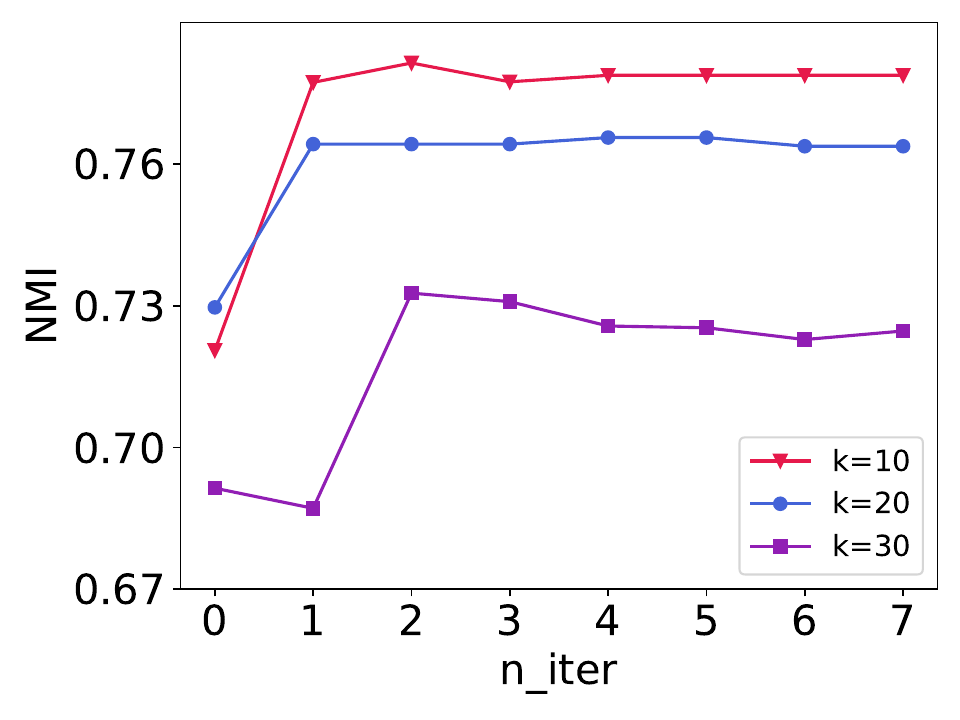}
    \includegraphics[width=0.24\columnwidth]{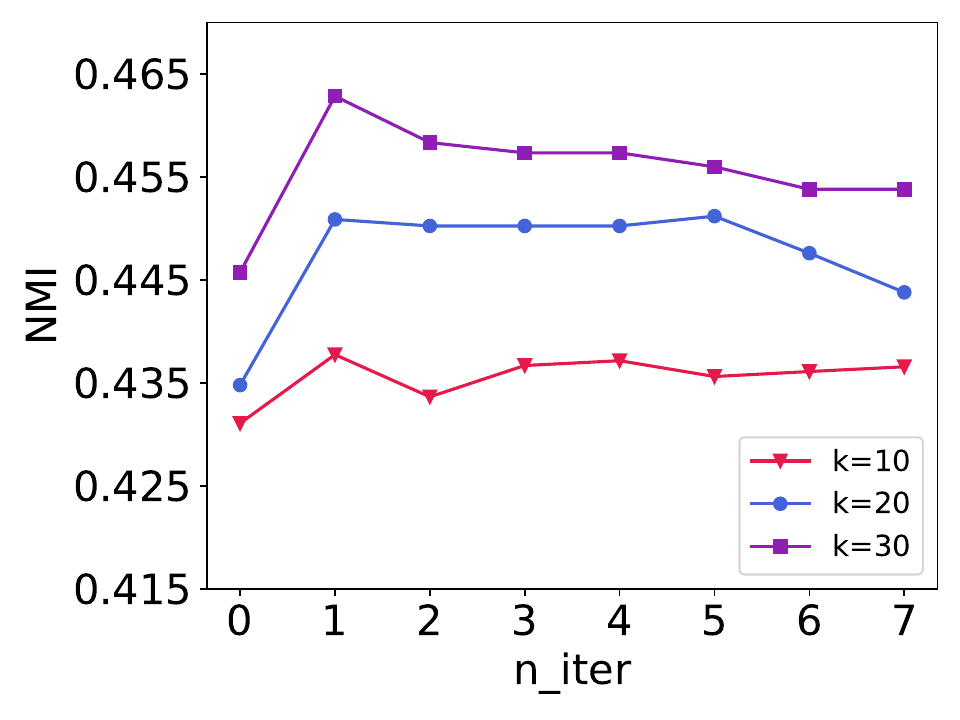}
    \includegraphics[width=0.24\columnwidth]{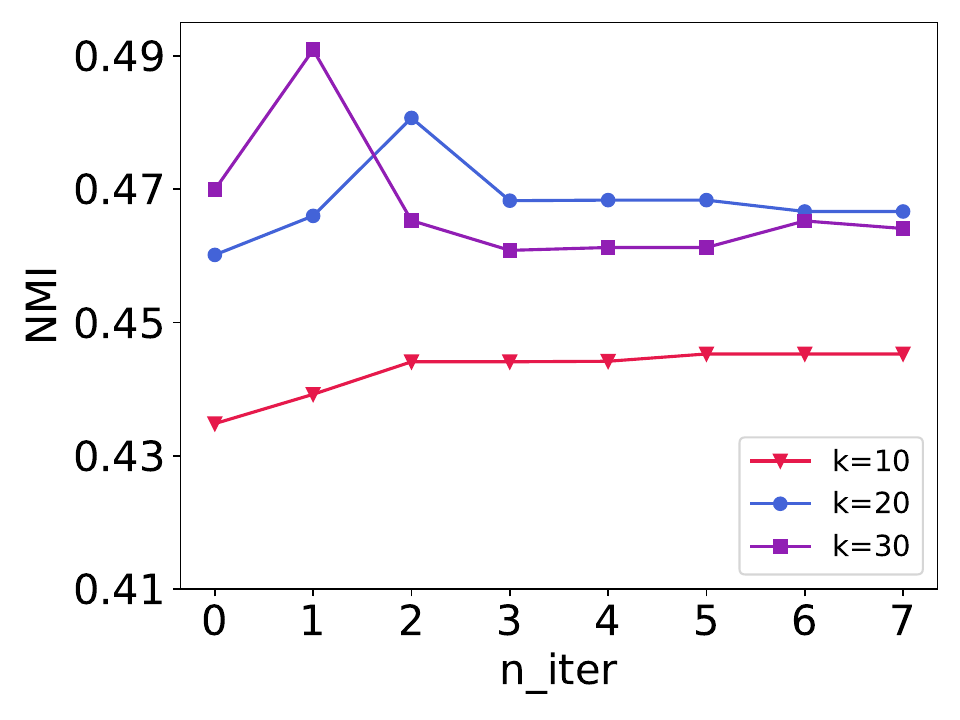}
\captionsetup{hypcap=false}
\captionof{figure}{Impact of hyperparameter $n\_iter$ on the NMI metric in ML task under different $k$ (top-left: S Curve, top-right: Swiss Roll, bottom-left: Truncated Sphere, bottom-right: Gaussian Surface).}
\label{fig:ml_n_iter_nmi}
\end{minipage}
\end{center}

\begin{center}
\begin{minipage}{\columnwidth}
\centering
	\includegraphics[width=0.24\columnwidth]{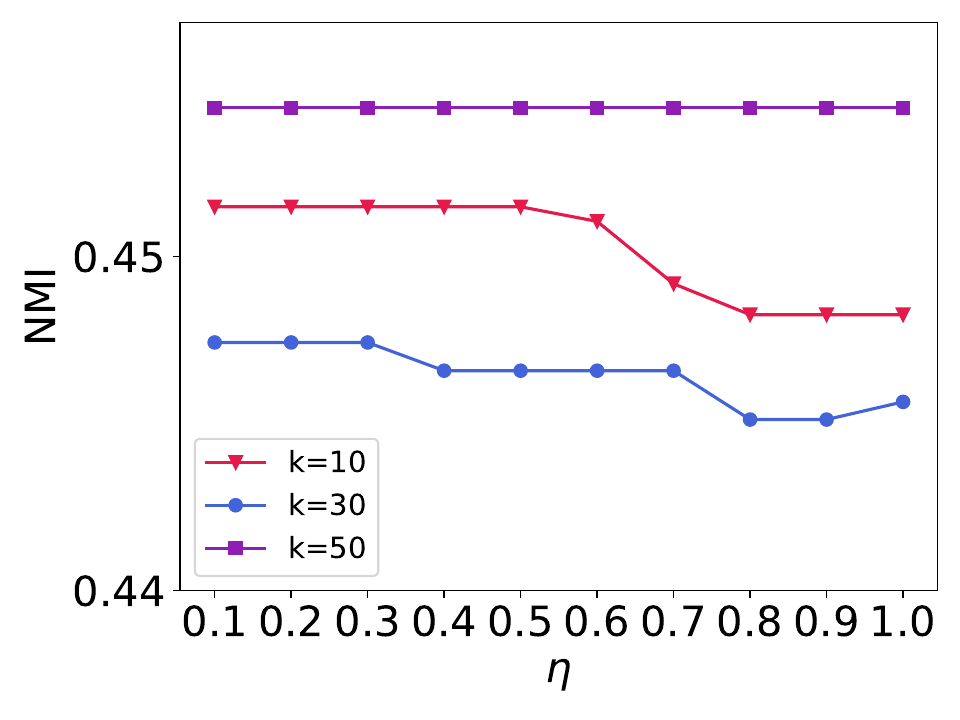}
    \includegraphics[width=0.24\columnwidth]{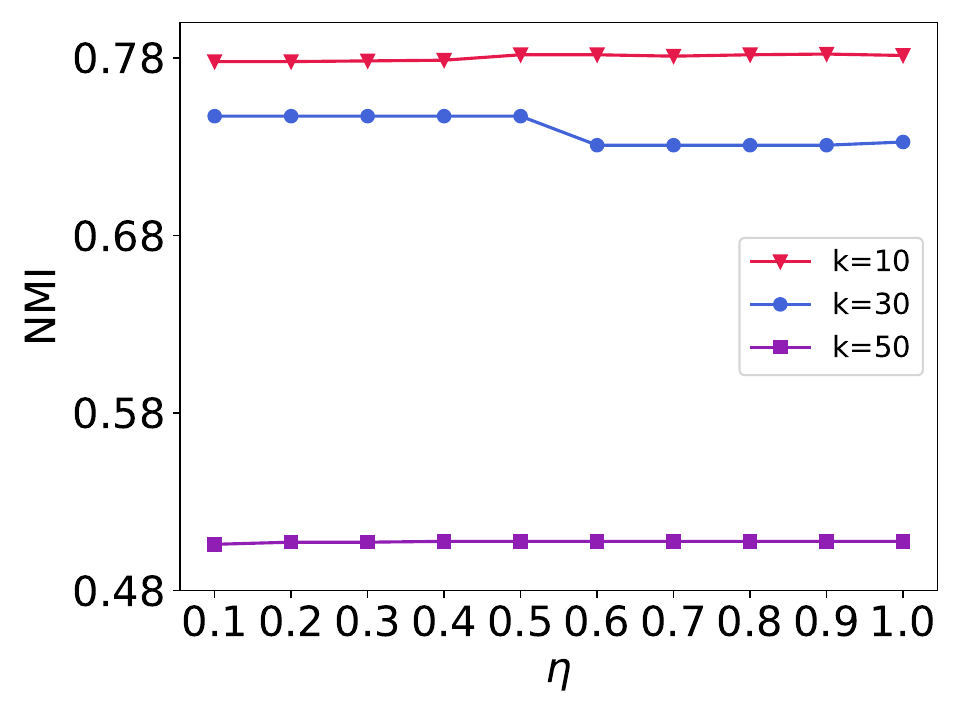}
    \includegraphics[width=0.24\columnwidth]{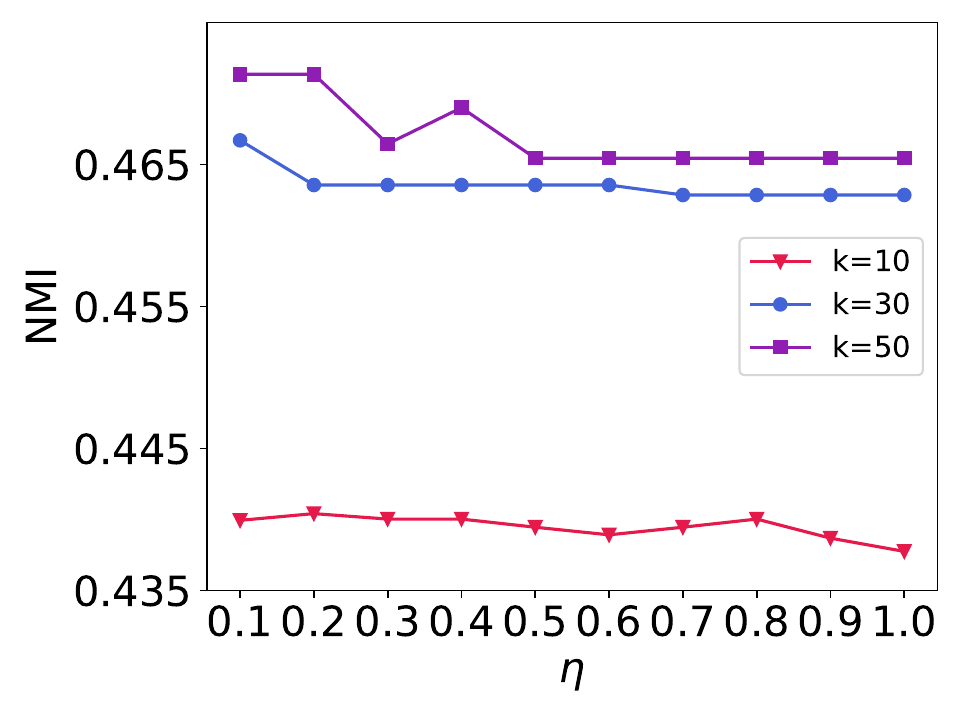}
    \includegraphics[width=0.24\columnwidth]{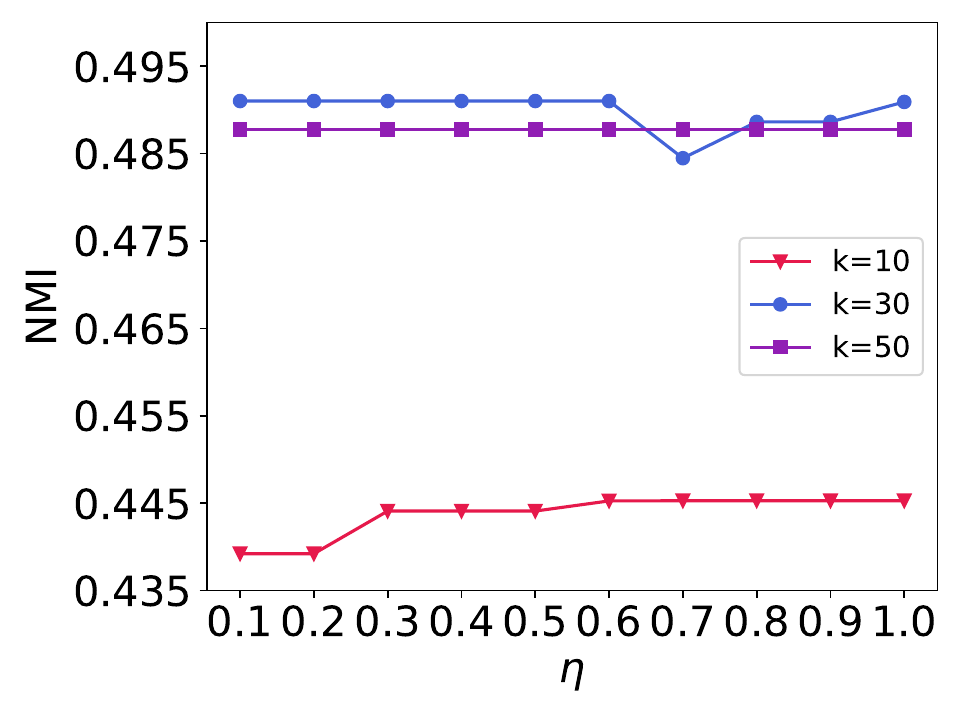}
\captionsetup{hypcap=false}
\captionof{figure}{Impact of hyperparameter $\eta$ on the NMI metric in ML task under different $k$ (top-left: S Curve, top-right: Swiss Roll, bottom-left: Truncated Sphere, bottom-right: Gaussian Surface).}
\label{fig:ml_eta_nmi}
\end{minipage}
\end{center}

\subsection{Abbreviations}
We list all the abbreviations used in this paper and their full forms in Table~\ref{tab:abbr} in alphabetical order.

\begin{table}[!t]
\centering
\footnotesize
\caption{Abbreviations and their full forms}
\label{tab:abbr}
\setlength{\tabcolsep}{1mm}{
\begin{tabular}{llll}
\hline
  \multicolumn{1}{l}{\textbf{Abbr.}} & \multicolumn{1}{c}{\textbf{Full Form}} & \multicolumn{1}{l}{\textbf{Abbr.}} & \multicolumn{1}{c}{\textbf{Full Form}}\\ \hline
  ACC & Accuracy & Isomap & Isometric Mapping \\
ARI & Adjusted Rand Index & LLE   & Locally Linear Embedding\\
    DGSL & Dynamic Graph Structure Learning & LEM    & Laplacian Eigenmaps\\
    DGSL-RCF & Dynamic Graph Structure Learning based on RCF & LTSA    & Local Tangent Space Alignment\\
    DGSL-OCF & Dynamic Graph Structure Learning based on OCF & ML  & Manifold Learning\\
    DML  & Deep Metric Learning & NMI & Normalized Mutual Information\\
    F1 & F1-measure & OCF & Ollivier-Ricci Curvature Flow \\
    GRL  & Geometric Representation Learning & OT & Optimal Transportation\\
GSL  & Graph Structure Learning & RCF & Resistance Curvature Flow \\
HLLE  & Hessian-based Locally Linear Embedding & SPD & Symmetric Positive Definite\\
\hline
\end{tabular}}
\end{table}

\end{document}